\documentclass[sn-mathphys]{sn-jnl}

\usepackage[version=4]{mhchem}
\usepackage{physics}
\usepackage{makecell}
\usepackage{bm}

\jyear{2023}%

\theoremstyle{thmstyleone}%
\theoremstyle{thmstyletwo}%

\theoremstyle{thmstylethree}%

\newcolumntype{?}{!{\vrule width 1pt}}
\renewcommand{\figurename}{Figure}
\renewcommand{\thefigure}{\arabic{figure}}
\renewcommand{\tablename}{Table}
\renewcommand{\thetable}{\arabic{table}}

\newcommand{\beginextendeddata}{
    \setcounter{table}{0}
    \renewcommand{\thetable}{\arabic{table}}
    \renewcommand{\tablename}{Extended Data Table}
    \setcounter{figure}{0}
    \renewcommand{\thefigure}{\arabic{figure}}
    \renewcommand{\figurename}{Extended Data Figure}
 }

\begin{document}

\title[MassFormer]{MassFormer: Tandem Mass Spectrum Prediction for Small Molecules using Graph Transformers}


\author[1,4,5]{\fnm{Adamo} \sur{Young}}\email{ayoung@cs.toronto.edu}

\author*[1,2,4]{\fnm{Hannes} \sur{R{\"o}st}}\email{hannes.rost@utoronto.ca}

\author*[1,3,5,6]{\fnm{Bo} \sur{Wang}}\email{bowang@vectorinstitute.ai}


\affil[1]{\orgdiv{Department of Computer Science}, \orgname{University of Toronto}}

\affil[2]{\orgdiv{Department of Molecular Genetics}, \orgname{University of Toronto}}

\affil[3]{\orgdiv{Department of Laboratory Medicine and Pathobiology}, \orgname{University of Toronto}}

\affil[4]{\orgdiv{Terrence Donnelly Centre}, \orgname{University of Toronto}}

\affil[5]{\orgname{Vector Institute}}

\affil[6]{\orgdiv{Peter Munk Cardiac Centre}, \orgname{University Health Network}}

\abstract{Tandem mass spectra capture fragmentation patterns that provide key structural information about a molecule. Although mass spectrometry is applied in many areas, the vast majority of small molecules lack experimental reference spectra. For over seventy years, spectrum prediction has remained a key challenge in the field. Existing deep learning methods do not leverage global structure in the molecule, potentially resulting in difficulties when generalizing to new data. In this work we propose a new model, MassFormer, for accurately predicting tandem mass spectra. MassFormer uses a graph transformer architecture to model long-distance relationships between atoms in the molecule. The transformer module is initialized with parameters obtained through a chemical pre-training task, then fine-tuned on spectral data. MassFormer outperforms competing approaches for spectrum prediction on multiple datasets, and is able to recover prior knowledge about the effect of collision energy on the spectrum. By employing gradient-based attribution methods, we demonstrate that the model can identify relationships between fragment peaks. To further highlight MassFormer's utility, we show that it can match or exceed existing prediction-based methods on two spectrum identification tasks. We provide  open-source implementations of our model and baseline approaches, with the goal of encouraging future research in this area.}


\keywords{Mass Spectrometry, Metabolomics, Deep Learning, Chemistry, Transformers}



\maketitle

\section{Main}
\label{s:main}

    Mass spectrometry (MS, \cite{ms_textbook,msms_review}) is an analytical technique used for identifying and quantifying chemicals in a mixture. Molecules from the sample are ionized and then detected by a mass analyzer, which records information about the mass to charge ratio (m/z) of each ion in the form of a mass spectrum. Tandem mass spectrometry (MS/MS) is a variant of MS that includes a fragmentation step to isolate and break down charged molecules (called precursors) into smaller fragments. These ions appear as peaks in the fragment spectrum, and their m/z position and relative abundance can be used to make inferences about the molecular structure of the original precursor. When coupled with online liquid chromatography (LC), a technique for chemical separation, the combined LC-MS/MS workflow is a powerful tool for analyzing aqueous solutions. Because of its versatility, LC-MS/MS is commonly employed in a variety of domains, including proteomics \cite{msms_proteomics}, metabolomics \cite{msms_metabolomics,msms_small_mol}, forensics \cite{msms_forensics1,msms_forensics2}, and environmental chemistry \cite{msms_env}.

    For most molecules, it is not possible to accurately simulate the fragmentation that occurs in a mass spectrometer. In principle, theoretical physics provides the tools to understand this process; however, existing first principles simulations are too slow to be used in a high-throughput manner, and rely on approximations that limit their accuracy (for further discussion, see Supplementary Notes). This presents a fundamental problem for the field of mass spectrometry, limiting our ability to analyze the data that is collected. Developing better spectrum simulators is critical for improving our understanding of mass spectrometry and facilitating its more widespread application.
    
    Compound identification from MS/MS data is an important application of mass spectrometry, particularly in metabolomics. Practitioners often rely on database searches with large reference libraries, using spectrum similarity functions \cite{spec_sim} in combination with domain expertise to identify matches. However, these databases have relatively poor coverage, containing on the order of $10^6$ spectra which represent approximately $10^4$ unique compounds. This fails to cover even the relatively small set of human metabolites, which is on the order of $10^5$ according to the human metabolome database (HMDB, \cite{hmdb}). One strategy to overcome the scarcity of spectral libraries is to augment them with simulated (\textit{in silico}) mass spectra for a large number of compounds from a chemical structure database \cite{pubchem,hmdb,kegg}. This can dramatically increase coverage for the reference library and improve the chance of finding a match, further motivating the development of accurate spectrum prediction models.
    
    The increasing availability of public \cite{massbank,gnps,hmdb,respect} and commercial \cite{nist1,nist2,metlin,wiley} MS datasets makes data-driven solutions to spectrum prediction appealing. One of the most popular spectrum prediction methods, competitive fragmentation modelling (CFM, \cite{cfm,cfm3,cfm4}), combines combinatorial framgentation and data-driven probabilistic modelling to predict spectra. CFM has been shown to be effective at predicting and annotating spectra, and can be used for compound identification. However, due to its combinatorial bond-breaking algorithm, CFM struggles with modelling larger compounds (particularly those with multiple rings). Deep learning has proven to be a useful tool for tackling complex problems in a variety of domains \cite{alphazero,alphafold,stable_diffusion,gpt3}. Recently, two types of deep learning methods have been proposed for spectrum prediction: fully-connected neural networks based on molecular fingerprints \cite{neims} and graph neural networks that use a molecular graph representation \cite{gnn_msms,esp}. However, both of these strategies rely on local chemical structures, and do not easily model global interactions between far-away atoms in the molecule. Molecular fingerprints are typically restricted to capturing subgraphs of a fixed size, and cannot represent global structures. Graph neural networks, while more flexible, only model local interactions in a single layer and require increased depth for long-range interactions \cite{how_powerful_gnns}. However, excessive depth can often result in over-smoothing \cite{oversmooth1,oversmooth2}, presenting challenges for larger molecules. While each fragmentation event appears to be a local phenomenon (involving a single bond breakage or a local re-arrangement), the fragment ion intensity recorded by the instrument corresponds to the relative propensity of such a fragmentation event to occur compared to all other possible events. Therefore, a fragment ion intensity cannot be interpreted on a local level without taking the global context of the whole molecule into consideration. Concurrently to this work, a number of deep learning spectrum predictors have been proposed \cite{graff,scarf,3dmolms,rassp}: these works are discussed in the Supplementary Notes.
    
    In this work we adapt a state-of-the-art neural network architecture, the graph transformer \cite{graphormer}, to tackle MS/MS spectrum prediction. Graph transformers model pairwise interactions between all nodes in the graph, and use degree and shortest path information to capture topological properties. Our experiments demonstrate how adapting a pre-trained graph transformer model to spectrum prediction can result in state of the art performance. Through rigorous comparisons with strong baseline methods, we demonstrate that our model can more accurately predict spectra for held-out compounds on two different spectrum datasets. We further validate the quality of our model's predictions by investigating the effects of collision energy on the simulated spectra, and interpret MassFormer's predictions with gradient-based attribution methods. Finally, we demonstrate a realistic use case by applying the model to a spectrum identification problem.

\section{Results}
\label{s:results}

    \subsection{Overview}
    \label{ss:results:overview}
    
\begin{figure}[!htb]
        \centering
        \includegraphics[width=\textwidth]{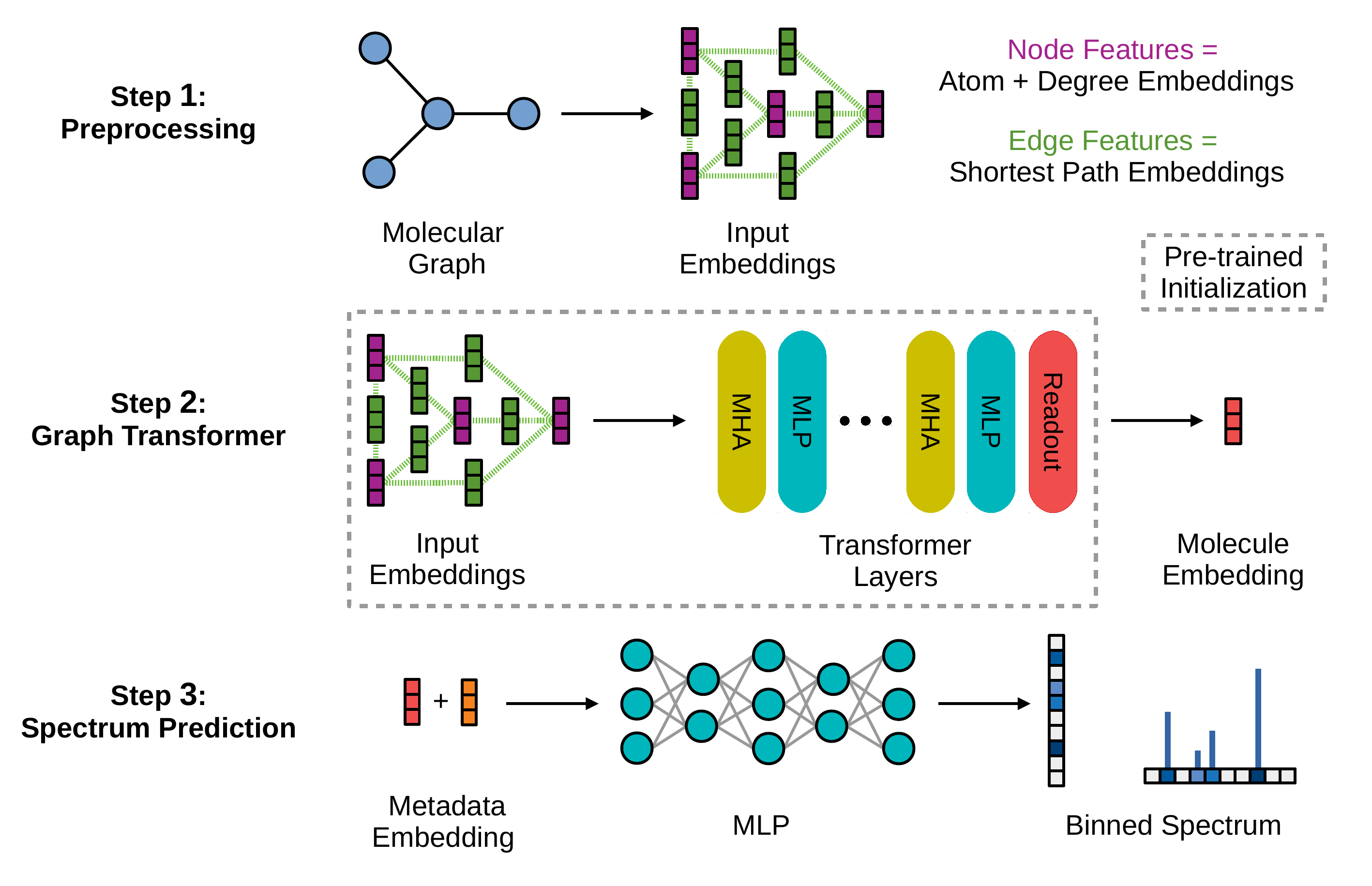}
        \vspace{1em}
        \caption{\textbf{Overview of the method.} Extraction of node and edge embeddings from the molecular graph, application of the graph transformer, extraction of the chemical embedding from the readout node, addition of spectral metadata (such as collision energy), and prediction of the binned spectrum. The parameters for the input embeddings and the graph transformer layers are initialized from a pre-trained model, and the entire model is fine-tuned on the spectrum prediction task. }
        \label{fig:overview}
    \end{figure}

    MassFormer uses a graph transformer architecture (Section \ref{ss:methods:gf}) to predict fragment spectra from an input molecule, represented as a molecular graph (Section \ref{ss:methods:feat}). A visual summary of our method is presented in Figure \ref{fig:overview}. The input graph is first preprocessed into node and edge embeddings. The former encode chemical information about the atoms, such as the element, and centrality properties like degree. The latter capture spatial relations between atoms in the molecule, combining information about shortest path length and edges visited along it. A full description of the featurization can be found in Section \ref{ss:methods:feat}. After preprocessing, the embeddings are passed to a graph transformer, which iteratively applies multi-head self-attention (MHA) and multi-layer perceptrons (MLP) to manipulate the data. The learned self-attention weights capture associations globally between all pairs of nodes, and are directly influenced by edge embeddings at each iteration. After several rounds of processing, the final embeddings are summarized into a single embedding that represents the entire molecule. This chemical representation is combined with spectrum metadata and passed to an MLP that makes a prediction in the form of a sparse positive vector. Each dimension of the vector represents a binned peak location, with the magnitude corresponding to the peak's intensity. The metadata describe important information about the precursor (such as the adduct formed during ionization) and the instrument (such as the collision energy), both of which influence the fragmentation process and resulting spectrum \cite{ms_textbook}. The input embedding and graph transformer parameters are initialized from a Graphormer model \cite{graphormer} that was pre-trained on a large chemical dataset (see Section \ref{ss:methods:pre-train}), then jointly tuned with the spectrum predictor MLP on the MS/MS dataset.

    \subsection{Predicting Spectra for Unseen Compounds}
    \label{ss:results:spec_sim}

    To quantitatively evaluate our model's predictions, we measured average cosine similarity on different types of held-out spectrum data. MassFormer was compared with two deep learning methods, a fingerprint (FP) neural network model and a Weisfeiler-Lehman (WLN) graph neural network model, as well as Competitive Fragmentation Modelling (CFM), a state of the art probabilistic method for spectrum prediction that does not use deep neural networks (see Section \ref{ss:methods:baseline_models} for more details on the baselines). The three deep learning models were trained on the same data: a $70\%$ portion of the NIST 2020 Tandem Mass Spectrometry dataset \cite{nist2}, with an additional $10\%$ Validation set for hyperparameter tuning and early stopping. However, to benchmark CFM we simply used the most recent publicly available pre-trained model \cite{cfm4}, since re-training CFM on our dataset would be non-trivial due to its design.\footnote{CFM was designed to predict Q-TOF spectra for \ce{[M}+\ce{H]+} precursors, while NIST contains Orbitrap spectra for a variety of adduct types.} These differences in training data introduced some ambiguity in comparisons between CFM and other models. 
    
    Model evaluation was performed on the NIST Test set (consisting of the remaining $20\%$ of spectra not used for training and validation), and an additional dataset from MassBank of North America (MoNA, \cite{massbank}). We considered two strategies for splitting the data: a simple random split by compound (InChiKey) and a more challenging split (Scaffold) that stratifies compounds by their Murcko scaffold \cite{scaffold,rdkit} before splitting. For more details on the datasets and training splits, refer to Section \ref{ss:methods:dset}. To allow for a fairer comparison with CFM, we focused on  \ce{[M}+\ce{H]+} spectra and removed compounds from the test sets that were overlapping with CFM's training set. In Extended Data Figure \ref{supfig:dl_sims}, we demonstrate results for the deep learning models on the full range of supported precursor adducts and confirm that they are consistent with the results on the \ce{[M}+\ce{H]+} subset.
    
    \begin{figure}[!htb]
        \begin{subfigure}{0.5\linewidth}
            \centering
            \includegraphics[width=\textwidth]{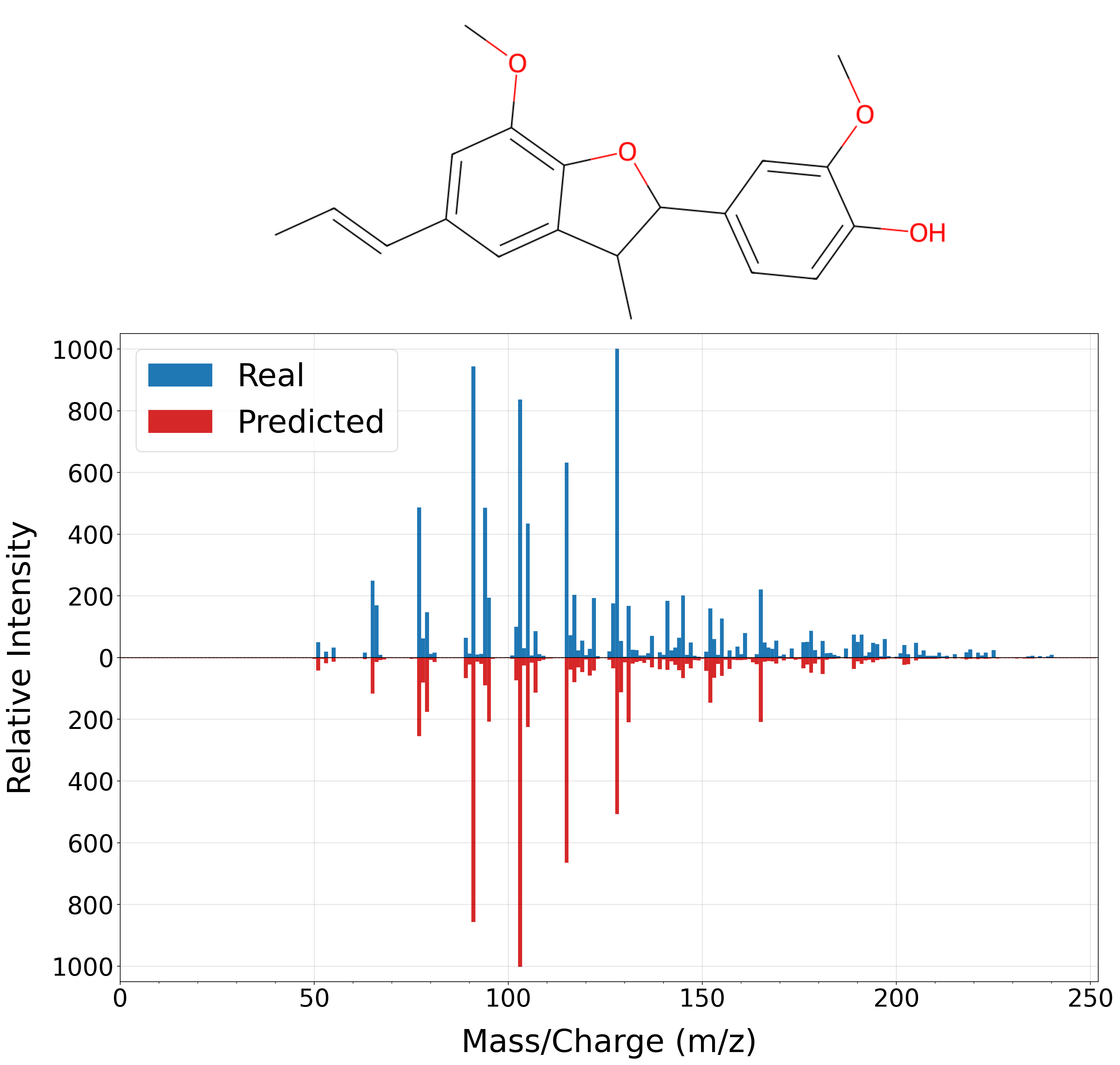}
            \caption{Cosine Similarity: $0.91$ }
            \label{fig:sim:example1}
        \end{subfigure}
        \begin{subfigure}{0.5\linewidth}
            \centering
            \includegraphics[width=\textwidth]{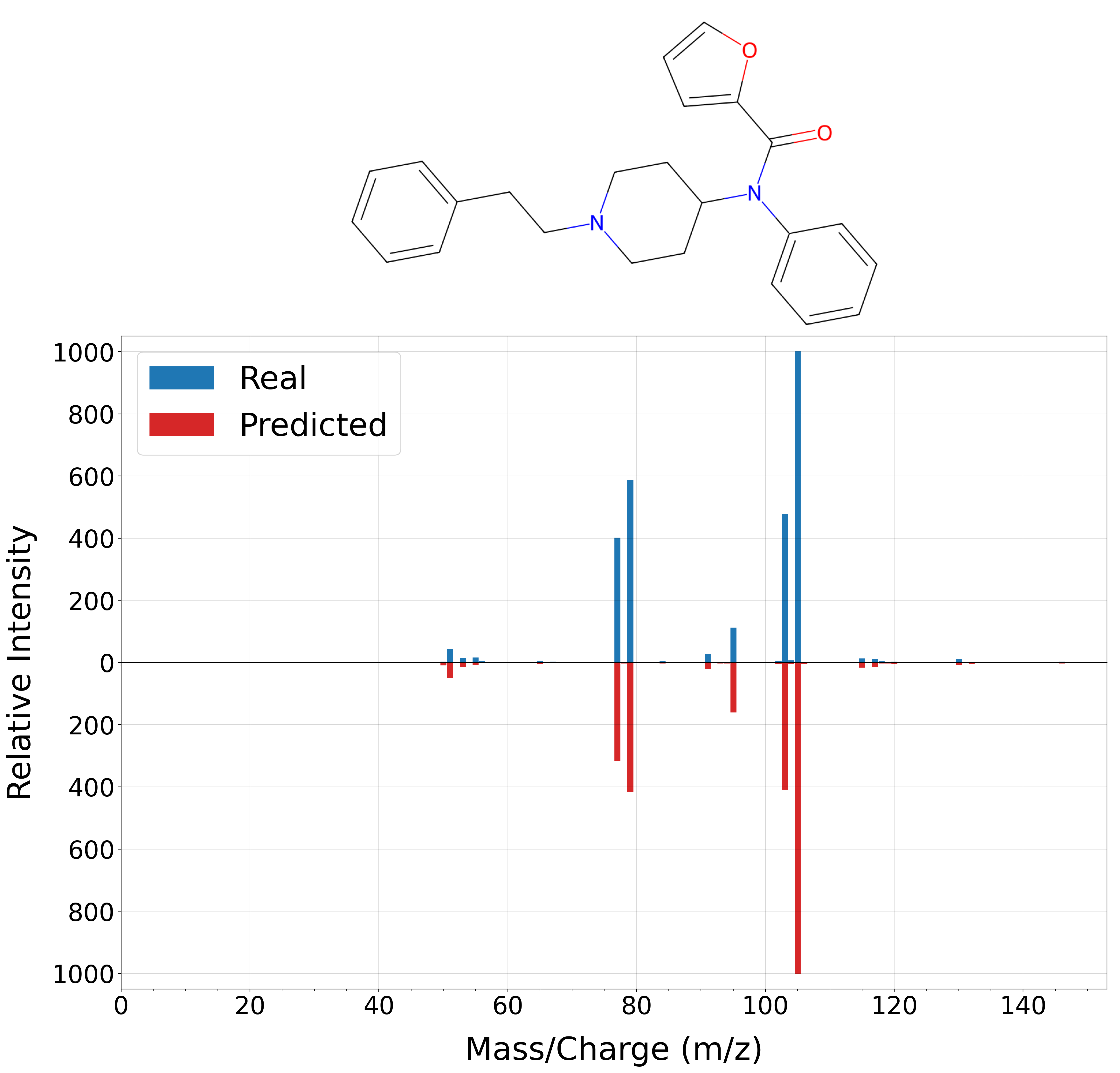}
            \caption{Cosine Similarity: $0.99$}
            \label{fig:sim:example2}
        \end{subfigure}
        \bigskip
        \begin{subfigure}{\linewidth}
            \centering
            \includegraphics[width=0.85\textwidth]{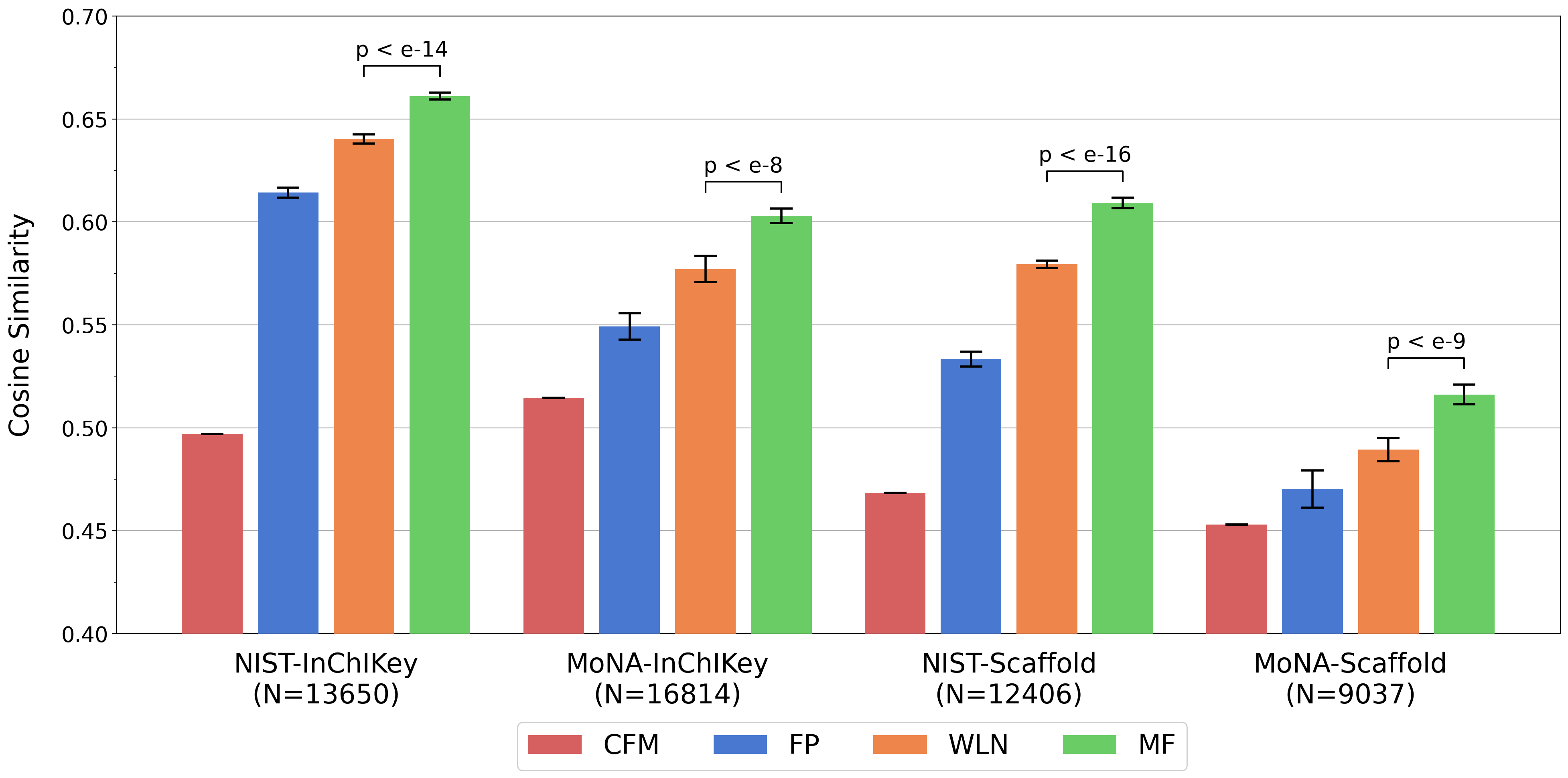}
            \caption{}
            \label{fig:sim:bars}
        \end{subfigure}
        \caption{ \textbf{Spectrum similarity experiments.} MassFormer can accurately predict spectra for Furanylfentanyl (\ref{fig:sim:example1}, an opiod analgesic) and Liparin A (\ref{fig:sim:example2}, a natural product), two examples from the NIST-Scaffold Test set. \ref{fig:sim:bars} compares the average cosine similarity of four models ( CFM = Competitive Fragmentation Modellin, FP = Fingerprint, WLN = Weisfeiler-Lehman Network, MF = MassFormer) on four different data splits. The number of molecules in the training set is provided for each split. To allow for direct comparison with CFM, we limited training and evaluation spectra to \ce{[M+}\ce{H]+} precursor adducts. MassFormer was consistently the best-performing deep learning model. All models struggled more on MoNA than NIST, and Scaffold splitting was more challenging than InChIKey. Averages and standard deviations from 10 independently trained models are reported (except for CFM, which was pre-trained). Statistical significance determined by one-sided Welch's $t$-test with \v{S}id\'{a}k correction. }
        \label{fig:sim}
    \end{figure}
    
    MassFormer reliably outperformed other models across different datasets and split criteria (Figure \ref{fig:sim}). The performance of the three deep learning models was correlated across data splits, reflecting the underlying differences in split difficulty. NIST-InChIKey was the easiest, resulting in the best performance; NIST-Scaffold exhibited split-induced distribution shift; MoNA-InChIKey contained spectra from a different dataset; MoNA-Scaffold presented both of the aforementioned challenges. Ground truth spectra for compounds that existed in both NIST and MoNA tended to be highly similar\footnote{When comparing ostensibly identical spectra from NIST and MoNA (same compound and metadata), on average they scored approximately 0.97 cosine similarity.}, suggesting that batch effect had limited impact. Thus, the decrease in performance on MoNA splits was most likely due to differences in compound and metadata coverage. Altogether, the experiments demonstrate that model performance strongly depended on data splitting technique, and that MassFormer was consistently superior across all splits.
    
    We also performed a more granular investigation of MassFormer's performance by comparing average similarity across the top ten most frequent chemical classes in our dataset (Extended Data Figure \ref{supfig:classyfire}), as indicated by ClassyFire \cite{classyfire}, an automatic chemical ontology tool. While most classes had similar performance, the model seemed to perform exceptionally well on ``lipids and lipid-like molecules". This is perhaps unsurprising, as some types of lipids have predictable fragmentation patterns, facilitating the success of rule-based fragmenters in lipid spectrum predition \cite{cfm3,lipidblast}. Conversely, simulating spectra for lipid-like compounds can be challenging for many generalist methods due to their large size \cite{cfm4}, so MassFormer's success in this area is encouraging.

    \subsection{Modelling the Effect of Collision Energy}
    \label{ss:results:ce}

\begin{figure}[!htb]
         \begin{subfigure}{\linewidth}
            \centering
            \includegraphics[width=\textwidth]{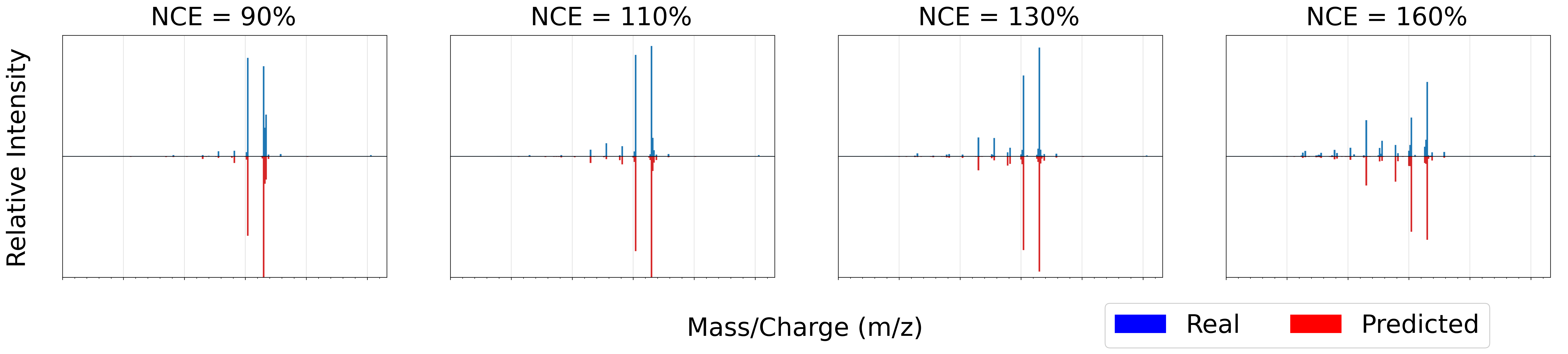}
            \caption{}
            \label{fig:ce:progression}
        \end{subfigure}
        \bigskip
        \begin{subfigure}{0.4\linewidth}
            \centering
            \raisebox{1.4em}{\includegraphics[width=0.7\textwidth]{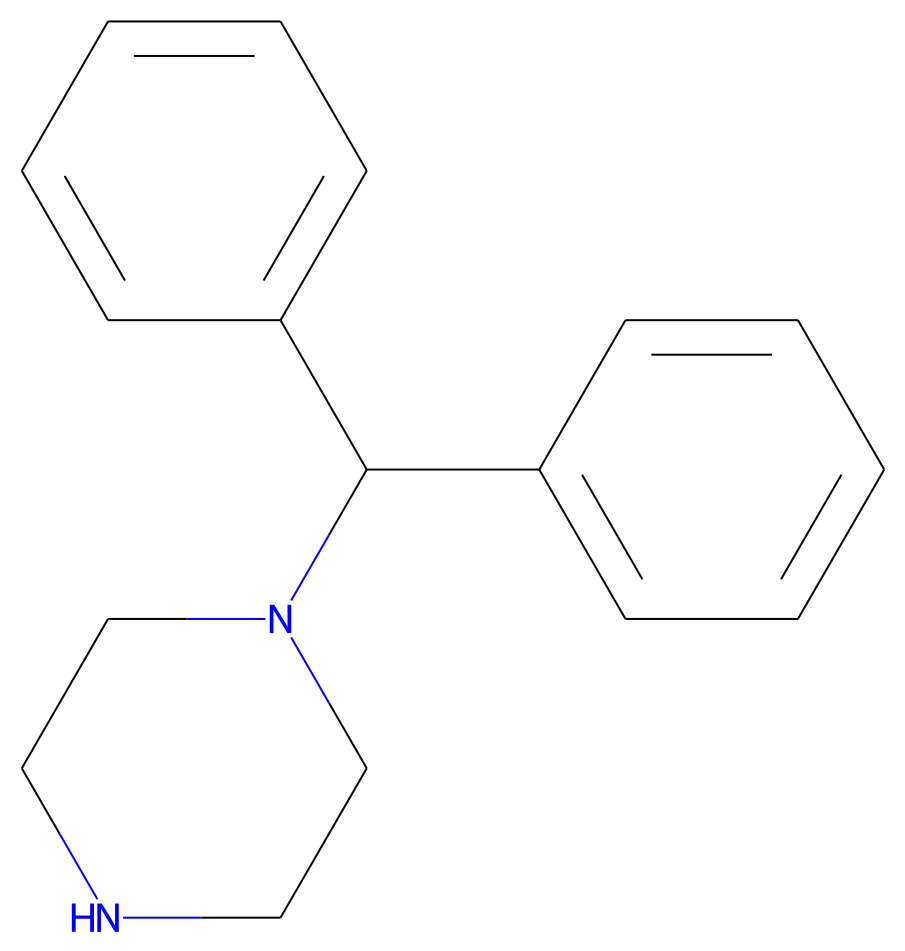}}
            \caption{ }
            \label{fig:ce:molecule}
        \end{subfigure}
        \begin{subfigure}{0.6\linewidth}
            \centering
            \includegraphics[width=0.90\textwidth]{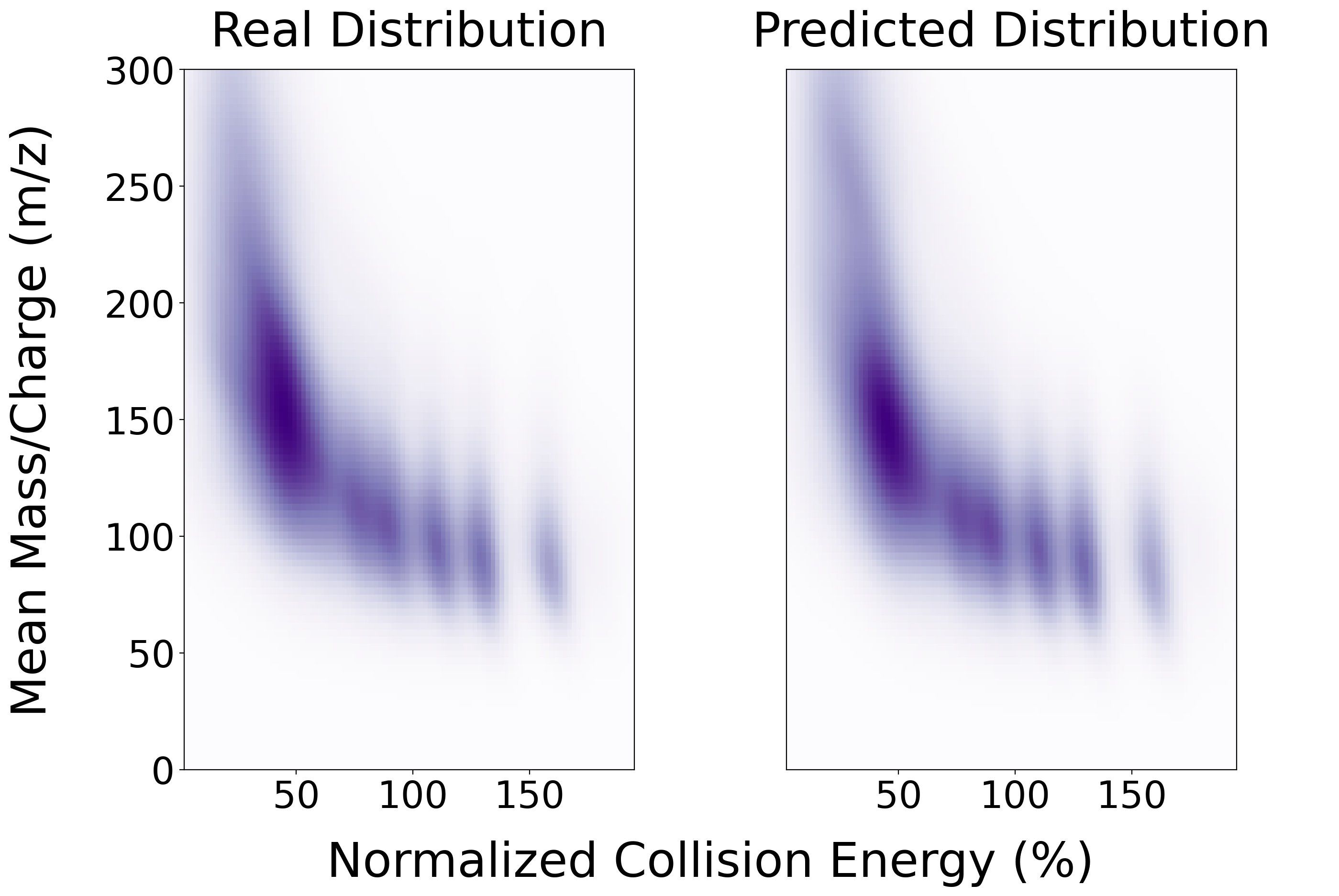}
            \caption{}
            \label{fig:ce:density}
        \end{subfigure}
        \caption{ \textbf{Collision energy experiments.} \ref{fig:ce:progression} demonstrates how increasing collision energy influences a compound's mass spectrum. Four spectra corresponding to benzhydrylpiperazine (\ref{fig:ce:molecule}, a piperazine derivative) at different normalized collision energies (NCEs) were selected from the NIST Test set. Notice how the mass distribution shifts to smaller fragments as NCE increases. MassFormer is able to predict the spectrum well at each of the four collision energies, and correctly models the shift in intensity. \ref{fig:ce:density} is a pair of density maps comparing the relationship between average mass-to-charge ratio and collision energy for real and predicted spectra from the NIST Test set. In addition to confirming the general negative correlation between collision energy and peak location, these plots demonstrate that MassFormer accurately models this property on held-out data. }
        \label{fig:ce}
    \end{figure}

    Collision energy is a key experimental parameter that can strongly influence the observed spectrum. Typically, increasing the collision energy results in more intense fragmentation, producing spectra with a higher proportion of smaller fragments. We wanted to investigate whether MassFormer could accurately model this effect. The relationship between collision energy and fragmentation is well represented in the NIST dataset, since each molecule was measured at $\approx 11$ different collision energies on average (see Table \ref{table:dset}). Figure \ref{fig:ce:progression} illustrates how collision energy  typically affects fragmentation. The four spectra (and associated spectrum predictions) all correspond to the same molecule at varying normalized collision energies (NCEs), an expression of  collision energy relative to precursor mass (see Equation \ref{eq:ace_nce}). As collision energy increases, the peak intensities in the real mass spectra shift to the left, with the model's predictions closely following this pattern. Figure \ref{fig:ce:density} shows this relationship more generally, by plotting the distribution of mean peak m/z in the spectrum, where each peak is weighted by its relative intensity. Qualitatively, the real and predicted distributions for the held-out spectra were difficult to distinguish.

    \subsection{Explaining Peak Predictions with Gradient-Based Attribution}
    \label{ss:results:explain}

    \begin{figure}[!htb]
        \begin{subfigure}{\linewidth}
            \centering
            \includegraphics[width=0.9\textwidth]{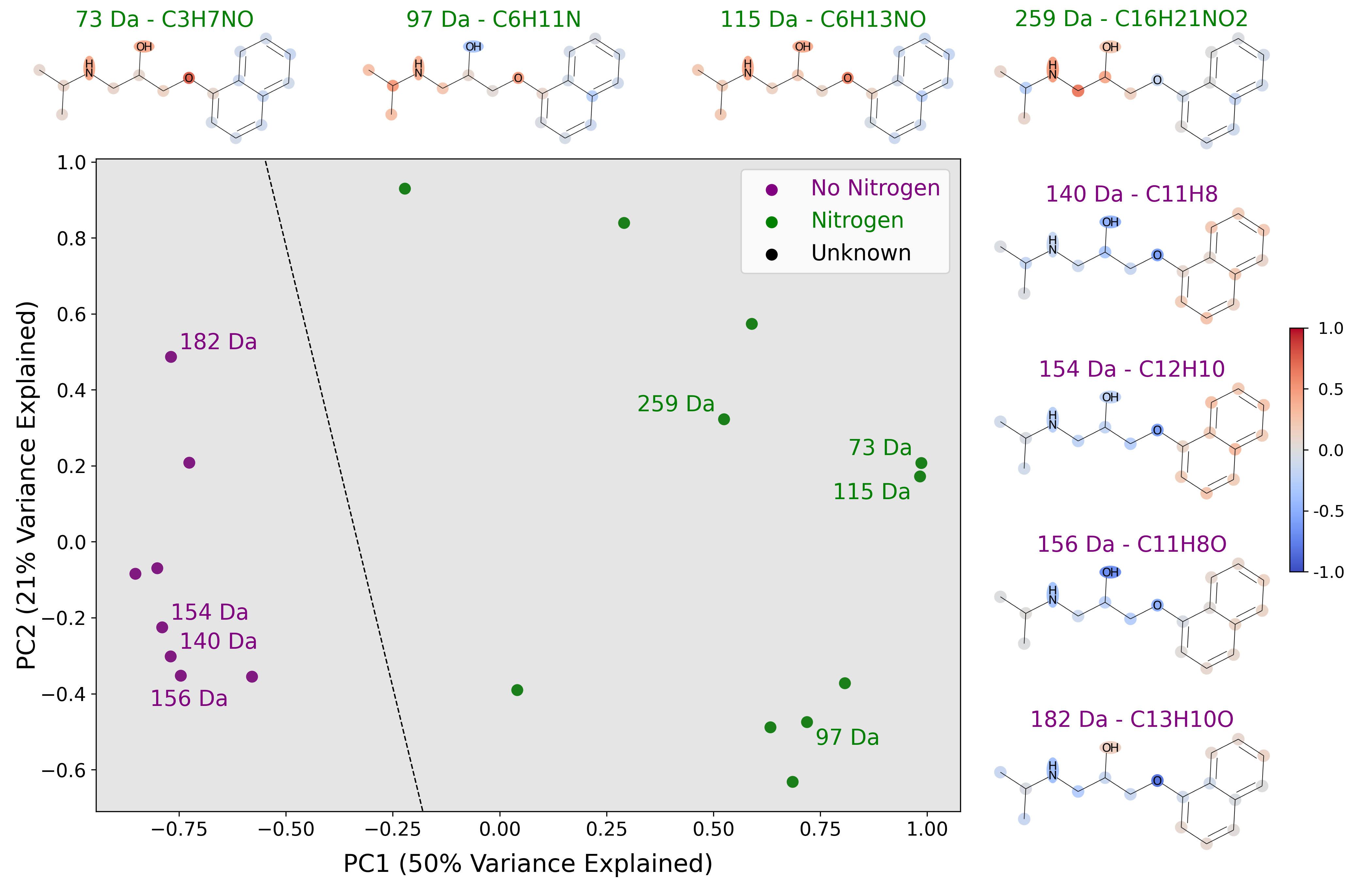}
            \caption{ }
            \label{fig:explain:spec}
        \end{subfigure}
        \bigskip
        \begin{subfigure}{0.5\linewidth}
            \centering
            \includegraphics[width=0.93\textwidth]{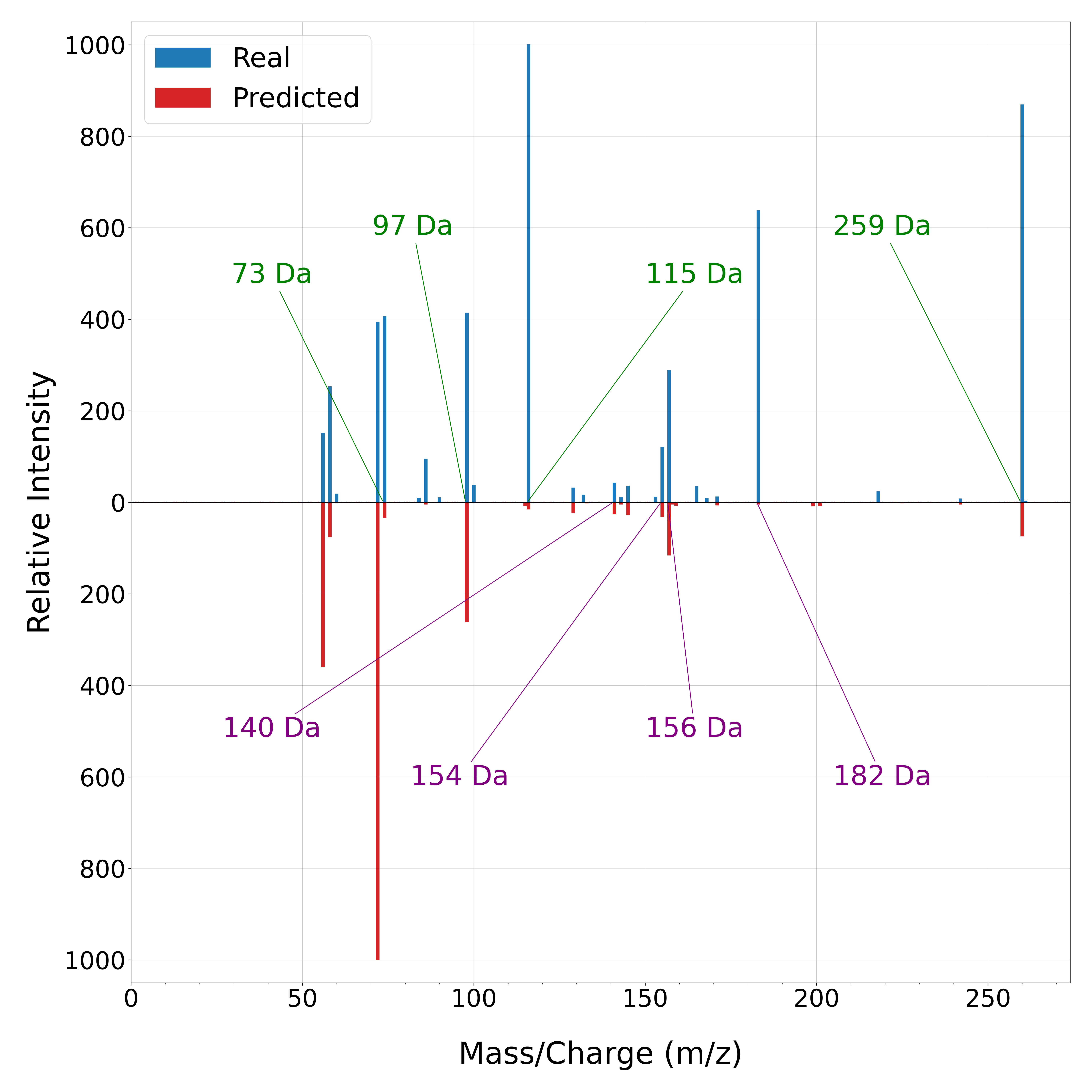}
            \caption{ }
            \label{fig:explain:pca}
        \end{subfigure}
        \begin{subfigure}{0.5\linewidth}
            \centering
            \includegraphics[width=0.93\textwidth]{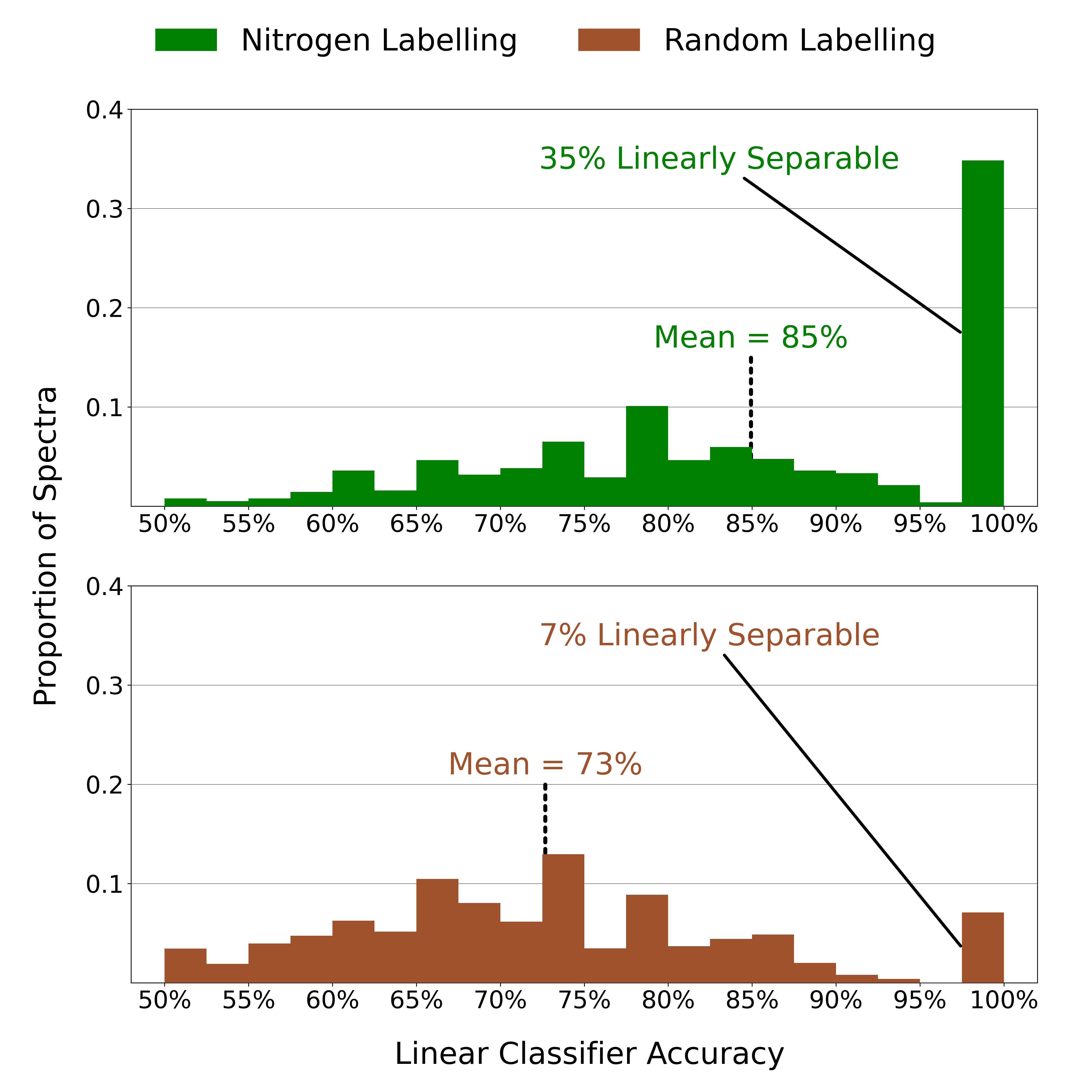}
            \caption{  }
            \label{fig:explain:scores}
        \end{subfigure}
        \caption{ \textbf{Explainability using Gradient Attributions.} GI attribution maps for each peak in Propanolol spectrum (\ref{fig:explain:spec}) were computed and then projected to two dimensions with PCA (\ref{fig:explain:pca}). The peaks were annotated with mass formulae using the SIRIUS toolkit, to allow assignment of binary labels based on heteroatom presence (in this example, Nitrogen). The projected GI maps for peaks that contained Nitrogen (in green) were linearly separable from those that did not (in purple), suggesting that these maps contained interpretable information about the peaks. The GI maps were normalized, with red values corresponding to positive attributions and blue values corresponding to negative attributions. \ref{fig:explain:scores} shows that this trend holds across the entire dataset. The distribution of optimal linear classification accuracy induced by the Nitrogen labelling strategy was markedly different from the random labelling distribution (higher accuracy indicates improved separability of the peaks). Furthermore, $35\%$ of the peaks were perfectly linearly separable when using the Nitrogen labelling, compared to only $7\%$ with the random baseline approach ($p < 10^{-48}$ by one-sided Welch's $t$-test).}
        \label{fig:explain}
    \end{figure}

    While neural networks are often viewed as black boxes, gradient-based attribution methods can provide some understanding of model behaviour. Broadly speaking, these methods work by taking the gradient of the model's output with respect to the input features to identify parts of the data that the model is most sensitive to. $\text{Gradient} \times \text{Input}$ (GI) attribution \cite{gi_1,gi_2} is one of the simplest methods for gradient-based attribution, and can be useful analyzing transformer models \cite{xai_transformer}. GI works by measuring the dot product between the input vector and its gradient: this can give information about the effects (positive or negative) of changing that feature on the model's output. For more details on GI computation refer to Section \ref{ss:methods:explain}.
    
    Each peak in a mass spectrum corresponds to a fragment of the original precursor. In organic compounds, non-Oxygen heteroatoms are much less common than the core Carbon/Hydrogen/Oxygen building blocks, so fragments that contain such heteroatoms are in some sense unique and related. We hypothesized that the model might be able to identify which peaks correspond to heteroatom-containing fragments. If this were true, one might expect the input dependencies for these peaks to be related: more concretely, the GI maps for such peaks would be correlated. A positive result would demonstrate that that, in some sense, the model understands which fragments are represented by each peak, and that such information can be revealed with gradient-based attribution methods.

\begin{figure}[!htb]
        \begin{subfigure}{0.50\linewidth}
            \centering
            \includegraphics[width=\textwidth]{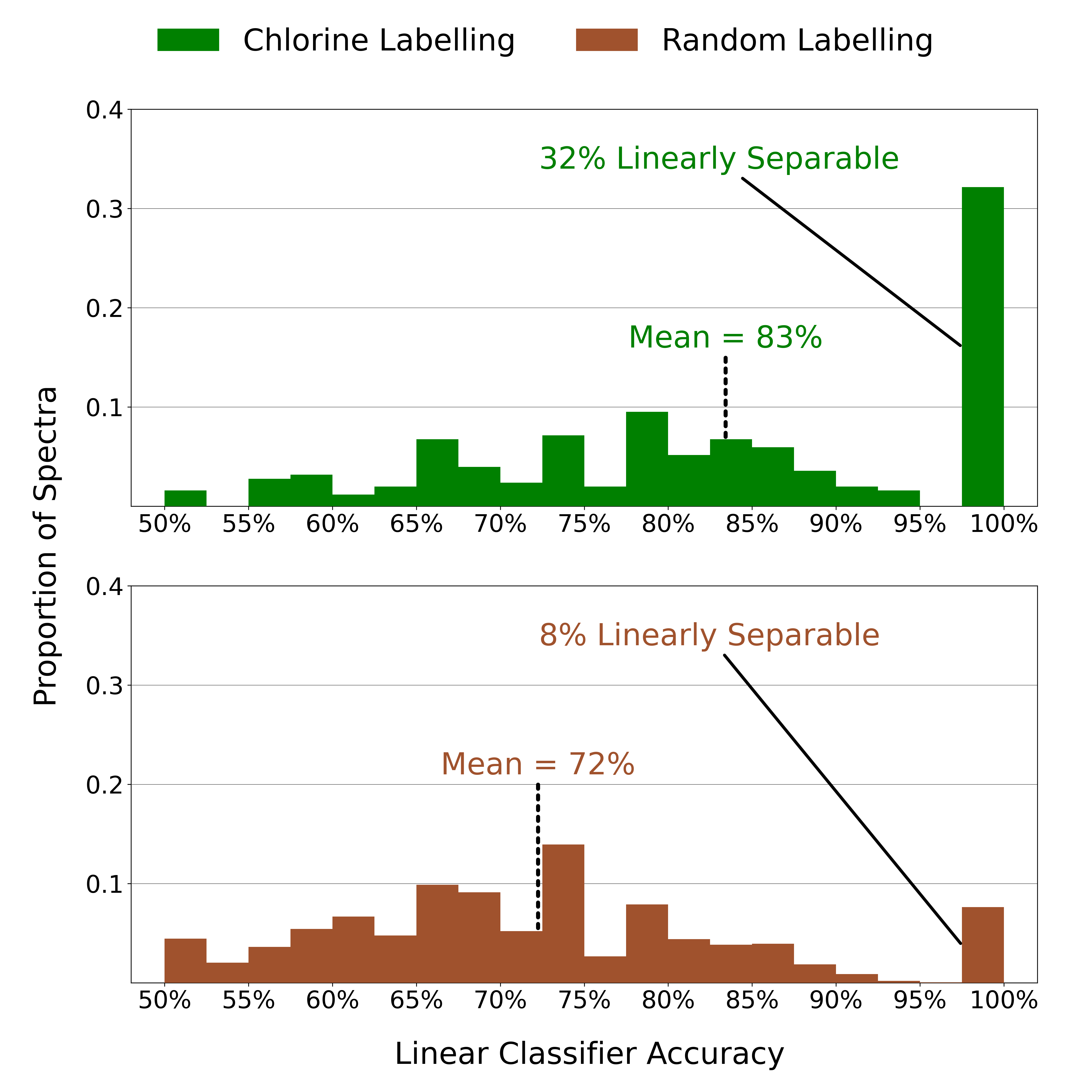}
            \caption{ Chlorine $(N = 752, p < 10^{-15})$}
            \label{supfig:explain:Cl}
        \end{subfigure}
        \begin{subfigure}{0.50\linewidth}
            \centering
            \includegraphics[width=\textwidth]{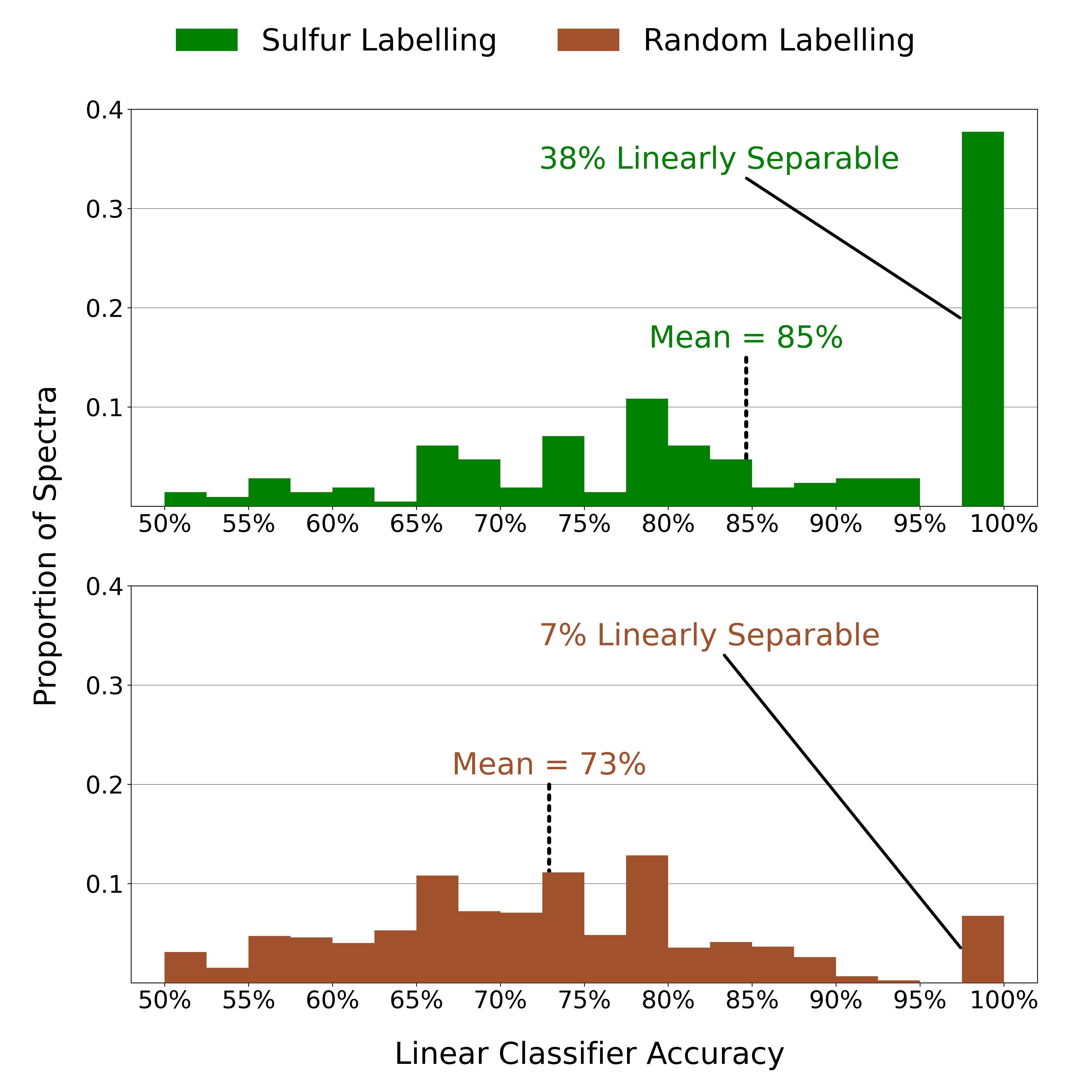}
            \caption{ Sulfur $(N = 212, p < 10^{-17})$ }
            \label{supfig:explain:S}
        \end{subfigure}
        \bigskip
        \begin{subfigure}{0.50\linewidth}
            \centering
            \includegraphics[width=\textwidth]{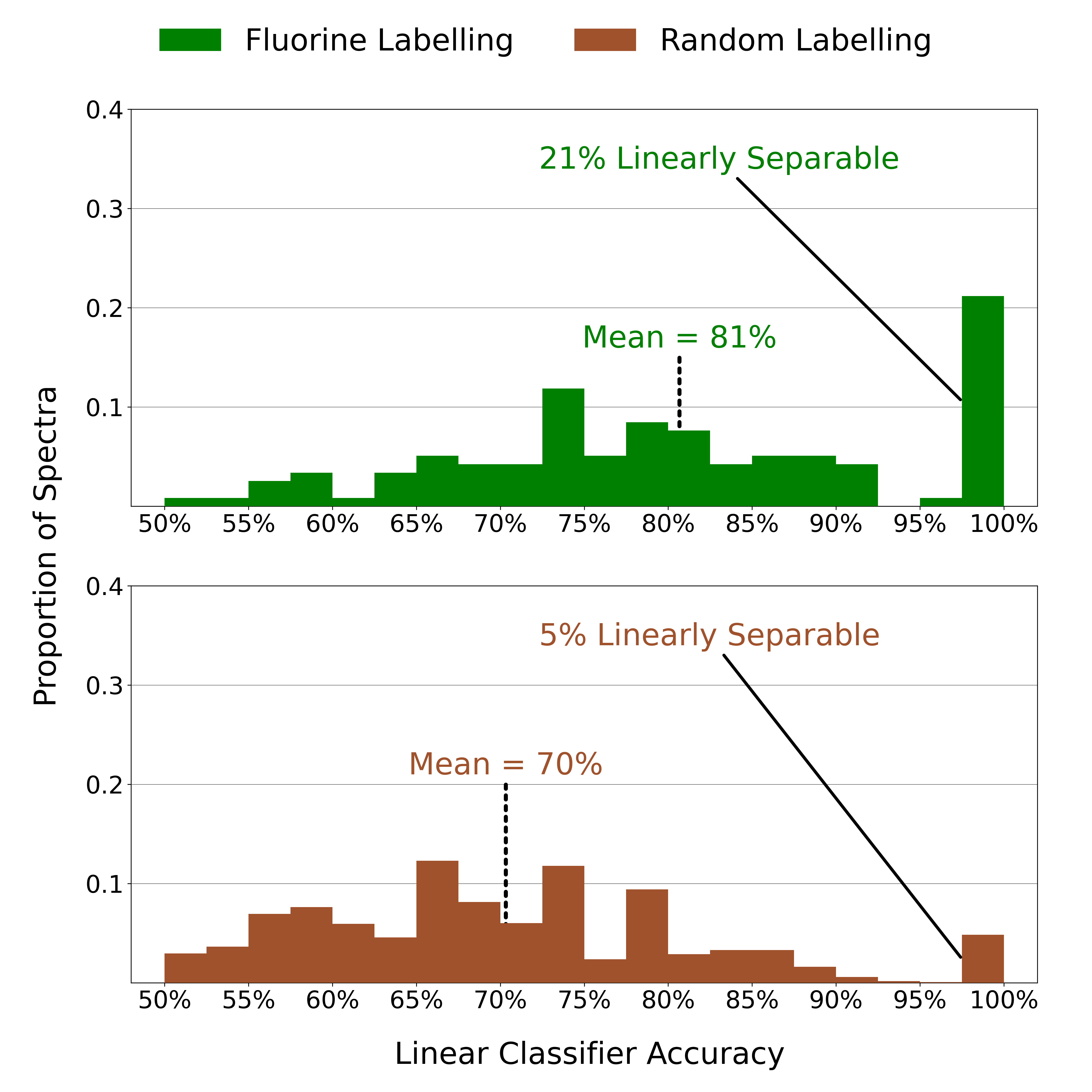}
            \caption{ Fluorine $(N = 118, p < 10^{-4})$ }
            \label{supfig:explain:F}
        \end{subfigure}
        \begin{subfigure}{0.50\linewidth}
            \centering
            \includegraphics[width=\textwidth]{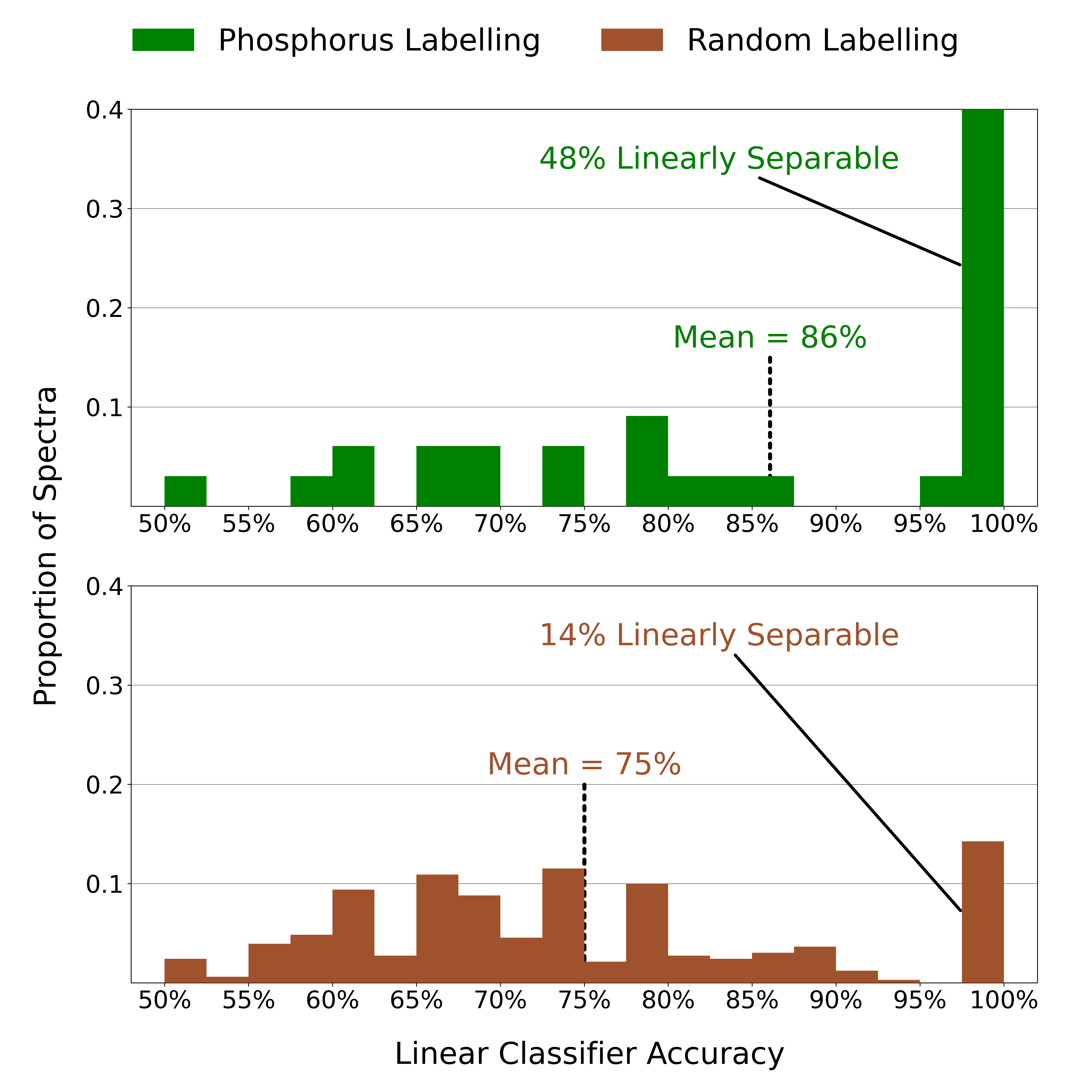}
            \caption{ Phosphorus $(N = 33, p < 10^{-3})$ }
            \label{supfig:explain:P}
        \end{subfigure}
        \caption{ \textbf{Additional Heteroatom Peak Separability Experiments.} Linear peak classification accuracy distributions, similar to Figure \ref{fig:explain:scores} but for the other non-Oxygen heteroatoms: Chlorine (\ref{supfig:explain:Cl}), Sulfur (\ref{supfig:explain:S}), Fluorine (\ref{supfig:explain:F}), Phosphorus (\ref{supfig:explain:P}). For each heteroatom, the distribution of optimal linear classification accuracy induced by the heteroatom labelling strategy was markedly different from the random labelling distribution (higher accuracy indicates improved separability of the peaks). Sample size and statistical significance (Welch's $t$-test with \v{S}id\'{a}k correction) for separability differences are provided for each plot.}
        \label{supfig:explain}
    \end{figure}

    For each peak, we computed a GI map using the gradient of the model's output at the peak location with respect to the input molecule's atom embeddings. These GI maps were high dimensional, so we projected them using principal component analysis (PCA) down to two dimensions for interpretation. Then, we applied the SIRIUS formula annotation tool \cite{sirius} to identify which peaks contained a heteroatom of interest. If a spectrum's GI map projections were linearly separable based on heteroatom labelling, this would be strong evidence that the GI maps capture heteroatom information about their corresponding peaks. Thus, we fitted a logistic regression model in the 2D projection space using the heteroatom peak labels, and reported its accuracy as a measurement of peak separability. The entire process is illustrated in Figure \ref{fig:explain}. The GI maps for a particular spectrum (Figure \ref{fig:explain:spec}) were computed, projected, and plotted in two dimensions (Figure \ref{fig:explain:pca}). In this example, it is clear that the Nitrogen and non-Nitrogen peaks are perfectly linearly separable. By repeating this analysis for many spectra from the dataset (Figure \ref{fig:explain:scores}), we showed that the Nitrogen labelling in the projected GI map space resulted in markedly higher linear separability than random labelling. In Extended Data Figure \ref{supfig:explain} we show that this pattern holds for four other heteroatoms (Cl,P,S,F). Altogether these data suggest that MassFormer can learn implicit heteroatom-based relationships between predicted peaks in the spectrum, and that these relationships can be revealed using gradient-based attribution methods.
    
    \subsection{Identifying Spectra by Ranking Candidates}
    \label{ss:results:spec_id}

    Spectrum identification is a major application that motivates the development of spectrum prediction models. MS compound identification tools (see Supplementary Notes for examples) are typically benchmarked on the Critical Assessment of Small Molecule Identification (CASMI) competition \cite{casmi2012,casmi2016}. The basic setup of the CASMI task is as follows: experimental spectra are acquired for a small set of compounds (called queries) that are not represented in existing public libraries, and competing methods are scored based on their ability to identify the correct structure. We performed a variation of the CASMI competition using MassFormer and the three comparison models from Section \ref{ss:results:spec_sim}: FP, WLN, and CFM. In this variation, we used positive mode \ce{[M}+\ce{H]+} spectra from the CASMI 2016 competition (124\footnote{There were 127 positive mode spectra in CASMI 2016, but we removed 3 compounds with charged atoms to maintain compatibility with our training setup, which assumes neutral molecules. This also allows for fairer comparison with CFM, which does not natively support charged atoms.} unique queries in total). We trained each of the three deep learning models on the same data (the NIST-InChIKey Train set, see Section \ref{ss:results:spec_sim}), with additional filtering to prevent data leakage of CASMI query compounds. As in the previous section, we used a pre-trained version of CFM, similar to what the authors did in the original CASMI 2016 competition \cite{casmi2016} as well as their most recent paper \cite{cfm4}. All models used the original candidate structure list provided by the contest organizers, which was constructed by searching ChemSpider \cite{chemspider} for compounds with masses similar to the precursor (see original paper \cite{casmi2016} for more details). Candidate rankings for each model were established based on the similarity of predicted candidate spectra with the query. The results are summarized in Tables \ref{table:rank_metrics} and \ref{table:norm_rank_metrics} (additional plots of the rank distributions can be found in Extended Data Figure \ref{supfig:id}). Overall, the deep learning models seemed to consistently outperform CFM in all metrics. MassFormer was generally superior to other methods, except for CASMI Top-1 (where it was outperformed by the Fingerprint model). While it would be possible to improve scores of all methods with a more refined candidate structure selection \cite{sirius,cfm4}, we stuck with a simple approach to demonstrate a worst-case scenario where there is little \textit{a priori} information about the query compound other than its mass and precursor adduct.

    \begin{table}[!htb]
        \begin{center}
        \resizebox{\textwidth}{!}{
        \renewcommand{\arraystretch}{1.3}
        \begin{tabular}{c|c|c|c|c|c|c}
         & \multicolumn{3}{c|}{CASMI} & \multicolumn{3}{c}{pCASMI} \\
        \hline
        Model & Rank ($\Downarrow$) & Top-1 ($\Uparrow$) & Top-5 ($\Uparrow$) & Rank ($\Downarrow$) & Top-1 ($\Uparrow$) & Top-5 ($\Uparrow$) \\
        \hline
        CFM   &   $46.0$ &  $.25$ &  $.44$ &  $391.5$ &  $.07$ &  $.23$ \\ 
        FP    &   $42.2 \pm 5.4$ &  $\bm{.34 \pm .03}$ &  $.60 \pm .02$ &  $135.9 \pm 9.5$ &  $\bm{.19 \pm .02}$ &  $.44 \pm .02$ \\
        WLN   &  $47.0 \pm 11.8$ &  $.24 \pm .04$ &  $.56 \pm .02$ &   $91.8 \pm 8.5$ &  $.16 \pm .01$ &  $.41 \pm .02$ \\
        MF    &   $\bm{27.1 \pm 5.3}$ &  $.29 \pm .02$ &  $\bm{.61 \pm .03}$ &   $\bm{85.4 \pm 5.8}$ &  $.18 \pm .02$ &  $\bm{.47 \pm .01}$ \\
        \end{tabular}
        }
        \vspace{0.5em}
        \caption{\textbf{Spectrum Identification Ranking Metrics.} Ranking metrics for the CASMI and pCASMI spectrum identification tasks. Rank corresponds to the average rank of the correct candidate. Top-$k$ accuracy reports the frequency with which the correct candidate is in set of the $k$ highest ranked candidates. MassFormer (MF) performed best  across all metrics except Top-1, where it was outperformed by the fingerprint model (FP) in both CASMI and pCASMI. Averages and standard deviations from 10 independently trained models are reported, except for CFM (which was pre-trained). \label{table:rank_metrics}}
        \end{center}
    \end{table}

    \begin{table}[!htb]
        \begin{center}
        \resizebox{\textwidth}{!}{
        \renewcommand{\arraystretch}{1.3}
        \begin{tabular}{c|c|c|c|c|c|c}
         & \multicolumn{3}{c|}{CASMI} & \multicolumn{3}{c}{pCASMI} \\
        \hline
        Model & N-Rank ($\Downarrow$) & Top-1\% ($\Uparrow$) & Top-5\% ($\Uparrow$) & N-Rank ($\Downarrow$) & Top-1\% ($\Uparrow$) & Top-5\% ($\Uparrow$) \\
        \hline
        CFM   &  $.091$ &  $.54$ &  $.68$ & $.198$ &  $.26$ &  $.44$ \\
        FP    &  $.050 \pm .005$ &  $\bm{.65 \pm .03}$ &  $.81 \pm .02$ &  $.097 \pm .006$ &  $.45 \pm 0.02$ &  $.66 \pm .02$ \\
        WLN   &  $.055 \pm .006$ &  $.58 \pm .03$ &  $.79 \pm 0.02$ &  $.073 \pm .003$ &  $.47 \pm 0.02$ &  $.70 \pm .03$ \\
        MF    &  $\bm{.036 \pm .005}$ &  $.63 \pm .03$ &  $\bm{.84 \pm 0.02}$ &  $\bm{.065 \pm .004}$ &  $\bm{.52 \pm 0.02}$ &  $\bm{.74 \pm .02}$ \\
        \end{tabular}
        }
        \end{center}
        \vspace{0.5em}
        \caption{\textbf{Spectrum Identification Normalized Ranking Metrics.} The normalized counterparts of the ranking metrics displayed in Table \ref{table:rank_metrics} for the CASMI and pCASMI spectrum identification tasks. Normalizing by the number of candidates reduces sensitivity to query spectra with large candidate sets. MassFormer (MF) seemed to perform best across all metrics, except for CASMI Top-1 where the fingerprint model (FP) was slightly better. Averages and standard deviations from 10 independently trained models are reported, except for CFM (which was pre-trained). \label{table:norm_rank_metrics}}
    \end{table}

    CASMI is a useful way of evaluating spectrum identification methods, but it has a number of limitations. Because of the small number of query spectra (124 compounds) and limited coverage of precursor adducts (only \ce{[M}+\ce{H]+} for positive mode) and collision energies (20,35,50 NCE), it is difficult to draw conclusions about the performance of models with more diverse spectrum configurations. Additionally, many of the compounds in the CASMI 2016 set have high similarity with existing compounds in modern spectral libraries. To address these concerns, we proposed a new spectrum identification task called pseudo-CASMI (pCASMI). Full details about the setup of this challenge can be found in Section \ref{ss:methods:spec_id}. Briefly, we selected outlier compounds from the NIST dataset to create the query list. The results of this challenge are also included in Tables \ref{table:rank_metrics} and \ref{table:norm_rank_metrics}. Overall, pCASMI was a more challenging task, with lower scores for all methods when compared with CASMI. MassFormer performed strongly in all metrics. Interestingly, the WLN model seemed to have problems with average rank in the CASMI competition, but was significantly better in pCASMI. Unsurprisingly, CFM performed worse in pCASMI than CASMI, likely because the query spectra represent a more diverse set of collision energies for which CFM was neither designed nor optimized.

    Altogether, these experiments demonstrate how gains in spectrum prediction will often, but not always, translate to improvements in spectrum identification. MassFormer was the superior spectrum predictor, and this was reflected in its strong ranking metrics. However, the Fingerprint model performed surprisingly well in Top-$k$, sometimes eclipsing other methods (like MassFormer and WLN) which were far better at predicting spectra. Spectrum identification requires making predictions for a large number of candidate structures, many of which are out-of-distribution with respect to the training set. Performance on an independent and identically distributed (IID) held-out test set does not fully describe a model's behaviour on out-of-distribution data, which is generally more difficult to estimate. In fact, optimized models which perform similarly on an in-distribution validation set often perform differently on out-of-distribution data \cite{id_ood_1,id_ood_2,id_ood_3}.

\section{Discussion}
\label{s:discussion}

    In this work we introduced MassFormer, a novel method for predicting MS/MS spectra for small molecules using a graph transformer architecture. We validated our model's performance with two independent MS datasets, NIST and MoNA, and showed that it can produce realistic spectra. We verified that the model captures prior knowledge about the fragmentation process by investigating the effect of collision energy on the predicted spectrum. Using gradient-based attributions, we demonstrated model explainability by showing it can identify peaks with similar element composition. We benchmarked the model with two different molecule ranking task and showed that it can be useful for MS/MS spectrum identification. Our work represents one of the first open-source deep learning spectrum predictors for MS/MS data, with extensive benchmarking and implementations of other models from the literature.

    Our method has a number of limitations. The current model is restricted to positive mode ESI Orbitrap spectra with specific precursor adducts. Extending the model to accommodate additional ionization modes (such as EI and negative mode ESI), adduct types, and instrument types (such as Q-TOF) would broaden MassFormer's impact. By supporting additional mass spectrum modalities, the model would be able to leverage new sources of MS data. Prediction resolution is another limitation: all experiments in this work use spectra binned to 1 Da, but many MS/MS spectra have much higher resolution (in NIST, up to 0.0001 Da for Orbitrap spectra). This additional resolution can help distinguish between fragments with similar masses, and is particularly useful for larger molecules. Finally, while our model is explainable to some degree (through gradient-based attribution), it does not provide true fragment peak annotations like some other methods \cite{cfm4,sirius}. Structural and formula fragment annotations are often useful for practitioners: they can improve confidence in the model's predictions by allowing experienced users to manually validate predicted peak patterns against prior knowledge about fragmentation mechanisms. Additionally, when predicting spectra for identification purposes, peak annotations can be helpful for inferring the identity of the correct compound (or compound class), even when the overall prediction contains considerable noise. Developing a deep learning model that is capable of producing these annotations would be a valuable contribution to the field.

    MassFormer has a number of exciting applications, largely focused on MS-based compound identification. We have already demonstrated how MassFormer can be used to identify a spectrum given a list of candidate structures. SIRIUS \cite{sirius} is a popular tool for identification that does not rely on spectrum prediction. It may be possible to combine MassFormer with SIRIUS (or another existing tool) to improve structure identifications. For example, SIRIUS uses information in the query spectrum to predict chemical features (such as precursor formulae and fingerprints) that help with identification. These features could be used refine a set of candidate compounds that would subsequently be ranked by predicted spectrum similarity with the query. Alternatively, forgoing a candidate database entirely, MassFormer could instead be combined with a generative model for small molecules to help search the chemical space for high-scoring spectrum matches, as seen in other works \cite{msnovelist,massgenie}.  Finally, MassFormer has potential for use in decoy generation \cite{cosmic,msms_significance}, which plays an important role in false discovery rate (FDR) calibration for untargeted metabolomics experiments. Target-decoy methods work by introducing a number of pseudo-random data points (realistic but synthetic spectra) that are ground truth negatives, allowing for empirical estimation and tuning of confidence thresholds to meet an FDR criterion. Applying MassFormer to predict noisy spectra for compounds that are unlikely to exist already in the sample (such as pesticide molecules in human blood) is a potentially powerful strategy for generating realistic decoys, which would improve FDR estimates and reduce the chance of incorrect compound identifications.

\section{Methods}
\label{s:methods}

    \subsection{Problem Formulation}
    \label{ss:methods:problem}
    
    Spectrum prediction can be viewed as a supervised learning problem, with a dataset $\{x^i,z^i,y^i\}_{i=1}^n$ where $x^i$ is a molecule and $y^i$ is its spectrum under experimental conditions $z^i$. The goal is to learn the parameters $\theta$ of the prediction function $f_\theta: \mathcal{X} \times \mathcal{Z} \rightarrow \mathcal{Y}$, where $\mathcal{X}$ is the space of chemicals and $\mathcal{Y}$ is the spectrum space. Mass spectra can be represented as a set of peaks, each of which have an m/z location and an intensity. By discretizing the peak locations into $m$ fixed-width bins (as was done in \cite{neims,gnn_msms}), a mass spectrum can be represented as an $m$-dimensional sparse vector, where each peak at location $j$ has intensity $y_j\geq0$. The problem of spectrum prediction can thus be formulated as vector regression, with $\mathcal{Y} = \mathbb{R}^m \succeq 0$. The spectral metadata $z \in \mathcal{Z}$ (such as collision energy and precursor adduct) are provided as side information to the input molecule $x$. 
    
    \subsection{Chemical Featurization}
    \label{ss:methods:feat}
    
    The featurization of an input molecule $x$ is critical, as it influences the structure of the prediction function $f_\theta$ and can have an impact on downstream performance. Molecular fingerprints (also called molecular descriptors) represent molecules using hand-designed chemical features. Common feature choices include presence of predefined substructures (MACCS, \cite{maccs}) and hashed local substructure counts (ECFP, \cite{ecfp}). Molecular graph representations capture the structure of a molecule by explicitly representing atoms as nodes and bonds as edges. The node features can encode various chemical properties associated with the atom (i.e. element, formal charge, number of bonded hydrogens), while the edge features can encode bond information (i.e. bond type, aromaticity). Such representations naturally lend themselves to graph neural networks and graph transformers (see Section \ref{ss:methods:gf} below), and can be more expressive than fingerprints. Our graph representations do not include any stereochemical, conformer, or 3D coordinate information since such information is not provided in our MS datasets.

    The node, edge, and spectrum metadata features used for MassFormer are summarized in Table \ref{table:features}. Note that the node and edge features were chosen to be identical to the pre-trained Graphormer model (see Section \ref{ss:methods:pre-train}), to preserve compatibility. However, some of these features used in this model (formal charge, radical state, stereochemical information) were not applicable to our data, and have been omitted.

    \begin{table}[!htb]
        \begin{center}
        \renewcommand{\arraystretch}{1.6}
        \begin{tabular}{c|c|c}
             Feature Type & Feature Name & Values \\
             \hline
             Node & Element & C,N,O,P,S,F,Cl \\
             Node & Degree & $0,\dots,10,11+$ \\
             Node & Number of Hydrogens & $0,\dots,8,9+$ \\
             Node & Orbital Hybridization & \makecell{SP,SP2,SP3,\\SP3D,SP3D2,Other} \\
             Node & Aromaticity & True,False \\
             Node & Ring Membership & True,False \\
             \hline
             Edge & Bond Type & \makecell{Single,Double,Triple,\\Aromatic,Other} \\ 
             Edge & Bond Conjugation & True,False \\
             \hline
             Metadata & Normalized Collision Energy & $(0,200]$ \\
             Metadata & Precursor Adduct & \makecell{\ce{[M}+\ce{H]+}, \ce{[M}+\ce{H}-\ce{H2O]+},\\ \ce{[M}+\ce{H}-\ce{2H2O]+}, \ce{[M}+\ce{2H]^2+},\\ \ce{[M}+\ce{H}-\ce{NH3]+}, \ce{[M}+\ce{Na]+}} \\
             Metadata & Precursor Mass (Da) & $(0,1000]$
        \end{tabular}
        \vspace{1em}
        \caption{ \textbf{Featurization.} Input features provided to the MassFormer model. Node features capture atom information, edge features capture bond information, and metadata features capture information specific to the spectrum. \label{table:features}}
        \end{center}
    \end{table}
    
    \subsection{Chemical Graph Transformer}
    \label{ss:methods:gf}

    Transformers \cite{transformer} are a family of neural networks characterized by their use of attention to model sequences. Originally developed for neural machine translation \cite{nmt}, transformer models have proven useful in a number of domains, from computer vision \cite{vit} to reinforcement learning \cite{rl_transformer}, achieving state of the art performance even in problems that do not naturally lend themselves to sequence modelling. They represent an input sequence as a set of embeddings, each of which capture the meaning and position of a single element. By interleaving layers of multi-head attention (MHA) with small multi-layer perceptrons (MLPs), transformers iteratively process the set of embeddings and learn relationships between elements.
    
    A number of graph transformers have been proposed \cite{memory_gnn,graphit,mat,grover,graphormer}, motivated by the ability to model pairwise global interactions between all nodes in the graph. Many graph neural networks (i.e. graph attention network \cite{gat}) are similar in structure to graph transformers but can only model local relations in a single layer, and require large depth to model interactions over longer distances \cite{how_powerful_gnns}. Graph transformers are useful when the input graph is small enough such that the quadratic memory footprint of the attention mechanism does not become prohibitively expensive. They have achieved state-of-the-art performance in both 2D (graph) and 3D molecule property prediction tasks \cite{ogb_lsc,graphormer}.
    
    Our approach adapts the Graphormer \cite{graphormer} architecture, a recent graph transformer model that boasts impressive results on chemical property prediction tasks \cite{ogb,graphormer}, particularly in the low-data regime (on the order of $10^4$ samples). The distinguishing characteristic of this graph transformer is its unique positional encoding scheme. The model uses shortest path information between nodes, and associated edge embeddings along that path, as a form of relative positional encoding. The shortest path information is computed as a preprocessing step for each graph, using the Floyd-Warshall algorithm \cite{floyd}. 
    
    The attention mechanism $a_{ij}$ is described in detail in Equations \ref{eq:attn_1} and \ref{eq:attn_2}, where $h_i,h_j \in \mathbb{R}^{d' \times 1}$ are representations for nodes $i$ and $j$ respectively. $W_{K},W_{Q} \in \mathbb{R}^{d \times d'}$ are the standard learnable key and query projection matrices. $b_{ij} \in \mathbb{R}$ is a learnable scalar indexed by the shortest path distance between $i$ and $j$ (which is always a positive integer). $c_{ij} \in \mathbb{R}$ is the edge embedding term, described by Equation \ref{eq:attn_2}. Allowing a slight abuse of notation, $e_p \in \mathbb{R}^{d \times 1}$ is the embedding corresponding to the $p^{th}$ edge in the shortest path between $i$ and $j$, and $w_p \in \mathbb{R}^{d \times 1}$ is a learnable weight for that position. 
    
    \begin{align}
        a_{ij} &= \texttt{softmax} \left( \frac{(W_Q h_i)^T(W_K h_j)}{\sqrt{d}} + b_{ij} + c_{ij} \right) \label{eq:attn_1} \\
        c_{ij} &= \frac{1}{N} \sum_{p} w_{p}^T e_p \label{eq:attn_2}
    \end{align}
    
    For graph-level prediction tasks, it is useful to add a readout node to the input graph that can extract graph-level embeddings, similar to the CLS token in NLP transformers \cite{bert}. This ``fake" node is initialized with a unique embedding and connected to all other nodes in the graph with a special edge type. In the final layer of the transformer, the readout node's embedding is interpreted as a summarized representation of the input graph and can be used for downstream property prediction.
    
    \subsection{Pre-training and Fine-tuning}
    \label{ss:methods:pre-train}

    Adapting a pre-trained model can offer improved performance over training a radomly initialized model from scratch, particularly when data scarcity is a concern. We initialized the parameters of our graph transformer module and the learned input node and edge embeddings (see Figure \ref{fig:overview}) with the corresponding parameters of a pre-trained Graphormer model. This model was originally trained on the PCQM4Mv2 dataset \cite{ogb_lsc,pcq}, a large dataset of approximately 4 million 2D molecular graphs and their Density Functional Theory (DFT) simulated HOMO-LUMO energy gap. The pre-training task was a supervised graph-level regression problem of predicting the energy gap. While this task is not directly related to mass spectrometry, the roughly 100-fold larger compound coverage of the PCQM4Mv2 dataset provided an opportunity for the model to learn general chemical representations that transfer to the spectrometry task. In Extended Data Table \ref{suptable:ablations}, we performed model ablations to determine relative contributions of different aspects of the fine-tuning process. Our experiments demonstrate that using pre-trained weights is necessary to scale up the number of model parameters while maintaining training stability and performance. We also found re-initializing the statistics of the layer normalization \cite{layernorm} modules to be helpful. The other key component of MassFormer, the spectrum prediction MLP, was initialized randomly. Both modules were fine-tuned jointly for 20 epochs using a linearly decaying learning rate. For full details on the training procedure, please refer to the code repository (Section \ref{s:code_availability}).

    \subsection{Loss and Similarity Calculations}
    \label{ss:methods:loss}

    Since MS peak intensities are relative \footnote{The unnormalized value of a peak intensity can provide useful information in some contexts (such as quantification and noise removal), but reference spectra have already been normalized and denoised.}, it is advisable to use loss functions that are invariant to scaling. We choose cosine distance (Eq. \ref{eq:cd_loss}) as the loss function, where $\hat{y}=f_\theta(x,z)$ is the predicted spectrum and $y$ is the real spectrum. Cosine similarity is commonly used to compare spectra, so minimizing the cosine distance (thus maximizing similarity) is a natural choice and has been shown to work well in other models \cite{neims,gnn_msms}. 
    
    \begin{align}
        CD(y,\hat{y}) &= 1 - \frac{y^T \hat{y}}{||y||_2 ||\hat{y}||_2} = 1 - \frac{\sum_{i=1}^m y_i \hat{y}_i}{\sqrt{\sum_{j=1}^m {y_j^2} \sum_{k=1}^m {\hat{y}_k^2}}}
        \label{eq:cd_loss}
    \end{align}

\begin{figure}[!htb]
        \begin{subfigure}{\linewidth}
            \centering
            \includegraphics[width=0.8\textwidth]{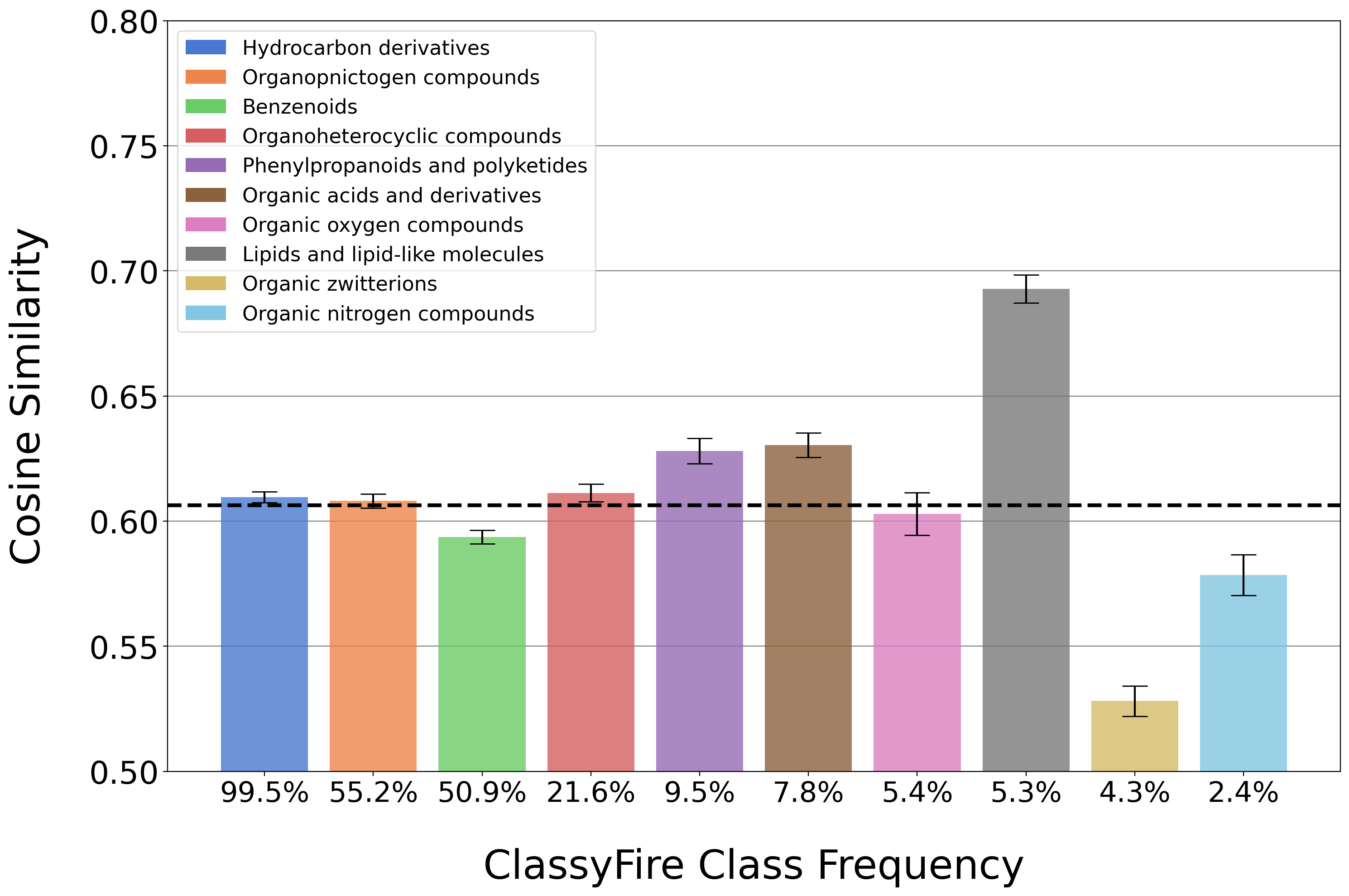}
            \caption{MassFormer}
            \label{supfig:classyfire:mf}
        \end{subfigure}
        \bigskip
        \begin{subfigure}{\linewidth}
            \centering
            \includegraphics[width=0.8\textwidth]{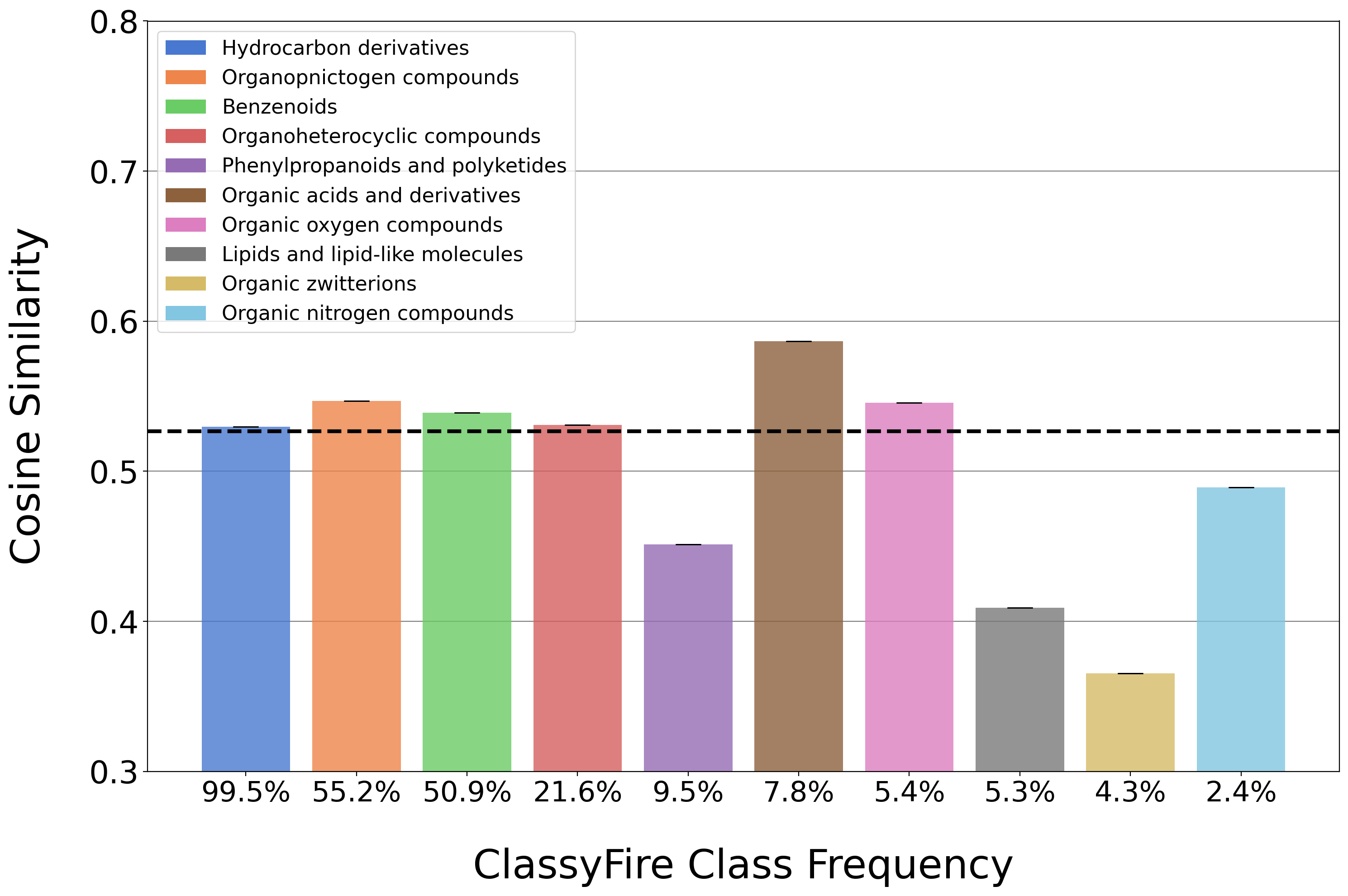}
            \caption{CFM}
            \label{supfig:classyfire:cfm}
        \end{subfigure}
        \caption{ \textbf{ClassyFire Similarity Experiments.} Performance on the top 10 most frequent chemical classes, as identified by ClassyFire, in the NIST-Scaffold Test Set (\ce{[M}+\ce{H]+} adducts only). Scores reported for both MassFormer (\ref{supfig:classyfire:mf}) and CFM (\ref{supfig:classyfire:cfm}). The chemical classes are sorted from most to least frequent on the x-axis, and are not disjoint. Note the difference in y-axis range between the two plots. The average performance for each model (across all compounds) is indicated by a black dashed line. MassFormer had superior performance to CFM in each category. Strikingly, MassFormer seemed to perform best on ``lipids and lipid-like molecules", which is one class that CFM seems to struggle with. This could be explained by the fact that many lipids are known to have simple fragmentation rules, which might make them easier to learn. However, since lipids can be quite large, any algorithm which relies on combinatorial fragmentation (like CFM) would struggle. Averages and standard deviations from 10 independently trained models are reported (except for CFM, which was pre-trained).}
        \label{supfig:classyfire}
    \end{figure}

\begin{figure}[!htb]
        \begin{subfigure}{\linewidth}
            \centering
            \includegraphics[width=0.8\textwidth]{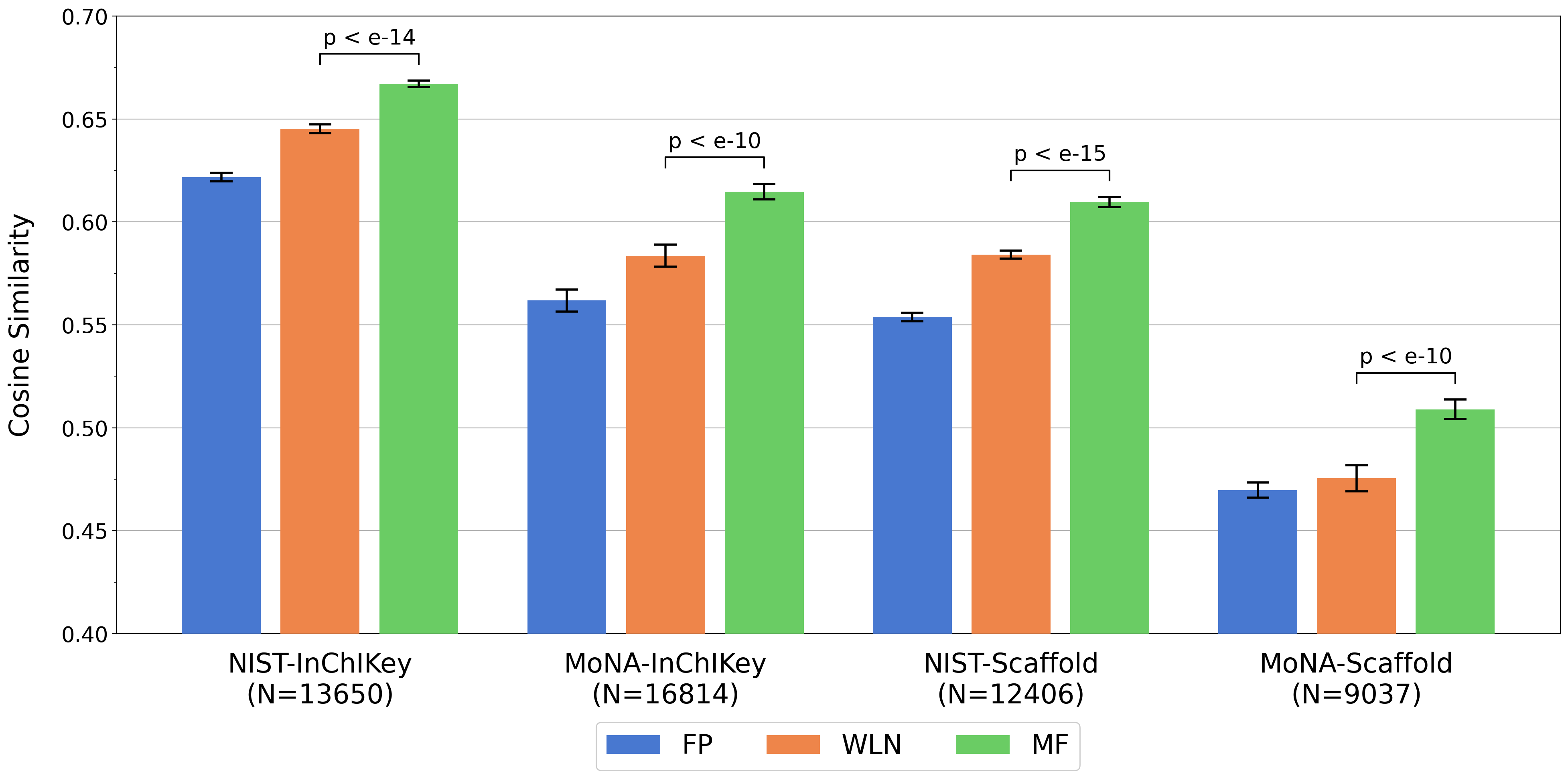}
            \caption{\ce{[M}+\ce{H]+} Adduct}
            \label{supfig:dl_sims:mh}
        \end{subfigure}
        \bigskip
        \begin{subfigure}{\linewidth}
            \centering
            \includegraphics[width=0.8\textwidth]{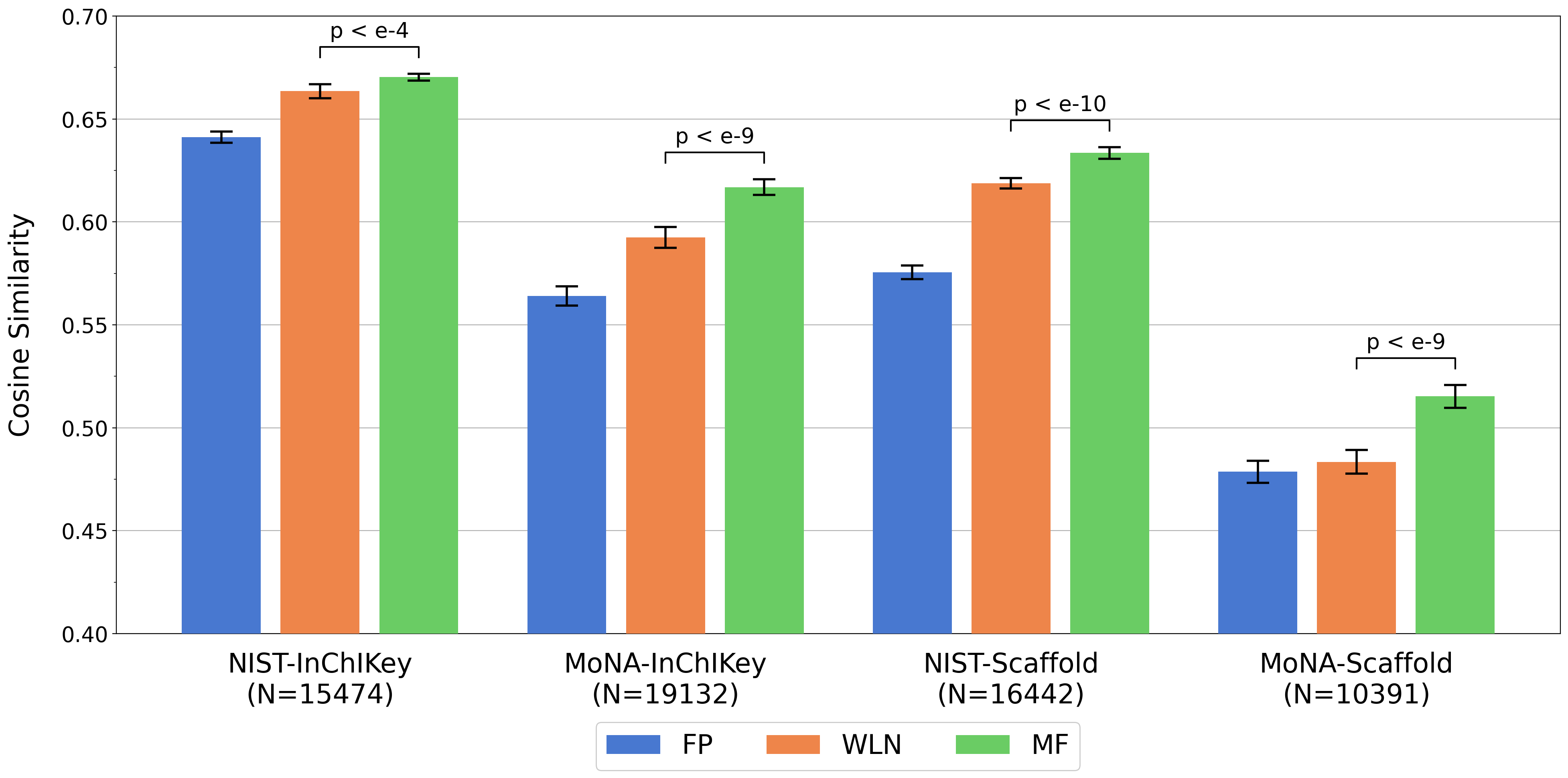}
            \caption{All Adducts}
            \label{supfig:dl_sims:main}
        \end{subfigure}
        \caption{ \textbf{Additional Spectrum Similarity Experiments.} These experiments were similar to Figure \ref{fig:sim:bars}, but did not involve filtering compounds from the test set based on overlap with CFM's training set. This allowed for larger test set sizes and more rigorous comparisons between deep learning approaches. In \ref{supfig:dl_sims:mh}, all models were trained and evaluated only on \ce{[M}+\ce{H]+} spectra, but in \ref{supfig:dl_sims:main} all six supported precursor adducts were included. MassFormer boasted strong performance in both cases. Training set sizes are indicated for each split. Averages and standard deviations from 10 independently trained models are reported. Statistical significance determined by one-sided Welch's $t$-test with \v{S}id\'{a}k correction.}
        \label{supfig:dl_sims}
    \end{figure}

    The spectral similarity experiments (Figure \ref{fig:sim}, Extended Data Figures \ref{supfig:classyfire} and \ref{supfig:dl_sims}, Extended Data Table \ref{suptable:ablations}) report average cosine similarity using a particular method of aggregating the results across spectra. Since there were often multiple spectra with different collision energies corresponding to the same precursor, both target and predicted spectra were averaged (in binned spectrum space) across collision energy before computing similarity. This helped prevent inflated similarity scores resulting from spectra with very few peaks, which tends to happen when the collision energy is either too high or too low. These similarity scores were subsequently averaged per molecule (i.e. across precursor adducts), then averaged again across molecules. This effectively down-weighted the individual importance of spectra corresponding to molecules with multiple precursor adducts in the dataset.
    
    \subsection{Gradient-based Feature Attribution}
    \label{ss:methods:explain}

    $\text{Gradient} \times \text{Input}$ (GI, \cite{gi_1,gi_2}) is an attribution method that assigns importance scores to input variables based on the sensitivity of the model's predictions to changes in those variables, estimated using gradients. It satisfies the conservation axiom, an underlying assumption for many explainable AI approaches which posits that ``scores assigned to input variables and forming the explanation must sum to the output of the network" \cite{xai_transformer}. If a model is sensitive to changes in certain parts of the input, these features are taken to be more important for making a correct prediction. GI methods compute the attribution score as the the dot product of the gradient vector with the input. Inputs with scores close to 0 are unimportant, while those with large positive or negative scores are interpreted as contributing positively or negatively (respectively) to the prediction.

    More formally, let $x \in \mathbb{R}^D$ be an input vector with $D$ dimensions, let $y \in \mathbb{R}^K$ be an output value associated with $x$, and $f_{\theta}(x): \mathbb{R}^D \rightarrow \mathbb{R}^K$ be a neural network. Let $\mathcal{L}(y,\hat{y}): \mathbb{R}^K \times \mathbb{R}^K \rightarrow \mathbb{R}$ be a scalar loss function (for example, cosine distance). The GI score for the model on this input vector is defined precisely by Equation \ref{eq:gi}:
    
    \begin{align}
        \text{GI}(x,y) &= x \cdot \nabla_x \mathcal{L}(y,f_{\theta}(x)) \label{eq:gi}
    \end{align}

    While GI score calculation is generally applicable to any type of fully differentiable neural network, there are certain aspects of the transformer architecture (the self-attention mechanism \cite{transformer} and layer normalization module \cite{layernorm}) that violate the conservation axiom and, in practice, reduce the quality of model explanations \cite{xai_transformer}. To address this problem, we made slight modifications to these modules when calculating GI scores, as recommended in \cite{xai_transformer} (refer to their paper for full details).

    In our experiments, the neural network $f_{\theta}$ in Equation \ref{eq:gi} was MassFormer, which maps input molecules to $K$-dimensional binned spectrum vectors (where $K = 1000$). To compute the GI scores for a peak at particular location $k$, we used a loss function that zeros out all other peaks in the spectrum. More precisely, we defined the loss $\mathcal{L}_k$ for peak $k$ using Equation \ref{eq:peak_loss}:
    
    \begin{align}
        \mathcal{L}_k(x) = e_k \cdot f_{\theta}(x)
        \label{eq:peak_loss}
    \end{align}
    
    In the above formulation, $e_k$ is the $k^{th}$ standard basis vector for $\mathbb{R}^K$ (in other words, a one-hot vector where the $k^{th}$ entry is 1). 
    
    Note that MassFormer takes a molecular graph as input, which is subsequently preprocessed into different types of embeddings (see Figure \ref{fig:overview} and Section \ref{ss:methods:feat}). We computed attribution scores only with respect to the element embeddings, as our experiments in Section \ref{ss:results:explain} involved discriminating peaks by element composition. To visualize the GI attribution maps (i.e. Figure \ref{fig:explain:pca}), we estimated GI scores for each atom and then $L_2$ normalized across atoms, to remove information about gradient magnitude and focus only on direction. We employed a slight variation of the aforementioned GI computation (Equation \ref{eq:gi}) for the linear projections in Section \ref{ss:results:explain}: instead of summing over the embedding dimension to produce a single scalar score per atom, we performed PCA on the unreduced GI map. Summing over the dimensions is useful for producing an easily interpretable score, but might remove variation that could inform the PCA projection.
    
    \subsection{Datasets and Training Splits}
    \label{ss:methods:dset}
    
    We used the NIST 2020 MS/MS dataset \cite{nist2} for both training and evaluation. NIST is a commercial dataset notable for its large coverage (over 1 million tandem spectra in total), standardized spectrum acquisition protocol, and high degree of manual validation. We also used the publicly available MassBank of North America (MoNA) as a held-out evaluation set. Massbank is an important dataset for the MS community, as it is one of the largest open online repository of MS data for small molecules. This dataset contained spectra from a variety of other online repositories, including GNPS \cite{gnps}, HMDB \cite{hmdb}, and ReSpect \cite{respect}. For simplicity, we only considered spectra from Orbitrap instruments that use HCD collision \cite{ms_textbook}, as this corresponds to the largest subset of data in NIST. Furthermore, we restricted the dataset to only include positive mode spectra with 6 of the most highly occurring precursor adducts (\ce{[M}+\ce{H]+}, \ce{[M}+\ce{H}-\ce{H2O]+}, \ce{[M}+\ce{H}-\ce{2H2O]+}, \ce{[M}+\ce{2H]^2+}, \ce{[M}+\ce{H}-\ce{NH3]+}, \ce{[M}+\ce{Na]+}). The dataset statistics are summarized in Table \ref{table:dset}. After filtering, NIST is a much larger dataset than MoNA, both in terms of number of spectra and compounds. For this reason, we relied on NIST data to train the models and used MoNA exclusively for evaluation.

    \begin{table}[!htb]
        \begin{center}
        \renewcommand{\arraystretch}{1.3}
        \begin{tabular}{c|c|c|c}
        Dataset & \# Spectra & \# Compounds & Average \# NCE / Compound\\
        \hline
        NIST & 375,406 & 22,105 & 11.33 \\
        MoNA & 13,225 & 1,376 & 9.29 \\
        \end{tabular}
        \end{center}
        \vspace{1em}
        \caption{Summary statistics for each MS dataset, after filtering for positive mode Orbitrap spectra and desired precursor adducts (\ce{[M}+\ce{H]+}, \ce{[M}+\ce{H}-\ce{H2O]+}, \ce{[M}+\ce{H}-\ce{2H2O]+}, \ce{[M}+\ce{2H]^2+}, \ce{[M}+\ce{H}-\ce{NH3]+}, \ce{[M}+\ce{Na]+}). \label{table:dset}}
    \end{table}

    Two kinds of data splitting techniques were employed in Section \ref{ss:results:spec_sim}. Both methods involved splitting spectra based on compound identity, to avoid leakage of spectra that differ only in metadata (such as collision energy or precursor type) but not in structure. The ``InChIKey" split used non-sterochemical InChIKey strings \cite{inchi}, which are hashed chemical string representations, as molecular identifiers. This approach was, in essence, a simple random split based on compound identity. In contrast, the ``Scaffold" split used Murcko Scaffolds \cite{scaffold} to coarsely cluster compounds before splitting in a manner such that all the compounds (and associated spectra) from one cluster would end up in the same partition. Scaffold splitting introduces distributional shift between training and test data, and is commonly used to evaluate deep learning models in small molecule applications \cite{moleculenet}.

\subsection{Baseline Models}
    \label{ss:methods:baseline_models}
    
    We compare our method with two related deep learning models, based on existing approaches from the literature. The Fingerprint model combines ECFP, MACCS, and RDKit fingerprints (all of which are available from the RDKit library \cite{rdkit}) and uses those directly as the chemical embedding. The WLN model uses a molecular graph representation in combination with a Weisfeiler-Lehman Network (\cite{wln}), which is a particular kind of graph neural network, to produce the chemical embedding. These models are based on two previously published MS prediction models: \cite{neims} and \cite{gnn_msms} respectively. However, both models required reimplementation for direct comparison: the former \cite{neims} was trained on a different type of MS data (EI-MS), while the latter \cite{gnn_msms} did not have available public code. \footnote{We communicated with the authors of \cite{gnn_msms} and designed our WLN model to be similar to their best configuration.} Compared to MassFormer's 56 million parameters, the WLN model had approximately 6 million parameters and the FP model had approximately 35 million parameters.

    \begin{table}[!htb]
        \begin{center}
        \renewcommand{\arraystretch}{1.3}
        \begin{tabular}{c|c|c}
        Dataset & InChIKey Overlap/Total & Scaffold Overlap/Total \\
        \hline
        NIST & 2,728/22,114 & 1,372/6,893 \\
        MoNA & 499/1,372 & 362/649 \\
        CASMI (Queries) & 48/124 & 58/92 \\
        CASMI (Candidates) & 139/197,465 & 580/71,477 \\
        pCASMI (Queries) & 32/300 & 37/300 \\
        pCASMI (Candidates) & 114/650,225 & 820/192,980 \\
        \end{tabular}
        \end{center}
        \vspace{1em}
        \caption{Structure overlap of the CFM training set (4054 molecules, 1576 scaffolds) with other datasets used in the experiments. Overlap reported in terms of identical InChIKeys (ignoring stereochemical information) and identical Scaffolds. \label{table:cfm_overlap}}
    \end{table}
    
    In addition, we compare our model to Competitive Fragmentation Modelling (CFM, \cite{cfm,cfm4}), a well-known tool for small molecule spectrum prediction. CFM is not a deep learning approach: it uses combinatorial fragmentation to determine the set of possible peak locations, then fits a probabilistic model to predict the relative intensity of the fragments. CFM was designed to predict Q-TOF spectra at 3 specific collision energies (10,20,40) and with a limited set of precursor adducts (for positive mode, only \ce{[M}+\ce{H]+}). It is not straightforward to re-train CFM on the NIST dataset, which contains roughly 10x more spectra than CFM's original training set, from a different kind of instrument (Orbitrap) and covering a wider range of precursor adducts and collision energies. CFM makes assumptions about fragmentation that are specific to the types of spectra that it models, complicating application to new datasets. Additionally, scaling CFM's training procedure to such a large dataset would be challenging. For these reasons, we used the most recent pre-trained version of the model \cite{cfm4} for all experiments. This version was trained on a 4054-molecule subset of the Metlin dataset \cite{metlin}: see Table \ref{table:cfm_overlap} for information about the overlap of CFM's training set with the other datasets. In our experiments with CFM, we mapped the input normalized collision energy to whichever of the three CFM-supported absolute collision energies was closest. To convert from normalized to absolute collision energy, we used Equation \ref{eq:ace_nce}, where $m(p)$ is the precursor mass and $c(p)$ is the charge factor, which is 1.0 for singly charged precursors and 0.9 for doubly charged precursors.

    \begin{align}
        \text{ACE} = \frac{m(p) \times c(p) \times \text{NCE}}{500}
        \label{eq:ace_nce}
    \end{align}
    
    CFM also contains a rule-based module for lipid spectrum prediction \cite{cfm3}. We preferentially employed this module to make predictions instead of the combinatorial/probabilistic approach, wherever applicable.

    \subsection{Spectrum Identification Task Setup}
    \label{ss:methods:spec_id}

    CASMI 2016 contains Orbitrap spectra for 188 unique compounds, 127 of which are positive mode (\ce{[M}+\ce{H]+} adducts). Each spectrum was merged over three normalized collision energies: 20, 35, and 50. Additionally, each query had an associated candidate list, which was originally developed by searching the ChemSpider database \cite{chemspider} for compounds with similar masses to the precursor. After preprocessing, there were on average 1250 candidate compounds per spectrum.
    
    To construct the pCASMI dataset, we selected compounds from the NIST dataset with unique Murcko Scaffolds. We sampled 400 of such compounds in total, stratified by molecular weight: 75 with weight $< 200$ Da, 75 with weight in $[200, 300)$ Da, 75 with weight in $[300, 400)$ Da, and 75 with weight $\geq 400$ Da. The motivation was to select a diverse group of outlier compounds to use as queries in the identification task. For each compound, all \ce{[M}+\ce{H]+} spectra (of any collision energy) were used as query information. The candidate sets were established by sampling compounds from PubChem (up to 10k) with molecular mass within 0.5 ppm of the true query mass. We filtered the candidate list to remove rare elements and multi-molecular compounds. Stereochemical information was also removed, and stereoisomeric candidates were deduplicated. After preprocessing, there were on average 2201 candidates per spectrum, which is roughly double that of the CASMI challenge.

\begin{figure}[!htb]
        \begin{subfigure}{\linewidth}
            \centering
            \includegraphics[width=0.9\textwidth]{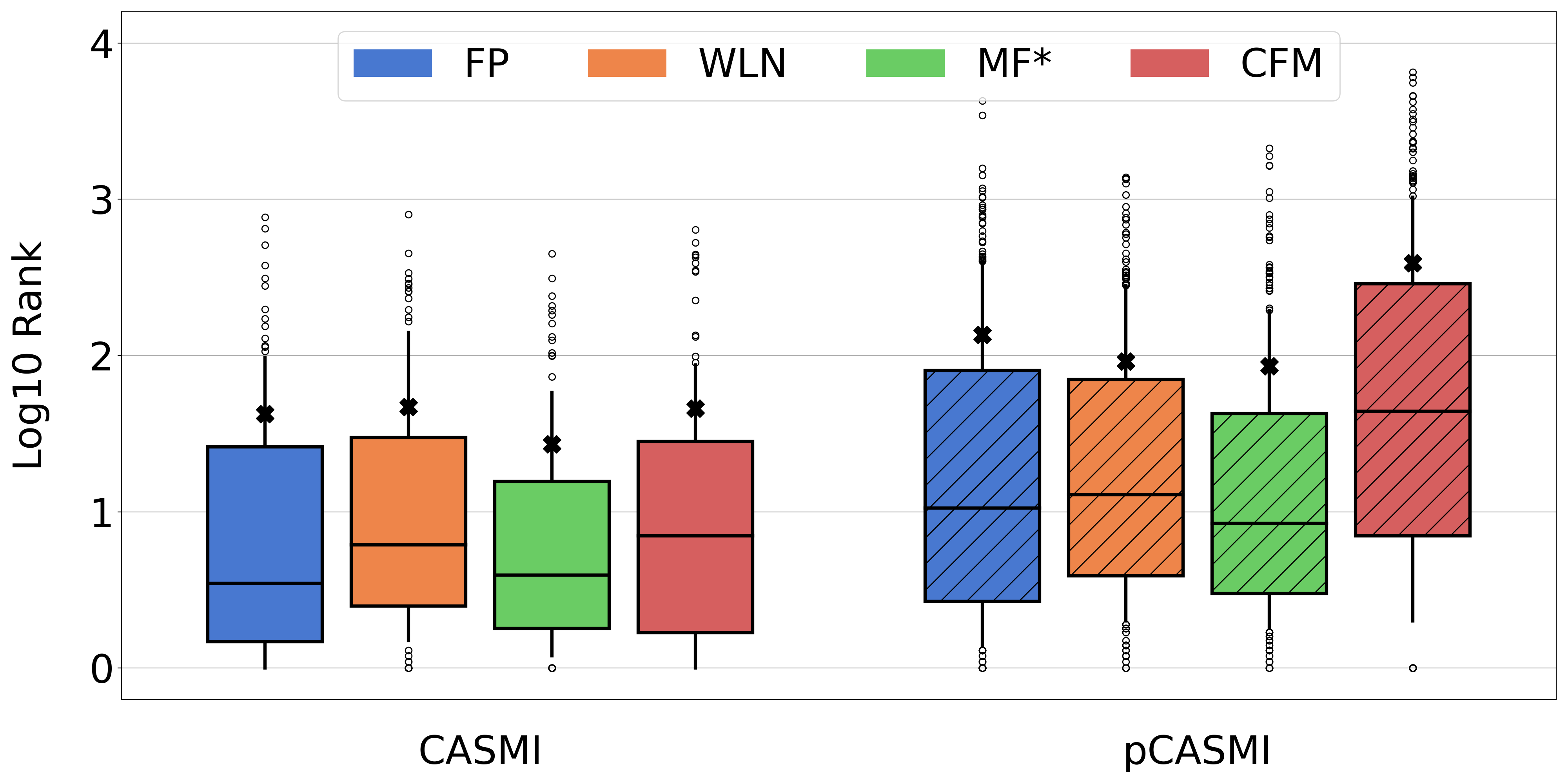}
            \caption{Log Rank Distribution}
            \label{supfig:id:both_rank_dist}
        \end{subfigure}
        \begin{subfigure}{\linewidth}
            \centering
            \includegraphics[width=0.9\textwidth]{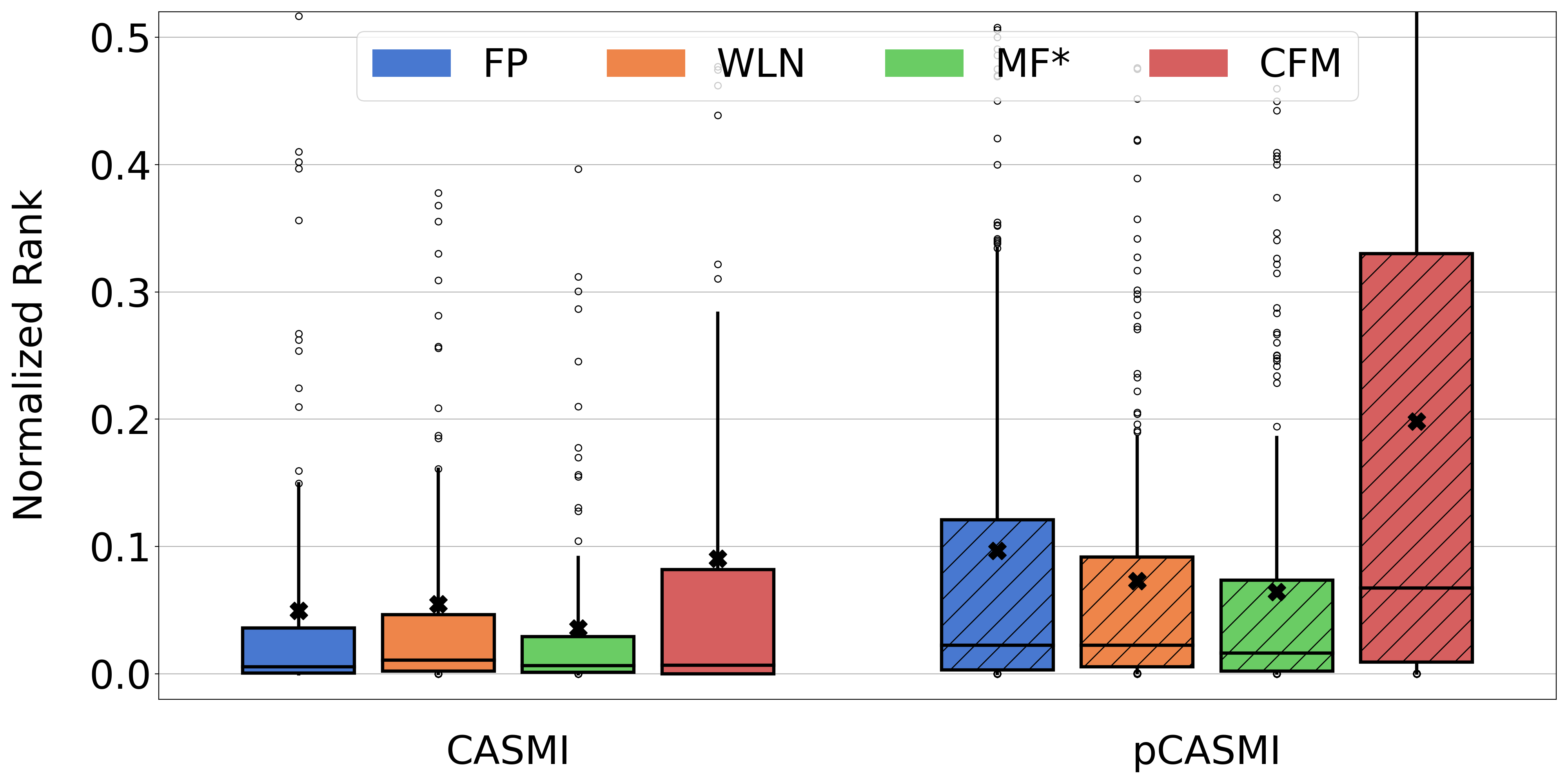}
            \caption{Log Normalized Rank Distribution}
            \label{supfig:id:both_norm_rank_dist}
        \end{subfigure}
        \vspace{0.1em}
        \caption{ \textbf{Spectrum Identification Ranking Distributions.} The unnormalized (\ref{supfig:id:both_rank_dist}) and normalized (\ref{supfig:id:both_norm_rank_dist}) distributions of the true candidate rank, for CASMI and pCASMI queries. Note that for rank metrics, a lower score is better. MassFormer's rank distributions are more strongly skewed towards lower values, with fewer extreme outliers. Boxplot lines represent median and interquartile range, whiskers represent 1.5 times the interquartile range, and the ``X" symbol represents the mean.}
        \label{supfig:id}
    \end{figure}

    The models were scored using an array of different unnormalized (Table \ref{table:rank_metrics}) and normalized (\ref{supfig:id}, Table \ref{table:norm_rank_metrics}) ranking metrics, inspired by those typically employed in CASMI competitions \cite{casmi2016}. Average Rank (Table \ref{table:rank_metrics}) corresponds to the average rank of the true candidate compound, with 1 being the best score. Its normalized counterpart, Average Normalized Rank (Table \ref{table:norm_rank_metrics}) corresponds to the average rank of the true candidate expressed as a fraction of the total number of candidates, with 0 being the best score and 1 being the worst. Top-$k$ Accuracy (Table \ref{table:rank_metrics}) represents the frequency with which the true candidate is ranked in the top $k$ candidates, and ranges from 0 to 1. Top-$k\%$ Accuracy (Table \ref{table:norm_rank_metrics}) is the normalized equivalent, measuring how often the correct candidate appeared in the top $k\%$ of candidates. In contrast to the average rank metrics, Top-$k$ and Top-$k\%$ metrics do not strongly penalize ranking the correct candidate extremely poorly (any rank outside of the top $k$ or $k\%$ is equally bad). Orthogonally, the normalized metrics are less sensitive to differences caused by variation in the number of candidates per query, which can be useful. 
    
    \subsection{Implementation Details}
    \label{ss:methods:imp}

    All models were implemented in PyTorch \cite{pytorch}. Our model, MassFormer, used a modified version of the Graphormer v2 implementation \cite{graphormer}. The WLN model used the Deep Graph Library \cite{dgl,dgllife} package for geometric deep learning. We also adapted some code from \cite{gnn_msms} for spectrum preprocessing. Before benchmarking the FP and WLN baseline models, we ran a bayesian hyperparameter sweep using Weights \& Biases \cite{wandb}, with budget of 100 initializations, to find the best-performing configuration on the NIST validation set. The list of hyperparameters that we optimized is as follows: learning rate, weight decay, dropout, minibatch size, and network-architecture specific parameters (such as hidden dimension and number of layers). For MassFormer we stuck with the Graphormer ``Large" architecture, which had available pre-trained weights. For full model details and hyperparameter configurations, please see the code repository (linked in Section \ref{s:code_availability}).

\section{Data Availability}
\label{s:data_availability}

    We used the NIST 2020 LC-ESI-MS/MS library, which is commercial and can be purchased from NIST or licensed distributors. In our code repository (see Section \ref{s:code_availability}), we include instructions for exporting the NIST data and preprocessing it to work with our models. We also used data hosted on MoNA (the ``LC-MS/MS Spectra" partition, \href{https://mona.fiehnlab.ucdavis.edu/downloads}{}) and data from the CASMI 2016 Competition (the ``Challenge Data" section, \href{http://casmi-contest.org/2016/challenges-cat2+3.shtml}{}). Exact versions of these data are available on Zenodo (\href{https://doi.org/10.5281/zenodo.7874421}{}), and can be preprocessed following instructions in our code repository.
    
\section{Code Availability}
\label{s:code_availability}

    The code for data preprocessing, model implementations, and experiment configurations is open-source (BSD-2-Clause license) and can be found in this GitHub repository: \href{https://github.com/Roestlab/massformer/}{}.

\clearpage

\bibliography{sn-bibliography}


\begin{thebibliography}{109}
\ifx \bisbn   \undefined \def \bisbn  #1{ISBN #1}\fi
\ifx \binits  \undefined \def \binits#1{#1}\fi
\ifx \bauthor  \undefined \def \bauthor#1{#1}\fi
\ifx \batitle  \undefined \def \batitle#1{#1}\fi
\ifx \bjtitle  \undefined \def \bjtitle#1{#1}\fi
\ifx \bvolume  \undefined \def \bvolume#1{\textbf{#1}}\fi
\ifx \byear  \undefined \def \byear#1{#1}\fi
\ifx \bissue  \undefined \def \bissue#1{#1}\fi
\ifx \bfpage  \undefined \def \bfpage#1{#1}\fi
\ifx \blpage  \undefined \def \blpage #1{#1}\fi
\ifx \burl  \undefined \def \burl#1{\textsf{#1}}\fi
\ifx \doiurl  \undefined \def \doiurl#1{\url{https://doi.org/#1}}\fi
\ifx \betal  \undefined \def \betal{\textit{et al.}}\fi
\ifx \binstitute  \undefined \def \binstitute#1{#1}\fi
\ifx \binstitutionaled  \undefined \def \binstitutionaled#1{#1}\fi
\ifx \bctitle  \undefined \def \bctitle#1{#1}\fi
\ifx \beditor  \undefined \def \beditor#1{#1}\fi
\ifx \bpublisher  \undefined \def \bpublisher#1{#1}\fi
\ifx \bbtitle  \undefined \def \bbtitle#1{#1}\fi
\ifx \bedition  \undefined \def \bedition#1{#1}\fi
\ifx \bseriesno  \undefined \def \bseriesno#1{#1}\fi
\ifx \blocation  \undefined \def \blocation#1{#1}\fi
\ifx \bsertitle  \undefined \def \bsertitle#1{#1}\fi
\ifx \bsnm \undefined \def \bsnm#1{#1}\fi
\ifx \bsuffix \undefined \def \bsuffix#1{#1}\fi
\ifx \bparticle \undefined \def \bparticle#1{#1}\fi
\ifx \barticle \undefined \def \barticle#1{#1}\fi
\bibcommenthead
\ifx \bconfdate \undefined \def \bconfdate #1{#1}\fi
\ifx \botherref \undefined \def \botherref #1{#1}\fi
\ifx \url \undefined \def \url#1{\textsf{#1}}\fi
\ifx \bchapter \undefined \def \bchapter#1{#1}\fi
\ifx \bbook \undefined \def \bbook#1{#1}\fi
\ifx \bcomment \undefined \def \bcomment#1{#1}\fi
\ifx \oauthor \undefined \def \oauthor#1{#1}\fi
\ifx \citeauthoryear \undefined \def \citeauthoryear#1{#1}\fi
\ifx \endbibitem  \undefined \def \endbibitem {}\fi
\ifx \bconflocation  \undefined \def \bconflocation#1{#1}\fi
\ifx \arxivurl  \undefined \def \arxivurl#1{\textsf{#1}}\fi
\csname PreBibitemsHook\endcsname

\bibitem{ms_textbook}
\begin{bbook}
\bauthor{\bsnm{Gross}, \binits{J.H.}}:
\bbtitle{Mass {Spectrometry}}.
\bpublisher{Springer},
\blocation{Cham}
(\byear{2017}).
\doiurl{10.1007/978-3-319-54398-7}.
\burl{http://link.springer.com/10.1007/978-3-319-54398-7}
Accessed 2022-11-02
\end{bbook}
\endbibitem

\bibitem{msms_review}
\begin{bchapter}
\bauthor{\bsnm{Niessen}, \binits{W.M.A.}},
\bauthor{\bsnm{Falck}, \binits{D.}}:
\bctitle{Introduction to {Mass} {Spectrometry}, a {Tutorial}}.
In: \bbtitle{Analyzing {Biomolecular} {Interactions} by {Mass} {Spectrometry}},
pp. \bfpage{1}--\blpage{54}.
\bpublisher{John Wiley \& Sons, Ltd}, \blocation{???}
(\byear{2015}).
\doiurl{10.1002/9783527673391.ch1}.
\burl{https://onlinelibrary.wiley.com/doi/abs/10.1002/9783527673391.ch1}
Accessed 2021-10-01
\end{bchapter}
\endbibitem

\bibitem{msms_proteomics}
\begin{barticle}
\bauthor{\bsnm{Aebersold}, \binits{R.}},
\bauthor{\bsnm{Mann}, \binits{M.}}:
\batitle{{{M}ass-spectrometric exploration of proteome structure and
  function}}.
\bjtitle{Nature}
\bvolume{537}(\bissue{7620}),
\bfpage{347}--\blpage{355}
(\byear{2016})
\end{barticle}
\endbibitem

\bibitem{msms_metabolomics}
\begin{barticle}
\bauthor{\bsnm{Gowda}, \binits{G.A.N.}},
\bauthor{\bsnm{Djukovic}, \binits{D.}}:
\batitle{Overview of {Mass} {Spectrometry}-{Based} {Metabolomics}:
  {Opportunities} and {Challenges}}.
\bjtitle{Methods in molecular biology (Clifton, N.J.)}
\bvolume{1198},
\bfpage{3}--\blpage{12}
(\byear{2014}).
\doiurl{10.1007/978-1-4939-1258-2\_1}.
Accessed 2021-11-04
\end{barticle}
\endbibitem

\bibitem{msms_small_mol}
\begin{barticle}
\bauthor{\bsnm{De~Vijlder}, \binits{T.}},
\bauthor{\bsnm{Valkenborg}, \binits{D.}},
\bauthor{\bsnm{Lemière}, \binits{F.}},
\bauthor{\bsnm{Romijn}, \binits{E.P.}},
\bauthor{\bsnm{Laukens}, \binits{K.}},
\bauthor{\bsnm{Cuyckens}, \binits{F.}}:
\batitle{A tutorial in small molecule identification via electrospray
  ionization-mass spectrometry: {The} practical art of structural elucidation}.
\bjtitle{Mass Spectrometry Reviews}
\bvolume{37}(\bissue{5}),
\bfpage{607}--\blpage{629}
(\byear{2018}).
\doiurl{10.1002/mas.21551}.
Accessed 2021-11-05
\end{barticle}
\endbibitem

\bibitem{msms_forensics1}
\begin{barticle}
\bauthor{\bsnm{Peters}, \binits{F.T.}}:
\batitle{Recent advances of liquid chromatography–(tandem) mass spectrometry
  in clinical and forensic toxicology}.
\bjtitle{Clinical Biochemistry}
\bvolume{44}(\bissue{1}),
\bfpage{54}--\blpage{65}
(\byear{2011}).
\doiurl{10.1016/j.clinbiochem.2010.08.008}.
Accessed 2023-01-05
\end{barticle}
\endbibitem

\bibitem{msms_forensics2}
\begin{barticle}
\bauthor{\bsnm{Van~Bocxlaer}, \binits{J.F.}},
\bauthor{\bsnm{Clauwaert}, \binits{K.M.}},
\bauthor{\bsnm{Lambert}, \binits{W.E.}},
\bauthor{\bsnm{Deforce}, \binits{D.L.}},
\bauthor{\bparticle{Van~den} \bsnm{Eeckhout}, \binits{E.G.}},
\bauthor{\bsnm{De~Leenheer}, \binits{A.P.}}:
\batitle{Liquid chromatography-mass spectrometry in forensic toxicology}.
\bjtitle{Mass Spectrometry Reviews}
\bvolume{19}(\bissue{4}),
\bfpage{165}--\blpage{214}
(\byear{2000}).
\doiurl{10.1002/1098-2787(200007)19:4<165::AID-MAS1>3.0.CO;2-Y}
\end{barticle}
\endbibitem

\bibitem{msms_env}
\begin{barticle}
\bauthor{\bsnm{Lebedev}, \binits{A.T.}}:
\batitle{Environmental {Mass} {Spectrometry}}.
\bjtitle{Annual Review of Analytical Chemistry}
\bvolume{6}(\bissue{1}),
\bfpage{163}--\blpage{189}
(\byear{2013}).
\doiurl{10.1146/annurev-anchem-062012-092604}.
Accessed 2021-11-05
\end{barticle}
\endbibitem

\bibitem{spec_sim}
\begin{barticle}
\bauthor{\bsnm{Stein}, \binits{S.E.}},
\bauthor{\bsnm{Scott}, \binits{D.R.}}:
\batitle{Optimization and testing of mass spectral library search algorithms
  for compound identification}.
\bjtitle{Journal of the American Society for Mass Spectrometry}
\bvolume{5}(\bissue{9}),
\bfpage{859}--\blpage{866}
(\byear{1994}).
\doiurl{10.1016/1044-0305(94)87009-8}.
Accessed 2021-11-05
\end{barticle}
\endbibitem

\bibitem{hmdb}
\begin{barticle}
\bauthor{\bsnm{Wishart}, \binits{D.S.}},
\bauthor{\bsnm{Feunang}, \binits{Y.D.}},
\bauthor{\bsnm{Marcu}, \binits{A.}},
\bauthor{\bsnm{Guo}, \binits{A.C.}},
\bauthor{\bsnm{Liang}, \binits{K.}},
\bauthor{\bsnm{Vázquez-Fresno}, \binits{R.}},
\bauthor{\bsnm{Sajed}, \binits{T.}},
\bauthor{\bsnm{Johnson}, \binits{D.}},
\bauthor{\bsnm{Li}, \binits{C.}},
\bauthor{\bsnm{Karu}, \binits{N.}},
\bauthor{\bsnm{Sayeeda}, \binits{Z.}},
\bauthor{\bsnm{Lo}, \binits{E.}},
\bauthor{\bsnm{Assempour}, \binits{N.}},
\bauthor{\bsnm{Berjanskii}, \binits{M.}},
\bauthor{\bsnm{Singhal}, \binits{S.}},
\bauthor{\bsnm{Arndt}, \binits{D.}},
\bauthor{\bsnm{Liang}, \binits{Y.}},
\bauthor{\bsnm{Badran}, \binits{H.}},
\bauthor{\bsnm{Grant}, \binits{J.}},
\bauthor{\bsnm{Serra-Cayuela}, \binits{A.}},
\bauthor{\bsnm{Liu}, \binits{Y.}},
\bauthor{\bsnm{Mandal}, \binits{R.}},
\bauthor{\bsnm{Neveu}, \binits{V.}},
\bauthor{\bsnm{Pon}, \binits{A.}},
\bauthor{\bsnm{Knox}, \binits{C.}},
\bauthor{\bsnm{Wilson}, \binits{M.}},
\bauthor{\bsnm{Manach}, \binits{C.}},
\bauthor{\bsnm{Scalbert}, \binits{A.}}:
\batitle{{HMDB} 4.0: the human metabolome database for 2018}.
\bjtitle{Nucleic Acids Research}
\bvolume{46}(\bissue{D1}),
\bfpage{608}--\blpage{617}
(\byear{2018}).
\doiurl{10.1093/nar/gkx1089}
\end{barticle}
\endbibitem

\bibitem{pubchem}
\begin{barticle}
\bauthor{\bsnm{Kim}, \binits{S.}},
\bauthor{\bsnm{Chen}, \binits{J.}},
\bauthor{\bsnm{Cheng}, \binits{T.}},
\bauthor{\bsnm{Gindulyte}, \binits{A.}},
\bauthor{\bsnm{He}, \binits{J.}},
\bauthor{\bsnm{He}, \binits{S.}},
\bauthor{\bsnm{Li}, \binits{Q.}},
\bauthor{\bsnm{Shoemaker}, \binits{B.A.}},
\bauthor{\bsnm{Thiessen}, \binits{P.A.}},
\bauthor{\bsnm{Yu}, \binits{B.}},
\bauthor{\bsnm{Zaslavsky}, \binits{L.}},
\bauthor{\bsnm{Zhang}, \binits{J.}},
\bauthor{\bsnm{Bolton}, \binits{E.E.}}:
\batitle{{PubChem} 2019 update: improved access to chemical data}.
\bjtitle{Nucleic Acids Research}
\bvolume{47}(\bissue{Database issue}),
\bfpage{1102}--\blpage{1109}
(\byear{2019}).
\doiurl{10.1093/nar/gky1033}.
Accessed 2021-04-16
\end{barticle}
\endbibitem

\bibitem{kegg}
\begin{barticle}
\bauthor{\bsnm{Kanehisa}, \binits{M.}},
\bauthor{\bsnm{Furumichi}, \binits{M.}},
\bauthor{\bsnm{Sato}, \binits{Y.}},
\bauthor{\bsnm{Ishiguro-Watanabe}, \binits{M.}},
\bauthor{\bsnm{Tanabe}, \binits{M.}}:
\batitle{{KEGG}: integrating viruses and cellular organisms}.
\bjtitle{Nucleic Acids Research}
\bvolume{49}(\bissue{D1}),
\bfpage{545}--\blpage{551}
(\byear{2021}).
\doiurl{10.1093/nar/gkaa970}
\end{barticle}
\endbibitem

\bibitem{massbank}
\begin{barticle}
\bauthor{\bsnm{Horai}, \binits{H.}},
\bauthor{\bsnm{Arita}, \binits{M.}},
\bauthor{\bsnm{Kanaya}, \binits{S.}},
\bauthor{\bsnm{Nihei}, \binits{Y.}},
\bauthor{\bsnm{Ikeda}, \binits{T.}},
\bauthor{\bsnm{Suwa}, \binits{K.}},
\bauthor{\bsnm{Ojima}, \binits{Y.}},
\bauthor{\bsnm{Tanaka}, \binits{K.}},
\bauthor{\bsnm{Tanaka}, \binits{S.}},
\bauthor{\bsnm{Aoshima}, \binits{K.}},
\bauthor{\bsnm{Oda}, \binits{Y.}},
\bauthor{\bsnm{Kakazu}, \binits{Y.}},
\bauthor{\bsnm{Kusano}, \binits{M.}},
\bauthor{\bsnm{Tohge}, \binits{T.}},
\bauthor{\bsnm{Matsuda}, \binits{F.}},
\bauthor{\bsnm{Sawada}, \binits{Y.}},
\bauthor{\bsnm{Hirai}, \binits{M.Y.}},
\bauthor{\bsnm{Nakanishi}, \binits{H.}},
\bauthor{\bsnm{Ikeda}, \binits{K.}},
\bauthor{\bsnm{Akimoto}, \binits{N.}},
\bauthor{\bsnm{Maoka}, \binits{T.}},
\bauthor{\bsnm{Takahashi}, \binits{H.}},
\bauthor{\bsnm{Ara}, \binits{T.}},
\bauthor{\bsnm{Sakurai}, \binits{N.}},
\bauthor{\bsnm{Suzuki}, \binits{H.}},
\bauthor{\bsnm{Shibata}, \binits{D.}},
\bauthor{\bsnm{Neumann}, \binits{S.}},
\bauthor{\bsnm{Iida}, \binits{T.}},
\bauthor{\bsnm{Tanaka}, \binits{K.}},
\bauthor{\bsnm{Funatsu}, \binits{K.}},
\bauthor{\bsnm{Matsuura}, \binits{F.}},
\bauthor{\bsnm{Soga}, \binits{T.}},
\bauthor{\bsnm{Taguchi}, \binits{R.}},
\bauthor{\bsnm{Saito}, \binits{K.}},
\bauthor{\bsnm{Nishioka}, \binits{T.}}:
\batitle{{MassBank}: a public repository for sharing mass spectral data for
  life sciences}.
\bjtitle{Journal of Mass Spectrometry}
\bvolume{45}(\bissue{7}),
\bfpage{703}--\blpage{714}
(\byear{2010}).
\doiurl{10.1002/jms.1777}.
Accessed 2020-10-27
\end{barticle}
\endbibitem

\bibitem{gnps}
\begin{barticle}
\bauthor{\bsnm{Wang}, \binits{M.}},
\bauthor{\bsnm{Carver}, \binits{J.J.}},
\bauthor{\bsnm{Phelan}, \binits{V.V.}},
\bauthor{\bsnm{Sanchez}, \binits{L.M.}},
\bauthor{\bsnm{Garg}, \binits{N.}},
\bauthor{\bsnm{Peng}, \binits{Y.}},
\bauthor{\bsnm{Nguyen}, \binits{D.D.}},
\bauthor{\bsnm{Watrous}, \binits{J.}},
\bauthor{\bsnm{Kapono}, \binits{C.A.}},
\bauthor{\bsnm{Luzzatto-Knaan}, \binits{T.}},
\bauthor{\bsnm{Porto}, \binits{C.}},
\bauthor{\bsnm{Bouslimani}, \binits{A.}},
\bauthor{\bsnm{Melnik}, \binits{A.V.}},
\bauthor{\bsnm{Meehan}, \binits{M.J.}},
\bauthor{\bsnm{Liu}, \binits{W.T.}},
\bauthor{\bsnm{Crüsemann}, \binits{M.}},
\bauthor{\bsnm{Boudreau}, \binits{P.D.}},
\bauthor{\bsnm{Esquenazi}, \binits{E.}},
\bauthor{\bsnm{Sandoval-Calderón}, \binits{M.}},
\bauthor{\bsnm{Kersten}, \binits{R.D.}},
\bauthor{\bsnm{Pace}, \binits{L.A.}},
\bauthor{\bsnm{Quinn}, \binits{R.A.}},
\bauthor{\bsnm{Duncan}, \binits{K.R.}},
\bauthor{\bsnm{Hsu}, \binits{C.C.}},
\bauthor{\bsnm{Floros}, \binits{D.J.}},
\bauthor{\bsnm{Gavilan}, \binits{R.G.}},
\bauthor{\bsnm{Kleigrewe}, \binits{K.}},
\bauthor{\bsnm{Northen}, \binits{T.}},
\bauthor{\bsnm{Dutton}, \binits{R.J.}},
\bauthor{\bsnm{Parrot}, \binits{D.}},
\bauthor{\bsnm{Carlson}, \binits{E.E.}},
\bauthor{\bsnm{Aigle}, \binits{B.}},
\bauthor{\bsnm{Michelsen}, \binits{C.F.}},
\bauthor{\bsnm{Jelsbak}, \binits{L.}},
\bauthor{\bsnm{Sohlenkamp}, \binits{C.}},
\bauthor{\bsnm{Pevzner}, \binits{P.}},
\bauthor{\bsnm{Edlund}, \binits{A.}},
\bauthor{\bsnm{McLean}, \binits{J.}},
\bauthor{\bsnm{Piel}, \binits{J.}},
\bauthor{\bsnm{Murphy}, \binits{B.T.}},
\bauthor{\bsnm{Gerwick}, \binits{L.}},
\bauthor{\bsnm{Liaw}, \binits{C.C.}},
\bauthor{\bsnm{Yang}, \binits{Y.L.}},
\bauthor{\bsnm{Humpf}, \binits{H.U.}},
\bauthor{\bsnm{Maansson}, \binits{M.}},
\bauthor{\bsnm{Keyzers}, \binits{R.A.}},
\bauthor{\bsnm{Sims}, \binits{A.C.}},
\bauthor{\bsnm{Johnson}, \binits{A.R.}},
\bauthor{\bsnm{Sidebottom}, \binits{A.M.}},
\bauthor{\bsnm{Sedio}, \binits{B.E.}},
\bauthor{\bsnm{Klitgaard}, \binits{A.}},
\bauthor{\bsnm{Larson}, \binits{C.B.}},
\bauthor{\bsnm{P}, \binits{C.A.B.}},
\bauthor{\bsnm{Torres-Mendoza}, \binits{D.}},
\bauthor{\bsnm{Gonzalez}, \binits{D.J.}},
\bauthor{\bsnm{Silva}, \binits{D.B.}},
\bauthor{\bsnm{Marques}, \binits{L.M.}},
\bauthor{\bsnm{Demarque}, \binits{D.P.}},
\bauthor{\bsnm{Pociute}, \binits{E.}},
\bauthor{\bsnm{O'Neill}, \binits{E.C.}},
\bauthor{\bsnm{Briand}, \binits{E.}},
\bauthor{\bsnm{Helfrich}, \binits{E.J.N.}},
\bauthor{\bsnm{Granatosky}, \binits{E.A.}},
\bauthor{\bsnm{Glukhov}, \binits{E.}},
\bauthor{\bsnm{Ryffel}, \binits{F.}},
\bauthor{\bsnm{Houson}, \binits{H.}},
\bauthor{\bsnm{Mohimani}, \binits{H.}},
\bauthor{\bsnm{Kharbush}, \binits{J.J.}},
\bauthor{\bsnm{Zeng}, \binits{Y.}},
\bauthor{\bsnm{Vorholt}, \binits{J.A.}},
\bauthor{\bsnm{Kurita}, \binits{K.L.}},
\bauthor{\bsnm{Charusanti}, \binits{P.}},
\bauthor{\bsnm{McPhail}, \binits{K.L.}},
\bauthor{\bsnm{Nielsen}, \binits{K.F.}},
\bauthor{\bsnm{Vuong}, \binits{L.}},
\bauthor{\bsnm{Elfeki}, \binits{M.}},
\bauthor{\bsnm{Traxler}, \binits{M.F.}},
\bauthor{\bsnm{Engene}, \binits{N.}},
\bauthor{\bsnm{Koyama}, \binits{N.}},
\bauthor{\bsnm{Vining}, \binits{O.B.}},
\bauthor{\bsnm{Baric}, \binits{R.}},
\bauthor{\bsnm{Silva}, \binits{R.R.}},
\bauthor{\bsnm{Mascuch}, \binits{S.J.}},
\bauthor{\bsnm{Tomasi}, \binits{S.}},
\bauthor{\bsnm{Jenkins}, \binits{S.}},
\bauthor{\bsnm{Macherla}, \binits{V.}},
\bauthor{\bsnm{Hoffman}, \binits{T.}},
\bauthor{\bsnm{Agarwal}, \binits{V.}},
\bauthor{\bsnm{Williams}, \binits{P.G.}},
\bauthor{\bsnm{Dai}, \binits{J.}},
\bauthor{\bsnm{Neupane}, \binits{R.}},
\bauthor{\bsnm{Gurr}, \binits{J.}},
\bauthor{\bsnm{Rodríguez}, \binits{A.M.C.}},
\bauthor{\bsnm{Lamsa}, \binits{A.}},
\bauthor{\bsnm{Zhang}, \binits{C.}},
\bauthor{\bsnm{Dorrestein}, \binits{K.}},
\bauthor{\bsnm{Duggan}, \binits{B.M.}},
\bauthor{\bsnm{Almaliti}, \binits{J.}},
\bauthor{\bsnm{Allard}, \binits{P.M.}},
\bauthor{\bsnm{Phapale}, \binits{P.}},
\bauthor{\bsnm{Nothias}, \binits{L.F.}},
\bauthor{\bsnm{Alexandrov}, \binits{T.}},
\bauthor{\bsnm{Litaudon}, \binits{M.}},
\bauthor{\bsnm{Wolfender}, \binits{J.L.}},
\bauthor{\bsnm{Kyle}, \binits{J.E.}},
\bauthor{\bsnm{Metz}, \binits{T.O.}},
\bauthor{\bsnm{Peryea}, \binits{T.}},
\bauthor{\bsnm{Nguyen}, \binits{D.T.}},
\bauthor{\bsnm{VanLeer}, \binits{D.}},
\bauthor{\bsnm{Shinn}, \binits{P.}},
\bauthor{\bsnm{Jadhav}, \binits{A.}},
\bauthor{\bsnm{Müller}, \binits{R.}},
\bauthor{\bsnm{Waters}, \binits{K.M.}},
\bauthor{\bsnm{Shi}, \binits{W.}},
\bauthor{\bsnm{Liu}, \binits{X.}},
\bauthor{\bsnm{Zhang}, \binits{L.}},
\bauthor{\bsnm{Knight}, \binits{R.}},
\bauthor{\bsnm{Jensen}, \binits{P.R.}},
\bauthor{\bsnm{Palsson}, \binits{B.O.}},
\bauthor{\bsnm{Pogliano}, \binits{K.}},
\bauthor{\bsnm{Linington}, \binits{R.G.}},
\bauthor{\bsnm{Gutiérrez}, \binits{M.}},
\bauthor{\bsnm{Lopes}, \binits{N.P.}},
\bauthor{\bsnm{Gerwick}, \binits{W.H.}},
\bauthor{\bsnm{Moore}, \binits{B.S.}},
\bauthor{\bsnm{Dorrestein}, \binits{P.C.}},
\bauthor{\bsnm{Bandeira}, \binits{N.}}:
\batitle{{{S}haring and community curation of mass spectrometry data with
  {G}lobal {N}atural {P}roducts {S}ocial {M}olecular {N}etworking}}.
\bjtitle{Nat Biotechnol}
\bvolume{34}(\bissue{8}),
\bfpage{828}--\blpage{837}
(\byear{2016})
\end{barticle}
\endbibitem

\bibitem{respect}
\begin{barticle}
\bauthor{\bsnm{Sawada}, \binits{Y.}},
\bauthor{\bsnm{Nakabayashi}, \binits{R.}},
\bauthor{\bsnm{Yamada}, \binits{Y.}},
\bauthor{\bsnm{Suzuki}, \binits{M.}},
\bauthor{\bsnm{Sato}, \binits{M.}},
\bauthor{\bsnm{Sakata}, \binits{A.}},
\bauthor{\bsnm{Akiyama}, \binits{K.}},
\bauthor{\bsnm{Sakurai}, \binits{T.}},
\bauthor{\bsnm{Matsuda}, \binits{F.}},
\bauthor{\bsnm{Aoki}, \binits{T.}},
\bauthor{\bsnm{Hirai}, \binits{M.Y.}},
\bauthor{\bsnm{Saito}, \binits{K.}}:
\batitle{{RIKEN} tandem mass spectral database ({ReSpect}) for phytochemicals:
  {A} plant-specific {MS}/{MS}-based data resource and database}.
\bjtitle{Phytochemistry}
\bvolume{82},
\bfpage{38}--\blpage{45}
(\byear{2012}).
\doiurl{10.1016/j.phytochem.2012.07.007}.
Accessed 2023-03-09
\end{barticle}
\endbibitem

\bibitem{nist1}
\begin{barticle}
\bauthor{\bsnm{Stein}, \binits{S.}}:
\batitle{Mass {Spectral} {Reference} {Libraries}: {An} {Ever}-{Expanding}
  {Resource} for {Chemical} {Identification}}.
\bjtitle{Analytical Chemistry}
\bvolume{84}(\bissue{17}),
\bfpage{7274}--\blpage{7282}
(\byear{2012}).
\doiurl{10.1021/ac301205z}.
Accessed 2021-10-01
\end{barticle}
\endbibitem

\bibitem{nist2}
\begin{barticle}
\bauthor{\bsnm{Yang}, \binits{X.}},
\bauthor{\bsnm{Neta}, \binits{P.}},
\bauthor{\bsnm{Stein}, \binits{S.E.}}:
\batitle{Quality {Control} for {Building} {Libraries} from {Electrospray}
  {Ionization} {Tandem} {Mass} {Spectra}}.
\bjtitle{Analytical Chemistry}
\bvolume{86}(\bissue{13}),
\bfpage{6393}--\blpage{6400}
(\byear{2014}).
\doiurl{10.1021/ac500711m}.
Accessed 2021-07-05
\end{barticle}
\endbibitem

\bibitem{metlin}
\begin{barticle}
\bauthor{\bsnm{Guijas}, \binits{C.}},
\bauthor{\bsnm{Montenegro-Burke}, \binits{J.R.}},
\bauthor{\bsnm{Domingo-Almenara}, \binits{X.}},
\bauthor{\bsnm{Palermo}, \binits{A.}},
\bauthor{\bsnm{Warth}, \binits{B.}},
\bauthor{\bsnm{Hermann}, \binits{G.}},
\bauthor{\bsnm{Koellensperger}, \binits{G.}},
\bauthor{\bsnm{Huan}, \binits{T.}},
\bauthor{\bsnm{Uritboonthai}, \binits{W.}},
\bauthor{\bsnm{Aisporna}, \binits{A.E.}},
\bauthor{\bsnm{Wolan}, \binits{D.W.}},
\bauthor{\bsnm{Spilker}, \binits{M.E.}},
\bauthor{\bsnm{Benton}, \binits{H.P.}},
\bauthor{\bsnm{Siuzdak}, \binits{G.}}:
\batitle{{METLIN}: {A} {Technology} {Platform} for {Identifying} {Knowns} and
  {Unknowns}}.
\bjtitle{Analytical Chemistry}
\bvolume{90}(\bissue{5}),
\bfpage{3156}--\blpage{3164}
(\byear{2018}).
\doiurl{10.1021/acs.analchem.7b04424}.
Accessed 2020-11-25
\end{barticle}
\endbibitem

\bibitem{wiley}
\begin{botherref}
\oauthor{\bsnm{Wiley}, \binits{J.}},
\oauthor{\bsnm{Sons}}:
Wiley Registry of Mass Spectral Data, 12th Edition
(2020)
\end{botherref}
\endbibitem

\bibitem{cfm}
\begin{barticle}
\bauthor{\bsnm{Allen}, \binits{F.}},
\bauthor{\bsnm{Greiner}, \binits{R.}},
\bauthor{\bsnm{Wishart}, \binits{D.}}:
\batitle{Competitive fragmentation modeling of {ESI}-{MS}/{MS} spectra for
  putative metabolite identification}.
\bjtitle{Metabolomics}
\bvolume{11}(\bissue{1}),
\bfpage{98}--\blpage{110}
(\byear{2015}).
\doiurl{10.1007/s11306-014-0676-4}.
Accessed 2020-10-13
\end{barticle}
\endbibitem

\bibitem{cfm3}
\begin{barticle}
\bauthor{\bsnm{Djoumbou-Feunang}, \binits{Y.}},
\bauthor{\bsnm{Pon}, \binits{A.}},
\bauthor{\bsnm{Karu}, \binits{N.}},
\bauthor{\bsnm{Zheng}, \binits{J.}},
\bauthor{\bsnm{Li}, \binits{C.}},
\bauthor{\bsnm{Arndt}, \binits{D.}},
\bauthor{\bsnm{Gautam}, \binits{M.}},
\bauthor{\bsnm{Allen}, \binits{F.}},
\bauthor{\bsnm{Wishart}, \binits{D.S.}}:
\batitle{{CFM}-{ID} 3.0: {Significantly} {Improved} {ESI}-{MS}/{MS}
  {Prediction} and {Compound} {Identification}}.
\bjtitle{Metabolites}
\bvolume{9}(\bissue{4}),
\bfpage{72}
(\byear{2019}).
\doiurl{10.3390/metabo9040072}.
Accessed 2021-12-29
\end{barticle}
\endbibitem

\bibitem{cfm4}
\begin{barticle}
\bauthor{\bsnm{Wang}, \binits{F.}},
\bauthor{\bsnm{Liigand}, \binits{J.}},
\bauthor{\bsnm{Tian}, \binits{S.}},
\bauthor{\bsnm{Arndt}, \binits{D.}},
\bauthor{\bsnm{Greiner}, \binits{R.}},
\bauthor{\bsnm{Wishart}, \binits{D.S.}}:
\batitle{{CFM}-{ID} 4.0: {More} {Accurate} {ESI}-{MS}/{MS} {Spectral}
  {Prediction} and {Compound} {Identification}}.
\bjtitle{Analytical Chemistry}
(\byear{2021}).
\doiurl{10.1021/acs.analchem.1c01465}.
Accessed 2021-08-23
\end{barticle}
\endbibitem

\bibitem{alphazero}
\begin{botherref}
\oauthor{\bsnm{Silver}, \binits{D.}},
\oauthor{\bsnm{Hubert}, \binits{T.}},
\oauthor{\bsnm{Schrittwieser}, \binits{J.}},
\oauthor{\bsnm{Antonoglou}, \binits{I.}},
\oauthor{\bsnm{Lai}, \binits{M.}},
\oauthor{\bsnm{Guez}, \binits{A.}},
\oauthor{\bsnm{Lanctot}, \binits{M.}},
\oauthor{\bsnm{Sifre}, \binits{L.}},
\oauthor{\bsnm{Kumaran}, \binits{D.}},
\oauthor{\bsnm{Graepel}, \binits{T.}},
\oauthor{\bsnm{Lillicrap}, \binits{T.}},
\oauthor{\bsnm{Simonyan}, \binits{K.}},
\oauthor{\bsnm{Hassabis}, \binits{D.}}:
Mastering {Chess} and {Shogi} by {Self}-{Play} with a {General} {Reinforcement}
  {Learning} {Algorithm}.
arXiv.
arXiv:1712.01815 [cs]
(2017).
\doiurl{10.48550/arXiv.1712.01815}.
\url{http://arxiv.org/abs/1712.01815}
Accessed 2023-01-25
\end{botherref}
\endbibitem

\bibitem{alphafold}
\begin{barticle}
\bauthor{\bsnm{Jumper}, \binits{J.}},
\bauthor{\bsnm{Evans}, \binits{R.}},
\bauthor{\bsnm{Pritzel}, \binits{A.}},
\bauthor{\bsnm{Green}, \binits{T.}},
\bauthor{\bsnm{Figurnov}, \binits{M.}},
\bauthor{\bsnm{Ronneberger}, \binits{O.}},
\bauthor{\bsnm{Tunyasuvunakool}, \binits{K.}},
\bauthor{\bsnm{Bates}, \binits{R.}},
\bauthor{\bsnm{Žídek}, \binits{A.}},
\bauthor{\bsnm{Potapenko}, \binits{A.}},
\bauthor{\bsnm{Bridgland}, \binits{A.}},
\bauthor{\bsnm{Meyer}, \binits{C.}},
\bauthor{\bsnm{Kohl}, \binits{S.A.A.}},
\bauthor{\bsnm{Ballard}, \binits{A.J.}},
\bauthor{\bsnm{Cowie}, \binits{A.}},
\bauthor{\bsnm{Romera-Paredes}, \binits{B.}},
\bauthor{\bsnm{Nikolov}, \binits{S.}},
\bauthor{\bsnm{Jain}, \binits{R.}},
\bauthor{\bsnm{Adler}, \binits{J.}},
\bauthor{\bsnm{Back}, \binits{T.}},
\bauthor{\bsnm{Petersen}, \binits{S.}},
\bauthor{\bsnm{Reiman}, \binits{D.}},
\bauthor{\bsnm{Clancy}, \binits{E.}},
\bauthor{\bsnm{Zielinski}, \binits{M.}},
\bauthor{\bsnm{Steinegger}, \binits{M.}},
\bauthor{\bsnm{Pacholska}, \binits{M.}},
\bauthor{\bsnm{Berghammer}, \binits{T.}},
\bauthor{\bsnm{Bodenstein}, \binits{S.}},
\bauthor{\bsnm{Silver}, \binits{D.}},
\bauthor{\bsnm{Vinyals}, \binits{O.}},
\bauthor{\bsnm{Senior}, \binits{A.W.}},
\bauthor{\bsnm{Kavukcuoglu}, \binits{K.}},
\bauthor{\bsnm{Kohli}, \binits{P.}},
\bauthor{\bsnm{Hassabis}, \binits{D.}}:
\batitle{Highly accurate protein structure prediction with {AlphaFold}}.
\bjtitle{Nature}
\bvolume{596}(\bissue{7873}),
\bfpage{583}--\blpage{589}
(\byear{2021}).
\doiurl{10.1038/s41586-021-03819-2}.
\bcomment{Number: 7873 Publisher: Nature Publishing Group}.
Accessed 2023-01-25
\end{barticle}
\endbibitem

\bibitem{stable_diffusion}
\begin{botherref}
\oauthor{\bsnm{Rombach}, \binits{R.}},
\oauthor{\bsnm{Blattmann}, \binits{A.}},
\oauthor{\bsnm{Lorenz}, \binits{D.}},
\oauthor{\bsnm{Esser}, \binits{P.}},
\oauthor{\bsnm{Ommer}, \binits{B.}}:
High-{Resolution} {Image} {Synthesis} with {Latent} {Diffusion} {Models}.
arXiv.
arXiv:2112.10752 [cs]
(2022).
\doiurl{10.48550/arXiv.2112.10752}.
\url{http://arxiv.org/abs/2112.10752}
Accessed 2023-01-25
\end{botherref}
\endbibitem

\bibitem{gpt3}
\begin{botherref}
\oauthor{\bsnm{Brown}, \binits{T.B.}},
\oauthor{\bsnm{Mann}, \binits{B.}},
\oauthor{\bsnm{Ryder}, \binits{N.}},
\oauthor{\bsnm{Subbiah}, \binits{M.}},
\oauthor{\bsnm{Kaplan}, \binits{J.}},
\oauthor{\bsnm{Dhariwal}, \binits{P.}},
\oauthor{\bsnm{Neelakantan}, \binits{A.}},
\oauthor{\bsnm{Shyam}, \binits{P.}},
\oauthor{\bsnm{Sastry}, \binits{G.}},
\oauthor{\bsnm{Askell}, \binits{A.}},
\oauthor{\bsnm{Agarwal}, \binits{S.}},
\oauthor{\bsnm{Herbert-Voss}, \binits{A.}},
\oauthor{\bsnm{Krueger}, \binits{G.}},
\oauthor{\bsnm{Henighan}, \binits{T.}},
\oauthor{\bsnm{Child}, \binits{R.}},
\oauthor{\bsnm{Ramesh}, \binits{A.}},
\oauthor{\bsnm{Ziegler}, \binits{D.M.}},
\oauthor{\bsnm{Wu}, \binits{J.}},
\oauthor{\bsnm{Winter}, \binits{C.}},
\oauthor{\bsnm{Hesse}, \binits{C.}},
\oauthor{\bsnm{Chen}, \binits{M.}},
\oauthor{\bsnm{Sigler}, \binits{E.}},
\oauthor{\bsnm{Litwin}, \binits{M.}},
\oauthor{\bsnm{Gray}, \binits{S.}},
\oauthor{\bsnm{Chess}, \binits{B.}},
\oauthor{\bsnm{Clark}, \binits{J.}},
\oauthor{\bsnm{Berner}, \binits{C.}},
\oauthor{\bsnm{McCandlish}, \binits{S.}},
\oauthor{\bsnm{Radford}, \binits{A.}},
\oauthor{\bsnm{Sutskever}, \binits{I.}},
\oauthor{\bsnm{Amodei}, \binits{D.}}:
Language {Models} are {Few}-{Shot} {Learners}.
arXiv.
arXiv:2005.14165 [cs]
(2020).
\doiurl{10.48550/arXiv.2005.14165}.
\url{http://arxiv.org/abs/2005.14165}
Accessed 2023-01-25
\end{botherref}
\endbibitem

\bibitem{neims}
\begin{barticle}
\bauthor{\bsnm{Wei}, \binits{J.N.}},
\bauthor{\bsnm{Belanger}, \binits{D.}},
\bauthor{\bsnm{Adams}, \binits{R.P.}},
\bauthor{\bsnm{Sculley}, \binits{D.}}:
\batitle{Rapid prediction of electron–ionization mass spectrometry using
  neural networks}.
\bjtitle{ACS Central Science}
\bvolume{5}(\bissue{4}),
\bfpage{700}--\blpage{708}
(\byear{2019}).
\doiurl{10.1021/acscentsci.9b00085}.
Accessed 2021-03-03
\end{barticle}
\endbibitem

\bibitem{gnn_msms}
\begin{botherref}
\oauthor{\bsnm{Zhu}, \binits{H.}},
\oauthor{\bsnm{Liu}, \binits{L.}},
\oauthor{\bsnm{Hassoun}, \binits{S.}}:
Using {Graph} {Neural} {Networks} for {Mass} {Spectrometry} {Prediction}.
arXiv:2010.04661 [cs]
(2020).
Accessed 2021-02-20
\end{botherref}
\endbibitem

\bibitem{esp}
\begin{botherref}
\oauthor{\bsnm{Li}, \binits{X.}},
\oauthor{\bsnm{Zhu}, \binits{H.}},
\oauthor{\bsnm{Liu}, \binits{L.-p.}},
\oauthor{\bsnm{Hassoun}, \binits{S.}}:
Ensemble {Spectral} {Prediction} ({ESP}) {Model} for {Metabolite} {Annotation}.
arXiv:2203.13783 [cs, q-bio]
(2022).
arXiv: 2203.13783.
Accessed 2022-04-28
\end{botherref}
\endbibitem

\bibitem{how_powerful_gnns}
\begin{botherref}
\oauthor{\bsnm{Xu}, \binits{K.}},
\oauthor{\bsnm{Hu}, \binits{W.}},
\oauthor{\bsnm{Leskovec}, \binits{J.}},
\oauthor{\bsnm{Jegelka}, \binits{S.}}:
How {Powerful} are {Graph} {Neural} {Networks}?
arXiv:1810.00826 [cs, stat]
(2019).
arXiv: 1810.00826.
Accessed 2020-05-11
\end{botherref}
\endbibitem

\bibitem{oversmooth1}
\begin{botherref}
\oauthor{\bsnm{Chen}, \binits{D.}},
\oauthor{\bsnm{Lin}, \binits{Y.}},
\oauthor{\bsnm{Li}, \binits{W.}},
\oauthor{\bsnm{Li}, \binits{P.}},
\oauthor{\bsnm{Zhou}, \binits{J.}},
\oauthor{\bsnm{Sun}, \binits{X.}}:
Measuring and {Relieving} the {Over}-smoothing {Problem} for {Graph} {Neural}
  {Networks} from the {Topological} {View}.
arXiv.
arXiv:1909.03211 [cs, stat]
(2019).
\url{http://arxiv.org/abs/1909.03211}
Accessed 2023-03-08
\end{botherref}
\endbibitem

\bibitem{oversmooth2}
\begin{bchapter}
\bauthor{\bsnm{Liu}, \binits{M.}},
\bauthor{\bsnm{Gao}, \binits{H.}},
\bauthor{\bsnm{Ji}, \binits{S.}}:
\bctitle{Towards {Deeper} {Graph} {Neural} {Networks}}.
In: \bbtitle{Proceedings of the 26th {ACM} {SIGKDD} {International}
  {Conference} on {Knowledge} {Discovery} \& {Data} {Mining}},
pp. \bfpage{338}--\blpage{348}
(\byear{2020}).
\doiurl{10.1145/3394486.3403076}.
\bcomment{arXiv:2007.09296 [cs, stat]}.
\burl{http://arxiv.org/abs/2007.09296}
Accessed 2023-03-08
\end{bchapter}
\endbibitem

\bibitem{graff}
\begin{botherref}
\oauthor{\bsnm{Murphy}, \binits{M.}},
\oauthor{\bsnm{Jegelka}, \binits{S.}},
\oauthor{\bsnm{Fraenkel}, \binits{E.}},
\oauthor{\bsnm{Kind}, \binits{T.}},
\oauthor{\bsnm{Healey}, \binits{D.}},
\oauthor{\bsnm{Butler}, \binits{T.}}:
Efficiently predicting high resolution mass spectra with graph neural networks.
arXiv.
arXiv:2301.11419 [cs, q-bio]
(2023).
\url{http://arxiv.org/abs/2301.11419}
Accessed 2023-03-02
\end{botherref}
\endbibitem

\bibitem{scarf}
\begin{botherref}
\oauthor{\bsnm{Goldman}, \binits{S.}},
\oauthor{\bsnm{Bradshaw}, \binits{J.}},
\oauthor{\bsnm{Xin}, \binits{J.}},
\oauthor{\bsnm{Coley}, \binits{C.W.}}:
Prefix-tree {Decoding} for {Predicting} {Mass} {Spectra} from {Molecules}.
arXiv.
arXiv:2303.06470 [cs, q-bio]
(2023).
\doiurl{10.48550/arXiv.2303.06470}.
\url{http://arxiv.org/abs/2303.06470}
Accessed 2023-03-17
\end{botherref}
\endbibitem

\bibitem{3dmolms}
\begin{botherref}
\oauthor{\bsnm{Hong}, \binits{Y.}},
\oauthor{\bsnm{Li}, \binits{S.}},
\oauthor{\bsnm{Welch}, \binits{C.J.}},
\oauthor{\bsnm{Tichy}, \binits{S.}},
\oauthor{\bsnm{Ye}, \binits{Y.}},
\oauthor{\bsnm{Tang}, \binits{H.}}:
{3DMolMS}: {Prediction} of {Tandem} {Mass} {Spectra} from {Three} {Dimensional}
  {Molecular} {Conformations}.
bioRxiv.
Pages: 2023.03.15.532823 Section: New Results
(2023).
\doiurl{10.1101/2023.03.15.532823}.
\url{https://www.biorxiv.org/content/10.1101/2023.03.15.532823v1}
Accessed 2023-03-17
\end{botherref}
\endbibitem

\bibitem{rassp}
\begin{barticle}
\bauthor{\bsnm{Zhu}, \binits{R.L.}},
\bauthor{\bsnm{Jonas}, \binits{E.}}:
\batitle{Rapid {Approximate} {Subset}-{Based} {Spectra} {Prediction} for
  {Electron} {Ionization}–{Mass} {Spectrometry}}.
\bjtitle{Analytical Chemistry}
\bvolume{95}(\bissue{5}),
\bfpage{2653}--\blpage{2663}
(\byear{2023}).
\doiurl{10.1021/acs.analchem.2c02093}.
\bcomment{Publisher: American Chemical Society}.
Accessed 2023-03-23
\end{barticle}
\endbibitem

\bibitem{graphormer}
\begin{botherref}
\oauthor{\bsnm{Ying}, \binits{C.}},
\oauthor{\bsnm{Cai}, \binits{T.}},
\oauthor{\bsnm{Luo}, \binits{S.}},
\oauthor{\bsnm{Zheng}, \binits{S.}},
\oauthor{\bsnm{Ke}, \binits{G.}},
\oauthor{\bsnm{He}, \binits{D.}},
\oauthor{\bsnm{Shen}, \binits{Y.}},
\oauthor{\bsnm{Liu}, \binits{T.-Y.}}:
Do transformers really perform bad for graph representation?
Neural Information Processing Systems (NeurIPS)
(2021)
\end{botherref}
\endbibitem

\bibitem{scaffold}
\begin{barticle}
\bauthor{\bsnm{Bemis}, \binits{G.W.}},
\bauthor{\bsnm{Murcko}, \binits{M.A.}}:
\batitle{The {Properties} of {Known} {Drugs}. 1. {Molecular} {Frameworks}}.
\bjtitle{Journal of Medicinal Chemistry}
\bvolume{39}(\bissue{15}),
\bfpage{2887}--\blpage{2893}
(\byear{1996}).
\doiurl{10.1021/jm9602928}.
Accessed 2021-11-04
\end{barticle}
\endbibitem

\bibitem{rdkit}
\begin{botherref}
\oauthor{\bsnm{Landrum}, \binits{G.}}:
RDKit: Open-source cheminformatics.
\url{http://www.rdkit.org}
\end{botherref}
\endbibitem

\bibitem{classyfire}
\begin{barticle}
\bauthor{\bsnm{Djoumbou~Feunang}, \binits{Y.}},
\bauthor{\bsnm{Eisner}, \binits{R.}},
\bauthor{\bsnm{Knox}, \binits{C.}},
\bauthor{\bsnm{Chepelev}, \binits{L.}},
\bauthor{\bsnm{Hastings}, \binits{J.}},
\bauthor{\bsnm{Owen}, \binits{G.}},
\bauthor{\bsnm{Fahy}, \binits{E.}},
\bauthor{\bsnm{Steinbeck}, \binits{C.}},
\bauthor{\bsnm{Subramanian}, \binits{S.}},
\bauthor{\bsnm{Bolton}, \binits{E.}},
\bauthor{\bsnm{Greiner}, \binits{R.}},
\bauthor{\bsnm{Wishart}, \binits{D.S.}}:
\batitle{{ClassyFire}: automated chemical classification with a comprehensive,
  computable taxonomy}.
\bjtitle{Journal of Cheminformatics}
\bvolume{8}(\bissue{1}),
\bfpage{61}
(\byear{2016}).
\doiurl{10.1186/s13321-016-0174-y}.
Accessed 2023-01-05
\end{barticle}
\endbibitem

\bibitem{lipidblast}
\begin{barticle}
\bauthor{\bsnm{Kind}, \binits{T.}},
\bauthor{\bsnm{Liu}, \binits{K.-H.}},
\bauthor{\bsnm{Lee}, \binits{D.Y.}},
\bauthor{\bsnm{DeFelice}, \binits{B.}},
\bauthor{\bsnm{Meissen}, \binits{J.K.}},
\bauthor{\bsnm{Fiehn}, \binits{O.}}:
\batitle{{LipidBlast} in silico tandem mass spectrometry database for lipid
  identification}.
\bjtitle{Nature Methods}
\bvolume{10}(\bissue{8}),
\bfpage{755}--\blpage{758}
(\byear{2013}).
\doiurl{10.1038/nmeth.2551}.
\bcomment{Number: 8 Publisher: Nature Publishing Group}.
Accessed 2023-03-07
\end{barticle}
\endbibitem

\bibitem{gi_1}
\begin{botherref}
\oauthor{\bsnm{Shrikumar}, \binits{A.}},
\oauthor{\bsnm{Greenside}, \binits{P.}},
\oauthor{\bsnm{Kundaje}, \binits{A.}}:
Learning {Important} {Features} {Through} {Propagating} {Activation}
  {Differences}.
arXiv.
arXiv:1704.02685 [cs]
(2019).
\doiurl{10.48550/arXiv.1704.02685}.
\url{http://arxiv.org/abs/1704.02685}
Accessed 2023-02-13
\end{botherref}
\endbibitem

\bibitem{gi_2}
\begin{botherref}
\oauthor{\bsnm{Ancona}, \binits{M.}},
\oauthor{\bsnm{Ceolini}, \binits{E.}},
\oauthor{\bsnm{Öztireli}, \binits{C.}},
\oauthor{\bsnm{Gross}, \binits{M.}}:
Towards better understanding of gradient-based attribution methods for {Deep}
  {Neural} {Networks}.
arXiv.
arXiv:1711.06104 [cs, stat]
(2018).
\doiurl{10.48550/arXiv.1711.06104}.
\url{http://arxiv.org/abs/1711.06104}
Accessed 2023-02-13
\end{botherref}
\endbibitem

\bibitem{xai_transformer}
\begin{botherref}
\oauthor{\bsnm{Ali}, \binits{A.}},
\oauthor{\bsnm{Schnake}, \binits{T.}},
\oauthor{\bsnm{Eberle}, \binits{O.}},
\oauthor{\bsnm{Montavon}, \binits{G.}},
\oauthor{\bsnm{Müller}, \binits{K.-R.}},
\oauthor{\bsnm{Wolf}, \binits{L.}}:
{XAI} for {Transformers}: {Better} {Explanations} through {Conservative}
  {Propagation}.
arXiv.
arXiv:2202.07304 [cs]
(2022).
\url{http://arxiv.org/abs/2202.07304}
Accessed 2023-01-19
\end{botherref}
\endbibitem

\bibitem{sirius}
\begin{barticle}
\bauthor{\bsnm{Dührkop}, \binits{K.}},
\bauthor{\bsnm{Fleischauer}, \binits{M.}},
\bauthor{\bsnm{Ludwig}, \binits{M.}},
\bauthor{\bsnm{Aksenov}, \binits{A.A.}},
\bauthor{\bsnm{Melnik}, \binits{A.V.}},
\bauthor{\bsnm{Meusel}, \binits{M.}},
\bauthor{\bsnm{Dorrestein}, \binits{P.C.}},
\bauthor{\bsnm{Rousu}, \binits{J.}},
\bauthor{\bsnm{Böcker}, \binits{S.}}:
\batitle{{SIRIUS} 4: a rapid tool for turning tandem mass spectra into
  metabolite structure information}.
\bjtitle{Nature Methods}
\bvolume{16}(\bissue{4}),
\bfpage{299}--\blpage{302}
(\byear{2019}).
\doiurl{10.1038/s41592-019-0344-8}.
Accessed 2020-10-29
\end{barticle}
\endbibitem

\bibitem{casmi2012}
\begin{barticle}
\bauthor{\bsnm{Schymanski}, \binits{E.L.}},
\bauthor{\bsnm{Neumann}, \binits{S.}}:
\batitle{{CASMI}: {And} the {Winner} is . . .}
\bjtitle{Metabolites}
\bvolume{3}(\bissue{2}),
\bfpage{412}--\blpage{439}
(\byear{2013}).
\doiurl{10.3390/metabo3020412}.
\bcomment{Number: 2 Publisher: Multidisciplinary Digital Publishing Institute}.
Accessed 2022-01-12
\end{barticle}
\endbibitem

\bibitem{casmi2016}
\begin{barticle}
\bauthor{\bsnm{Schymanski}, \binits{E.L.}},
\bauthor{\bsnm{Ruttkies}, \binits{C.}},
\bauthor{\bsnm{Krauss}, \binits{M.}},
\bauthor{\bsnm{Brouard}, \binits{C.}},
\bauthor{\bsnm{Kind}, \binits{T.}},
\bauthor{\bsnm{Dührkop}, \binits{K.}},
\bauthor{\bsnm{Allen}, \binits{F.}},
\bauthor{\bsnm{Vaniya}, \binits{A.}},
\bauthor{\bsnm{Verdegem}, \binits{D.}},
\bauthor{\bsnm{Böcker}, \binits{S.}},
\bauthor{\bsnm{Rousu}, \binits{J.}},
\bauthor{\bsnm{Shen}, \binits{H.}},
\bauthor{\bsnm{Tsugawa}, \binits{H.}},
\bauthor{\bsnm{Sajed}, \binits{T.}},
\bauthor{\bsnm{Fiehn}, \binits{O.}},
\bauthor{\bsnm{Ghesquière}, \binits{B.}},
\bauthor{\bsnm{Neumann}, \binits{S.}}:
\batitle{Critical {Assessment} of {Small} {Molecule} {Identification} 2016:
  automated methods}.
\bjtitle{Journal of Cheminformatics}
\bvolume{9}(\bissue{1}),
\bfpage{22}
(\byear{2017}).
\doiurl{10.1186/s13321-017-0207-1}.
Accessed 2021-09-27
\end{barticle}
\endbibitem

\bibitem{chemspider}
\begin{barticle}
\bauthor{\bsnm{Pence}, \binits{H.E.}},
\bauthor{\bsnm{Williams}, \binits{A.}}:
\batitle{{ChemSpider}: {An} {Online} {Chemical} {Information} {Resource}}.
\bjtitle{Journal of Chemical Education}
\bvolume{87}(\bissue{11}),
\bfpage{1123}--\blpage{1124}
(\byear{2010}).
\doiurl{10.1021/ed100697w}.
\bcomment{Publisher: American Chemical Society}.
Accessed 2023-03-06
\end{barticle}
\endbibitem

\bibitem{id_ood_1}
\begin{botherref}
\oauthor{\bsnm{McCoy}, \binits{R.T.}},
\oauthor{\bsnm{Min}, \binits{J.}},
\oauthor{\bsnm{Linzen}, \binits{T.}}:
{BERTs} of a feather do not generalize together: {Large} variability in
  generalization across models with similar test set performance.
arXiv.
arXiv:1911.02969 [cs]
(2020).
\doiurl{10.48550/arXiv.1911.02969}.
\url{http://arxiv.org/abs/1911.02969}
Accessed 2023-01-16
\end{botherref}
\endbibitem

\bibitem{id_ood_2}
\begin{botherref}
\oauthor{\bsnm{Zhou}, \binits{X.}},
\oauthor{\bsnm{Nie}, \binits{Y.}},
\oauthor{\bsnm{Tan}, \binits{H.}},
\oauthor{\bsnm{Bansal}, \binits{M.}}:
The {Curse} of {Performance} {Instability} in {Analysis} {Datasets}:
  {Consequences}, {Source}, and {Suggestions}.
arXiv.
arXiv:2004.13606 [cs]
(2020).
\doiurl{10.48550/arXiv.2004.13606}.
\url{http://arxiv.org/abs/2004.13606}
Accessed 2023-01-16
\end{botherref}
\endbibitem

\bibitem{id_ood_3}
\begin{botherref}
\oauthor{\bsnm{D'Amour}, \binits{A.}},
\oauthor{\bsnm{Heller}, \binits{K.}},
\oauthor{\bsnm{Moldovan}, \binits{D.}},
\oauthor{\bsnm{Adlam}, \binits{B.}},
\oauthor{\bsnm{Alipanahi}, \binits{B.}},
\oauthor{\bsnm{Beutel}, \binits{A.}},
\oauthor{\bsnm{Chen}, \binits{C.}},
\oauthor{\bsnm{Deaton}, \binits{J.}},
\oauthor{\bsnm{Eisenstein}, \binits{J.}},
\oauthor{\bsnm{Hoffman}, \binits{M.D.}},
\oauthor{\bsnm{Hormozdiari}, \binits{F.}},
\oauthor{\bsnm{Houlsby}, \binits{N.}},
\oauthor{\bsnm{Hou}, \binits{S.}},
\oauthor{\bsnm{Jerfel}, \binits{G.}},
\oauthor{\bsnm{Karthikesalingam}, \binits{A.}},
\oauthor{\bsnm{Lucic}, \binits{M.}},
\oauthor{\bsnm{Ma}, \binits{Y.}},
\oauthor{\bsnm{McLean}, \binits{C.}},
\oauthor{\bsnm{Mincu}, \binits{D.}},
\oauthor{\bsnm{Mitani}, \binits{A.}},
\oauthor{\bsnm{Montanari}, \binits{A.}},
\oauthor{\bsnm{Nado}, \binits{Z.}},
\oauthor{\bsnm{Natarajan}, \binits{V.}},
\oauthor{\bsnm{Nielson}, \binits{C.}},
\oauthor{\bsnm{Osborne}, \binits{T.F.}},
\oauthor{\bsnm{Raman}, \binits{R.}},
\oauthor{\bsnm{Ramasamy}, \binits{K.}},
\oauthor{\bsnm{Sayres}, \binits{R.}},
\oauthor{\bsnm{Schrouff}, \binits{J.}},
\oauthor{\bsnm{Seneviratne}, \binits{M.}},
\oauthor{\bsnm{Sequeira}, \binits{S.}},
\oauthor{\bsnm{Suresh}, \binits{H.}},
\oauthor{\bsnm{Veitch}, \binits{V.}},
\oauthor{\bsnm{Vladymyrov}, \binits{M.}},
\oauthor{\bsnm{Wang}, \binits{X.}},
\oauthor{\bsnm{Webster}, \binits{K.}},
\oauthor{\bsnm{Yadlowsky}, \binits{S.}},
\oauthor{\bsnm{Yun}, \binits{T.}},
\oauthor{\bsnm{Zhai}, \binits{X.}},
\oauthor{\bsnm{Sculley}, \binits{D.}}:
Underspecification {Presents} {Challenges} for {Credibility} in {Modern}
  {Machine} {Learning}.
arXiv.
arXiv:2011.03395 [cs, stat]
(2020).
\doiurl{10.48550/arXiv.2011.03395}.
\url{http://arxiv.org/abs/2011.03395}
Accessed 2023-01-16
\end{botherref}
\endbibitem

\bibitem{msnovelist}
\begin{barticle}
\bauthor{\bsnm{Stravs}, \binits{M.A.}},
\bauthor{\bsnm{Dührkop}, \binits{K.}},
\bauthor{\bsnm{Böcker}, \binits{S.}},
\bauthor{\bsnm{Zamboni}, \binits{N.}}:
\batitle{{MSNovelist}: de novo structure generation from mass spectra}.
\bjtitle{Nature Methods}
\bvolume{19}(\bissue{7}),
\bfpage{865}--\blpage{870}
(\byear{2022}).
\doiurl{10.1038/s41592-022-01486-3}.
\bcomment{Number: 7 Publisher: Nature Publishing Group}.
Accessed 2022-12-21
\end{barticle}
\endbibitem

\bibitem{massgenie}
\begin{barticle}
\bauthor{\bsnm{Shrivastava}, \binits{A.D.}},
\bauthor{\bsnm{Swainston}, \binits{N.}},
\bauthor{\bsnm{Samanta}, \binits{S.}},
\bauthor{\bsnm{Roberts}, \binits{I.}},
\bauthor{\bsnm{Wright~Muelas}, \binits{M.}},
\bauthor{\bsnm{Kell}, \binits{D.B.}}:
\batitle{{MassGenie}: {A} {Transformer}-{Based} {Deep} {Learning} {Method} for
  {Identifying} {Small} {Molecules} from {Their} {Mass} {Spectra}}.
\bjtitle{Biomolecules}
\bvolume{11}(\bissue{12}),
\bfpage{1793}
(\byear{2021}).
\doiurl{10.3390/biom11121793}.
Accessed 2021-12-29
\end{barticle}
\endbibitem

\bibitem{cosmic}
\begin{botherref}
\oauthor{\bsnm{Hoffmann}, \binits{M.A.}},
\oauthor{\bsnm{Nothias}, \binits{L.-F.}},
\oauthor{\bsnm{Ludwig}, \binits{M.}},
\oauthor{\bsnm{Fleischauer}, \binits{M.}},
\oauthor{\bsnm{Gentry}, \binits{E.C.}},
\oauthor{\bsnm{Witting}, \binits{M.}},
\oauthor{\bsnm{Dorrestein}, \binits{P.C.}},
\oauthor{\bsnm{Dührkop}, \binits{K.}},
\oauthor{\bsnm{Böcker}, \binits{S.}}:
High-confidence structural annotation of metabolites absent from spectral
  libraries.
Nature Biotechnology,
1--11
(2021).
\doiurl{10.1038/s41587-021-01045-9}.
Bandiera\_abtest: a Cc\_license\_type: cc\_by Cg\_type: Nature Research
  Journals Primary\_atype: Research Publisher: Nature Publishing Group
  Subject\_term: Data processing;Molecular biology Subject\_term\_id:
  data-processing;molecular-biology.
Accessed 2021-11-26
\end{botherref}
\endbibitem

\bibitem{msms_significance}
\begin{barticle}
\bauthor{\bsnm{Scheubert}, \binits{K.}},
\bauthor{\bsnm{Hufsky}, \binits{F.}},
\bauthor{\bsnm{Petras}, \binits{D.}},
\bauthor{\bsnm{Wang}, \binits{M.}},
\bauthor{\bsnm{Nothias}, \binits{L.-F.}},
\bauthor{\bsnm{Dührkop}, \binits{K.}},
\bauthor{\bsnm{Bandeira}, \binits{N.}},
\bauthor{\bsnm{Dorrestein}, \binits{P.C.}},
\bauthor{\bsnm{Böcker}, \binits{S.}}:
\batitle{Significance estimation for large scale metabolomics annotations by
  spectral matching}.
\bjtitle{Nature Communications}
\bvolume{8}(\bissue{1}),
\bfpage{1494}
(\byear{2017}).
\doiurl{10.1038/s41467-017-01318-5}.
\bcomment{Number: 1 Publisher: Nature Publishing Group}.
Accessed 2022-12-21
\end{barticle}
\endbibitem

\bibitem{maccs}
\begin{barticle}
\bauthor{\bsnm{Durant}, \binits{J.L.}},
\bauthor{\bsnm{Leland}, \binits{B.A.}},
\bauthor{\bsnm{Henry}, \binits{D.R.}},
\bauthor{\bsnm{Nourse}, \binits{J.G.}}:
\batitle{Reoptimization of {MDL} keys for use in drug discovery}.
\bjtitle{Journal of Chemical Information and Computer Sciences}
\bvolume{42}(\bissue{6}),
\bfpage{1273}--\blpage{1280}
(\byear{2002}).
\doiurl{10.1021/ci010132r}
\end{barticle}
\endbibitem

\bibitem{ecfp}
\begin{barticle}
\bauthor{\bsnm{Rogers}, \binits{D.}},
\bauthor{\bsnm{Hahn}, \binits{M.}}:
\batitle{Extended-{Connectivity} {Fingerprints}}.
\bjtitle{Journal of Chemical Information and Modeling}
\bvolume{50}(\bissue{5}),
\bfpage{742}--\blpage{754}
(\byear{2010}).
\doiurl{10.1021/ci100050t}.
Accessed 2021-05-18
\end{barticle}
\endbibitem

\bibitem{transformer}
\begin{botherref}
\oauthor{\bsnm{Vaswani}, \binits{A.}},
\oauthor{\bsnm{Shazeer}, \binits{N.}},
\oauthor{\bsnm{Parmar}, \binits{N.}},
\oauthor{\bsnm{Uszkoreit}, \binits{J.}},
\oauthor{\bsnm{Jones}, \binits{L.}},
\oauthor{\bsnm{Gomez}, \binits{A.N.}},
\oauthor{\bsnm{Kaiser}, \binits{L.}},
\oauthor{\bsnm{Polosukhin}, \binits{I.}}:
Attention {Is} {All} {You} {Need}.
arXiv:1706.03762 [cs]
(2017).
Accessed 2021-10-27
\end{botherref}
\endbibitem

\bibitem{nmt}
\begin{barticle}
\bauthor{\bsnm{Tan}, \binits{Z.}},
\bauthor{\bsnm{Wang}, \binits{S.}},
\bauthor{\bsnm{Yang}, \binits{Z.}},
\bauthor{\bsnm{Chen}, \binits{G.}},
\bauthor{\bsnm{Huang}, \binits{X.}},
\bauthor{\bsnm{Sun}, \binits{M.}},
\bauthor{\bsnm{Liu}, \binits{Y.}}:
\batitle{Neural machine translation: {A} review of methods, resources, and
  tools}.
\bjtitle{AI Open}
\bvolume{1},
\bfpage{5}--\blpage{21}
(\byear{2020}).
\doiurl{10.1016/j.aiopen.2020.11.001}.
Accessed 2021-11-04
\end{barticle}
\endbibitem

\bibitem{vit}
\begin{botherref}
\oauthor{\bsnm{Dosovitskiy}, \binits{A.}},
\oauthor{\bsnm{Beyer}, \binits{L.}},
\oauthor{\bsnm{Kolesnikov}, \binits{A.}},
\oauthor{\bsnm{Weissenborn}, \binits{D.}},
\oauthor{\bsnm{Zhai}, \binits{X.}},
\oauthor{\bsnm{Unterthiner}, \binits{T.}},
\oauthor{\bsnm{Dehghani}, \binits{M.}},
\oauthor{\bsnm{Minderer}, \binits{M.}},
\oauthor{\bsnm{Heigold}, \binits{G.}},
\oauthor{\bsnm{Gelly}, \binits{S.}},
\oauthor{\bsnm{Uszkoreit}, \binits{J.}},
\oauthor{\bsnm{Houlsby}, \binits{N.}}:
An {Image} is {Worth} 16x16 {Words}: {Transformers} for {Image} {Recognition}
  at {Scale}.
arXiv:2010.11929 [cs]
(2021).
Accessed 2021-11-04
\end{botherref}
\endbibitem

\bibitem{rl_transformer}
\begin{botherref}
\oauthor{\bsnm{Janner}, \binits{M.}},
\oauthor{\bsnm{Li}, \binits{Q.}},
\oauthor{\bsnm{Levine}, \binits{S.}}:
Reinforcement {Learning} as {One} {Big} {Sequence} {Modeling} {Problem}.
arXiv:2106.02039 [cs]
(2021).
Accessed 2021-11-04
\end{botherref}
\endbibitem

\bibitem{memory_gnn}
\begin{botherref}
\oauthor{\bsnm{Khasahmadi}, \binits{A.H.}},
\oauthor{\bsnm{Hassani}, \binits{K.}},
\oauthor{\bsnm{Moradi}, \binits{P.}},
\oauthor{\bsnm{Lee}, \binits{L.}},
\oauthor{\bsnm{Morris}, \binits{Q.}}:
Memory-{Based} {Graph} {Networks}.
arXiv:2002.09518 [cs, stat]
(2020).
Accessed 2021-06-16
\end{botherref}
\endbibitem

\bibitem{graphit}
\begin{botherref}
\oauthor{\bsnm{Mialon}, \binits{G.}},
\oauthor{\bsnm{Chen}, \binits{D.}},
\oauthor{\bsnm{Selosse}, \binits{M.}},
\oauthor{\bsnm{Mairal}, \binits{J.}}:
{GraphiT}: {Encoding} {Graph} {Structure} in {Transformers}.
arXiv:2106.05667 [cs]
(2021).
Accessed 2021-06-16
\end{botherref}
\endbibitem

\bibitem{mat}
\begin{botherref}
\oauthor{\bsnm{Maziarka}, \binits{L.}},
\oauthor{\bsnm{Danel}, \binits{T.}},
\oauthor{\bsnm{Mucha}, \binits{S.}},
\oauthor{\bsnm{Rataj}, \binits{K.}},
\oauthor{\bsnm{Tabor}, \binits{J.}},
\oauthor{\bsnm{Jastrz\c{e}bski}, \binits{S.}}:
Molecule {Attention} {Transformer}.
arXiv:2002.08264 [physics, stat]
(2020).
Accessed 2021-02-22
\end{botherref}
\endbibitem

\bibitem{grover}
\begin{botherref}
\oauthor{\bsnm{Rong}, \binits{Y.}},
\oauthor{\bsnm{Bian}, \binits{Y.}},
\oauthor{\bsnm{Xu}, \binits{T.}},
\oauthor{\bsnm{Xie}, \binits{W.}},
\oauthor{\bsnm{Wei}, \binits{Y.}},
\oauthor{\bsnm{Huang}, \binits{W.}},
\oauthor{\bsnm{Huang}, \binits{J.}}:
Self-{Supervised} {Graph} {Transformer} on {Large}-{Scale} {Molecular} {Data}.
arXiv:2007.02835 [cs, q-bio]
(2020).
Accessed 2021-11-04
\end{botherref}
\endbibitem

\bibitem{gat}
\begin{botherref}
\oauthor{\bsnm{Veličković}, \binits{P.}},
\oauthor{\bsnm{Cucurull}, \binits{G.}},
\oauthor{\bsnm{Casanova}, \binits{A.}},
\oauthor{\bsnm{Romero}, \binits{A.}},
\oauthor{\bsnm{Liò}, \binits{P.}},
\oauthor{\bsnm{Bengio}, \binits{Y.}}:
Graph {Attention} {Networks}.
arXiv:1710.10903 [cs, stat]
(2018).
arXiv: 1710.10903.
Accessed 2021-04-16
\end{botherref}
\endbibitem

\bibitem{ogb_lsc}
\begin{botherref}
\oauthor{\bsnm{Hu}, \binits{W.}},
\oauthor{\bsnm{Fey}, \binits{M.}},
\oauthor{\bsnm{Ren}, \binits{H.}},
\oauthor{\bsnm{Nakata}, \binits{M.}},
\oauthor{\bsnm{Dong}, \binits{Y.}},
\oauthor{\bsnm{Leskovec}, \binits{J.}}:
{OGB}-{LSC}: {A} {Large}-{Scale} {Challenge} for {Machine} {Learning} on
  {Graphs}.
arXiv.
arXiv:2103.09430 [cs]
(2021).
\url{http://arxiv.org/abs/2103.09430}
Accessed 2023-01-05
\end{botherref}
\endbibitem

\bibitem{ogb}
\begin{botherref}
\oauthor{\bsnm{Hu}, \binits{W.}},
\oauthor{\bsnm{Fey}, \binits{M.}},
\oauthor{\bsnm{Zitnik}, \binits{M.}},
\oauthor{\bsnm{Dong}, \binits{Y.}},
\oauthor{\bsnm{Ren}, \binits{H.}},
\oauthor{\bsnm{Liu}, \binits{B.}},
\oauthor{\bsnm{Catasta}, \binits{M.}},
\oauthor{\bsnm{Leskovec}, \binits{J.}}:
Open {Graph} {Benchmark}: {Datasets} for {Machine} {Learning} on {Graphs}.
arXiv:2005.00687 [cs, stat]
(2021).
Accessed 2021-11-04
\end{botherref}
\endbibitem

\bibitem{floyd}
\begin{barticle}
\bauthor{\bsnm{Floyd}, \binits{R.W.}}:
\batitle{Algorithm 97: {Shortest} path}.
\bjtitle{Communications of the ACM}
\bvolume{5}(\bissue{6}),
\bfpage{345}
(\byear{1962}).
\doiurl{10.1145/367766.368168}.
Accessed 2021-11-05
\end{barticle}
\endbibitem

\bibitem{bert}
\begin{botherref}
\oauthor{\bsnm{Devlin}, \binits{J.}},
\oauthor{\bsnm{Chang}, \binits{M.-W.}},
\oauthor{\bsnm{Lee}, \binits{K.}},
\oauthor{\bsnm{Toutanova}, \binits{K.}}:
{BERT}: {Pre}-training of {Deep} {Bidirectional} {Transformers} for {Language}
  {Understanding}.
arXiv:1810.04805 [cs]
(2019).
Accessed 2021-11-04
\end{botherref}
\endbibitem

\bibitem{pcq}
\begin{barticle}
\bauthor{\bsnm{Nakata}, \binits{M.}},
\bauthor{\bsnm{Shimazaki}, \binits{T.}}:
\batitle{{PubChemQC} {Project}: {A} {Large}-{Scale} {First}-{Principles}
  {Electronic} {Structure} {Database} for {Data}-{Driven} {Chemistry}}.
\bjtitle{Journal of Chemical Information and Modeling}
\bvolume{57}(\bissue{6}),
\bfpage{1300}--\blpage{1308}
(\byear{2017}).
\doiurl{10.1021/acs.jcim.7b00083}.
Accessed 2021-11-04
\end{barticle}
\endbibitem

\bibitem{layernorm}
\begin{botherref}
\oauthor{\bsnm{Ba}, \binits{J.L.}},
\oauthor{\bsnm{Kiros}, \binits{J.R.}},
\oauthor{\bsnm{Hinton}, \binits{G.E.}}:
Layer {Normalization}.
arXiv.
arXiv:1607.06450 [cs, stat]
(2016).
\doiurl{10.48550/arXiv.1607.06450}.
\url{http://arxiv.org/abs/1607.06450}
Accessed 2023-02-13
\end{botherref}
\endbibitem

\bibitem{inchi}
\begin{barticle}
\bauthor{\bsnm{Heller}, \binits{S.R.}},
\bauthor{\bsnm{McNaught}, \binits{A.}},
\bauthor{\bsnm{Pletnev}, \binits{I.}},
\bauthor{\bsnm{Stein}, \binits{S.}},
\bauthor{\bsnm{Tchekhovskoi}, \binits{D.}}:
\batitle{{InChI}, the {IUPAC} {International} {Chemical} {Identifier}}.
\bjtitle{Journal of Cheminformatics}
\bvolume{7}(\bissue{1}),
\bfpage{23}
(\byear{2015}).
\doiurl{10.1186/s13321-015-0068-4}.
Accessed 2023-02-13
\end{barticle}
\endbibitem

\bibitem{moleculenet}
\begin{barticle}
\bauthor{\bsnm{Wu}, \binits{Z.}},
\bauthor{\bsnm{Ramsundar}, \binits{B.}},
\bauthor{\bsnm{Feinberg}, \binits{E.N.}},
\bauthor{\bsnm{Gomes}, \binits{J.}},
\bauthor{\bsnm{Geniesse}, \binits{C.}},
\bauthor{\bsnm{Pappu}, \binits{A.S.}},
\bauthor{\bsnm{Leswing}, \binits{K.}},
\bauthor{\bsnm{Pande}, \binits{V.}}:
\batitle{{MoleculeNet}: a benchmark for molecular machine learning}.
\bjtitle{Chemical Science}
\bvolume{9}(\bissue{2}),
\bfpage{513}--\blpage{530}
(\byear{2018}).
\doiurl{10.1039/C7SC02664A}.
\bcomment{Publisher: The Royal Society of Chemistry}.
Accessed 2023-02-13
\end{barticle}
\endbibitem

\bibitem{wln}
\begin{botherref}
\oauthor{\bsnm{Jin}, \binits{W.}},
\oauthor{\bsnm{Coley}, \binits{C.W.}},
\oauthor{\bsnm{Barzilay}, \binits{R.}},
\oauthor{\bsnm{Jaakkola}, \binits{T.}}:
Predicting {Organic} {Reaction} {Outcomes} with {Weisfeiler}-{Lehman}
  {Network}.
arXiv:1709.04555 [cs, stat]
(2017).
Accessed 2021-09-30
\end{botherref}
\endbibitem

\bibitem{pytorch}
\begin{botherref}
\oauthor{\bsnm{Paszke}, \binits{A.}},
\oauthor{\bsnm{Gross}, \binits{S.}},
\oauthor{\bsnm{Massa}, \binits{F.}},
\oauthor{\bsnm{Lerer}, \binits{A.}},
\oauthor{\bsnm{Bradbury}, \binits{J.}},
\oauthor{\bsnm{Chanan}, \binits{G.}},
\oauthor{\bsnm{Killeen}, \binits{T.}},
\oauthor{\bsnm{Lin}, \binits{Z.}},
\oauthor{\bsnm{Gimelshein}, \binits{N.}},
\oauthor{\bsnm{Antiga}, \binits{L.}},
\oauthor{\bsnm{Desmaison}, \binits{A.}},
\oauthor{\bsnm{Köpf}, \binits{A.}},
\oauthor{\bsnm{Yang}, \binits{E.}},
\oauthor{\bsnm{DeVito}, \binits{Z.}},
\oauthor{\bsnm{Raison}, \binits{M.}},
\oauthor{\bsnm{Tejani}, \binits{A.}},
\oauthor{\bsnm{Chilamkurthy}, \binits{S.}},
\oauthor{\bsnm{Steiner}, \binits{B.}},
\oauthor{\bsnm{Fang}, \binits{L.}},
\oauthor{\bsnm{Bai}, \binits{J.}},
\oauthor{\bsnm{Chintala}, \binits{S.}}:
{PyTorch}: {An} {Imperative} {Style}, {High}-{Performance} {Deep} {Learning}
  {Library}.
arXiv:1912.01703 [cs, stat]
(2019).
Accessed 2021-11-04
\end{botherref}
\endbibitem

\bibitem{dgl}
\begin{botherref}
\oauthor{\bsnm{Wang}, \binits{M.}},
\oauthor{\bsnm{Zheng}, \binits{D.}},
\oauthor{\bsnm{Ye}, \binits{Z.}},
\oauthor{\bsnm{Gan}, \binits{Q.}},
\oauthor{\bsnm{Li}, \binits{M.}},
\oauthor{\bsnm{Song}, \binits{X.}},
\oauthor{\bsnm{Zhou}, \binits{J.}},
\oauthor{\bsnm{Ma}, \binits{C.}},
\oauthor{\bsnm{Yu}, \binits{L.}},
\oauthor{\bsnm{Gai}, \binits{Y.}},
\oauthor{\bsnm{Xiao}, \binits{T.}},
\oauthor{\bsnm{He}, \binits{T.}},
\oauthor{\bsnm{Karypis}, \binits{G.}},
\oauthor{\bsnm{Li}, \binits{J.}},
\oauthor{\bsnm{Zhang}, \binits{Z.}}:
Deep {Graph} {Library}: {A} {Graph}-{Centric}, {Highly}-{Performant} {Package}
  for {Graph} {Neural} {Networks}.
arXiv:1909.01315 [cs, stat]
(2020).
Accessed 2021-05-18
\end{botherref}
\endbibitem

\bibitem{dgllife}
\begin{botherref}
\oauthor{\bsnm{Li}, \binits{M.}},
\oauthor{\bsnm{Zhou}, \binits{J.}},
\oauthor{\bsnm{Hu}, \binits{J.}},
\oauthor{\bsnm{Fan}, \binits{W.}},
\oauthor{\bsnm{Zhang}, \binits{Y.}},
\oauthor{\bsnm{Gu}, \binits{Y.}},
\oauthor{\bsnm{Karypis}, \binits{G.}}:
{DGL}-{LifeSci}: {An} {Open}-{Source} {Toolkit} for {Deep} {Learning} on
  {Graphs} in {Life} {Science}.
arXiv:2106.14232 [cs, q-bio]
(2021).
Accessed 2021-09-30
\end{botherref}
\endbibitem

\bibitem{wandb}
\begin{botherref}
\oauthor{\bsnm{Biewald}, \binits{L.}}:
Experiment Tracking with Weights and Biases.
Software available from wandb.com
(2020).
\url{https://www.wandb.com/}
\end{botherref}
\endbibitem

\bibitem{theory_ms_0}
\begin{barticle}
\bauthor{\bsnm{Rosenstock}, \binits{H.M.}},
\bauthor{\bsnm{Wallenstein}, \binits{M.B.}},
\bauthor{\bsnm{Wahrhaftig}, \binits{A.L.}},
\bauthor{\bsnm{Eyring}, \binits{H.}}:
\batitle{Absolute {Rate} {Theory} for {Isolated} {Systems} and the {Mass}
  {Spectra} of {Polyatomic} {Molecules}*}.
\bjtitle{Proceedings of the National Academy of Sciences}
\bvolume{38}(\bissue{8}),
\bfpage{667}--\blpage{678}
(\byear{1952}).
\doiurl{10.1073/pnas.38.8.667}.
\bcomment{Publisher: Proceedings of the National Academy of Sciences}.
Accessed 2023-01-15
\end{barticle}
\endbibitem

\bibitem{theory_ms_1}
\begin{bchapter}
\bauthor{\bsnm{Rosenstock}, \binits{H.M.}},
\bauthor{\bsnm{Krauss}, \binits{M.}}:
\bctitle{{CURRENT} {STATUS} {OF} {THE} {STATISTICAL} {THEORY} {OF} {MASS}
  {SPECTRA}}.
In: \beditor{\bsnm{Elliott}, \binits{R.M.}} (ed.)
\bbtitle{Advances in {Mass} {Spectrometry}},
pp. \bfpage{251}--\blpage{284}.
\bpublisher{Pergamon}, \blocation{???}
(\byear{1963}).
\doiurl{10.1016/B978-0-08-009775-6.50029-3}.
\burl{https://www.sciencedirect.com/science/article/pii/B9780080097756500293}
Accessed 2023-01-15
\end{bchapter}
\endbibitem

\bibitem{theory_ms_2}
\begin{barticle}
\bauthor{\bsnm{Lorquet}, \binits{J.C.}}:
\batitle{Whither the statistical theory of mass spectra?}
\bjtitle{Mass Spectrometry Reviews}
\bvolume{13}(\bissue{3}),
\bfpage{233}--\blpage{257}
(\byear{1994}).
\doiurl{10.1002/mas.1280130304}.
\bcomment{\_eprint:
  https://onlinelibrary.wiley.com/doi/pdf/10.1002/mas.1280130304}.
Accessed 2023-01-15
\end{barticle}
\endbibitem

\bibitem{theory_ms_3}
\begin{barticle}
\bauthor{\bsnm{Lorquet}, \binits{J.C.}}:
\batitle{Landmarks in the theory of mass spectra}.
\bjtitle{International Journal of Mass Spectrometry}
\bvolume{200}(\bissue{1}),
\bfpage{43}--\blpage{56}
(\byear{2000}).
\doiurl{10.1016/S1387-3806(00)00303-1}.
Accessed 2023-01-15
\end{barticle}
\endbibitem

\bibitem{lipid_rules}
\begin{bbook}
\bauthor{\bsnm{Murphy}, \binits{R.C.}}:
\bbtitle{Tandem {Mass} {Spectrometry} of {Lipids}},
(\byear{2014}).
\burl{https://books-rsc-org.myaccess.library.utoronto.ca/books/monograph/1091/Tandem-Mass-Spectrometry-of-Lipids}
Accessed 2023-01-31
\end{bbook}
\endbibitem

\bibitem{abc_xyz}
\begin{barticle}
\bauthor{\bsnm{Steen}, \binits{H.}},
\bauthor{\bsnm{Mann}, \binits{M.}}:
\batitle{The abc's (and xyz's) of peptide sequencing}.
\bjtitle{Nature Reviews Molecular Cell Biology}
\bvolume{5}(\bissue{9}),
\bfpage{699}--\blpage{711}
(\byear{2004}).
\doiurl{10.1038/nrm1468}.
\bcomment{Number: 9 Publisher: Nature Publishing Group}.
Accessed 2023-01-05
\end{barticle}
\endbibitem

\bibitem{prosit}
\begin{barticle}
\bauthor{\bsnm{Gessulat}, \binits{S.}},
\bauthor{\bsnm{Schmidt}, \binits{T.}},
\bauthor{\bsnm{Zolg}, \binits{D.P.}},
\bauthor{\bsnm{Samaras}, \binits{P.}},
\bauthor{\bsnm{Schnatbaum}, \binits{K.}},
\bauthor{\bsnm{Zerweck}, \binits{J.}},
\bauthor{\bsnm{Knaute}, \binits{T.}},
\bauthor{\bsnm{Rechenberger}, \binits{J.}},
\bauthor{\bsnm{Delanghe}, \binits{B.}},
\bauthor{\bsnm{Huhmer}, \binits{A.}},
\bauthor{\bsnm{Reimer}, \binits{U.}},
\bauthor{\bsnm{Ehrlich}, \binits{H.-C.}},
\bauthor{\bsnm{Aiche}, \binits{S.}},
\bauthor{\bsnm{Kuster}, \binits{B.}},
\bauthor{\bsnm{Wilhelm}, \binits{M.}}:
\batitle{Prosit: proteome-wide prediction of peptide tandem mass spectra by
  deep learning}.
\bjtitle{Nature Methods}
\bvolume{16}(\bissue{6}),
\bfpage{509}--\blpage{518}
(\byear{2019}).
\doiurl{10.1038/s41592-019-0426-7}.
\bcomment{Number: 6 Publisher: Nature Publishing Group}.
Accessed 2023-01-05
\end{barticle}
\endbibitem

\bibitem{hq_msms_peptide}
\begin{barticle}
\bauthor{\bsnm{Tiwary}, \binits{S.}},
\bauthor{\bsnm{Levy}, \binits{R.}},
\bauthor{\bsnm{Gutenbrunner}, \binits{P.}},
\bauthor{\bsnm{Salinas~Soto}, \binits{F.}},
\bauthor{\bsnm{Palaniappan}, \binits{K.K.}},
\bauthor{\bsnm{Deming}, \binits{L.}},
\bauthor{\bsnm{Berndl}, \binits{M.}},
\bauthor{\bsnm{Brant}, \binits{A.}},
\bauthor{\bsnm{Cimermancic}, \binits{P.}},
\bauthor{\bsnm{Cox}, \binits{J.}}:
\batitle{High-quality {MS}/{MS} spectrum prediction for data-dependent and
  data-independent acquisition data analysis}.
\bjtitle{Nature Methods}
\bvolume{16}(\bissue{6}),
\bfpage{519}--\blpage{525}
(\byear{2019}).
\doiurl{10.1038/s41592-019-0427-6}.
\bcomment{Number: 6 Publisher: Nature Publishing Group}.
Accessed 2023-01-05
\end{barticle}
\endbibitem

\bibitem{massfrontier}
\begin{botherref}
MassFrontier.
\url{https://www.thermofisher.com/ca/en/home/industrial/mass-spectrometry/liquid-chromatography-mass-spectrometry-lc-ms/lc-ms-software/multi-omics-data-analysis/mass-frontier-spectral-interpretation-software.html}
\end{botherref}
\endbibitem

\bibitem{msfragmenter}
\begin{botherref}
MS Fragmenter.
\url{https://www.acdlabs.com/products/spectrus-platform/ms-fragmenter/}
\end{botherref}
\endbibitem

\bibitem{metfrag1}
\begin{barticle}
\bauthor{\bsnm{Wolf}, \binits{S.}},
\bauthor{\bsnm{Schmidt}, \binits{S.}},
\bauthor{\bsnm{Müller-Hannemann}, \binits{M.}},
\bauthor{\bsnm{Neumann}, \binits{S.}}:
\batitle{In silico fragmentation for computer assisted identification of
  metabolite mass spectra}.
\bjtitle{BMC Bioinformatics}
\bvolume{11}(\bissue{1}),
\bfpage{148}
(\byear{2010}).
\doiurl{10.1186/1471-2105-11-148}.
Accessed 2021-10-27
\end{barticle}
\endbibitem

\bibitem{metfrag2}
\begin{barticle}
\bauthor{\bsnm{Ruttkies}, \binits{C.}},
\bauthor{\bsnm{Schymanski}, \binits{E.L.}},
\bauthor{\bsnm{Wolf}, \binits{S.}},
\bauthor{\bsnm{Hollender}, \binits{J.}},
\bauthor{\bsnm{Neumann}, \binits{S.}}:
\batitle{{MetFrag} relaunched: incorporating strategies beyond in silico
  fragmentation}.
\bjtitle{Journal of Cheminformatics}
\bvolume{8}(\bissue{1}),
\bfpage{3}
(\byear{2016}).
\doiurl{10.1186/s13321-016-0115-9}.
Accessed 2021-10-27
\end{barticle}
\endbibitem

\bibitem{msfinder1}
\begin{barticle}
\bauthor{\bsnm{Tsugawa}, \binits{H.}},
\bauthor{\bsnm{Kind}, \binits{T.}},
\bauthor{\bsnm{Nakabayashi}, \binits{R.}},
\bauthor{\bsnm{Yukihira}, \binits{D.}},
\bauthor{\bsnm{Tanaka}, \binits{W.}},
\bauthor{\bsnm{Cajka}, \binits{T.}},
\bauthor{\bsnm{Saito}, \binits{K.}},
\bauthor{\bsnm{Fiehn}, \binits{O.}},
\bauthor{\bsnm{Arita}, \binits{M.}}:
\batitle{Hydrogen {Rearrangement} {Rules}: {Computational} {MS}/{MS}
  {Fragmentation} and {Structure} {Elucidation} {Using} {MS}-{FINDER}
  {Software}}.
\bjtitle{Analytical Chemistry}
\bvolume{88}(\bissue{16}),
\bfpage{7946}--\blpage{7958}
(\byear{2016}).
\doiurl{10.1021/acs.analchem.6b00770}.
Accessed 2023-01-31
\end{barticle}
\endbibitem

\bibitem{msfinder2}
\begin{barticle}
\bauthor{\bsnm{Lai}, \binits{Z.}},
\bauthor{\bsnm{Tsugawa}, \binits{H.}},
\bauthor{\bsnm{Wohlgemuth}, \binits{G.}},
\bauthor{\bsnm{Mehta}, \binits{S.}},
\bauthor{\bsnm{Mueller}, \binits{M.}},
\bauthor{\bsnm{Zheng}, \binits{Y.}},
\bauthor{\bsnm{Ogiwara}, \binits{A.}},
\bauthor{\bsnm{Meissen}, \binits{J.}},
\bauthor{\bsnm{Showalter}, \binits{M.}},
\bauthor{\bsnm{Takeuchi}, \binits{K.}},
\bauthor{\bsnm{Kind}, \binits{T.}},
\bauthor{\bsnm{Beal}, \binits{P.}},
\bauthor{\bsnm{Arita}, \binits{M.}},
\bauthor{\bsnm{Fiehn}, \binits{O.}}:
\batitle{Identifying metabolites by integrating metabolome databases with mass
  spectrometry cheminformatics}.
\bjtitle{Nature Methods}
\bvolume{15}(\bissue{1}),
\bfpage{53}--\blpage{56}
(\byear{2018}).
\doiurl{10.1038/nmeth.4512}.
\bcomment{Number: 1 Publisher: Nature Publishing Group}.
Accessed 2023-03-07
\end{barticle}
\endbibitem

\bibitem{magma}
\begin{barticle}
\bauthor{\bsnm{Ridder}, \binits{L.}},
\bauthor{\bsnm{Hooft}, \binits{J.J.J.v.d.}},
\bauthor{\bsnm{Verhoeven}, \binits{S.}}:
\batitle{Automatic {Compound} {Annotation} from {Mass} {Spectrometry} {Data}
  {Using} {MAGMa}}.
\bjtitle{Mass Spectrometry}
\bvolume{3}(\bissue{Special\_Issue\_2}),
\bfpage{0033}--\blpage{0033}
(\byear{2014}).
\doiurl{10.5702/massspectrometry.S0033}
\end{barticle}
\endbibitem

\bibitem{comp_ei_ms_review}
\begin{barticle}
\bauthor{\bsnm{Bauer}, \binits{C.A.}},
\bauthor{\bsnm{Grimme}, \binits{S.}}:
\batitle{How to {Compute} {Electron} {Ionization} {Mass} {Spectra} from {First}
  {Principles}}.
\bjtitle{The Journal of Physical Chemistry A}
\bvolume{120}(\bissue{21}),
\bfpage{3755}--\blpage{3766}
(\byear{2016}).
\doiurl{10.1021/acs.jpca.6b02907}.
\bcomment{Publisher: American Chemical Society}.
Accessed 2022-11-09
\end{barticle}
\endbibitem

\bibitem{comp_ei_ms_1}
\begin{barticle}
\bauthor{\bsnm{Irikura}, \binits{K.K.}}:
\batitle{Ab {Initio} {Computation} of {Energy} {Deposition} {During} {Electron}
  {Ionization} of {Molecules}}.
\bjtitle{The Journal of Physical Chemistry A}
\bvolume{121}(\bissue{40}),
\bfpage{7751}--\blpage{7760}
(\byear{2017}).
\doiurl{10.1021/acs.jpca.7b07993}.
\bcomment{Publisher: American Chemical Society}.
Accessed 2023-01-15
\end{barticle}
\endbibitem

\bibitem{comp_ei_ms_2}
\begin{barticle}
\bauthor{\bsnm{Guerra}, \binits{M.}},
\bauthor{\bsnm{Parente}, \binits{F.}},
\bauthor{\bsnm{Indelicato}, \binits{P.}},
\bauthor{\bsnm{Santos}, \binits{J.P.}}:
\batitle{Modified binary encounter {Bethe} model for electron-impact
  ionization}.
\bjtitle{International Journal of Mass Spectrometry}
\bvolume{313},
\bfpage{1}--\blpage{7}
(\byear{2012}).
\doiurl{10.1016/j.ijms.2011.12.003}.
\bcomment{arXiv:1306.2826 [physics]}.
Accessed 2023-01-15
\end{barticle}
\endbibitem

\bibitem{qceims}
\begin{barticle}
\bauthor{\bsnm{Spackman}, \binits{P.R.}},
\bauthor{\bsnm{Bohman}, \binits{B.}},
\bauthor{\bsnm{Karton}, \binits{A.}},
\bauthor{\bsnm{Jayatilaka}, \binits{D.}}:
\batitle{Quantum chemical electron impact mass spectrum prediction for de novo
  structure elucidation: {Assessment} against experimental reference data and
  comparison to competitive fragmentation modeling}.
\bjtitle{International Journal of Quantum Chemistry}
\bvolume{118}(\bissue{2}),
\bfpage{25460}
(\byear{2018}).
\doiurl{10.1002/qua.25460}.
\bcomment{\_eprint: https://onlinelibrary.wiley.com/doi/pdf/10.1002/qua.25460}.
Accessed 2022-11-09
\end{barticle}
\endbibitem

\bibitem{qcxms}
\begin{barticle}
\bauthor{\bsnm{Koopman}, \binits{J.}},
\bauthor{\bsnm{Grimme}, \binits{S.}}:
\batitle{From {QCEIMS} to {QCxMS}: {A} {Tool} to {Routinely} {Calculate} {CID}
  {Mass} {Spectra} {Using} {Molecular} {Dynamics}}.
\bjtitle{Journal of the American Society for Mass Spectrometry}
\bvolume{32}(\bissue{7}),
\bfpage{1735}--\blpage{1751}
(\byear{2021}).
\doiurl{10.1021/jasms.1c00098}.
Accessed 2021-11-14
\end{barticle}
\endbibitem

\bibitem{qc_good_1}
\begin{barticle}
\bauthor{\bsnm{Schreckenbach}, \binits{S.A.}},
\bauthor{\bsnm{Anderson}, \binits{J.S.M.}},
\bauthor{\bsnm{Koopman}, \binits{J.}},
\bauthor{\bsnm{Grimme}, \binits{S.}},
\bauthor{\bsnm{Simpson}, \binits{M.J.}},
\bauthor{\bsnm{Jobst}, \binits{K.J.}}:
\batitle{Predicting the {Mass} {Spectra} of {Environmental} {Pollutants}
  {Using} {Computational} {Chemistry}: {A} {Case} {Study} and {Critical}
  {Evaluation}}.
\bjtitle{Journal of the American Society for Mass Spectrometry}
\bvolume{32}(\bissue{6}),
\bfpage{1508}--\blpage{1518}
(\byear{2021}).
\doiurl{10.1021/jasms.1c00078}.
\bcomment{Publisher: American Society for Mass Spectrometry. Published by the
  American Chemical Society. All rights reserved.}
Accessed 2022-11-09
\end{barticle}
\endbibitem

\bibitem{qc_good_2}
\begin{barticle}
\bauthor{\bsnm{Spackman}, \binits{P.R.}},
\bauthor{\bsnm{Bohman}, \binits{B.}},
\bauthor{\bsnm{Karton}, \binits{A.}},
\bauthor{\bsnm{Jayatilaka}, \binits{D.}}:
\batitle{Quantum chemical electron impact mass spectrum prediction for de novo
  structure elucidation: {Assessment} against experimental reference data and
  comparison to competitive fragmentation modeling}.
\bjtitle{International Journal of Quantum Chemistry}
\bvolume{118}(\bissue{2}),
\bfpage{25460}
(\byear{2018}).
\doiurl{10.1002/qua.25460}.
\bcomment{\_eprint: https://onlinelibrary.wiley.com/doi/pdf/10.1002/qua.25460}.
Accessed 2022-11-09
\end{barticle}
\endbibitem

\bibitem{qc_bad}
\begin{barticle}
\bauthor{\bsnm{Wang}, \binits{S.}},
\bauthor{\bsnm{Kind}, \binits{T.}},
\bauthor{\bsnm{Tantillo}, \binits{D.J.}},
\bauthor{\bsnm{Fiehn}, \binits{O.}}:
\batitle{Predicting in silico electron ionization mass spectra using quantum
  chemistry}.
\bjtitle{Journal of Cheminformatics}
\bvolume{12}(\bissue{1}),
\bfpage{63}
(\byear{2020}).
\doiurl{10.1186/s13321-020-00470-3}.
Accessed 2023-01-15
\end{barticle}
\endbibitem

\bibitem{resnet}
\begin{botherref}
\oauthor{\bsnm{He}, \binits{K.}},
\oauthor{\bsnm{Zhang}, \binits{X.}},
\oauthor{\bsnm{Ren}, \binits{S.}},
\oauthor{\bsnm{Sun}, \binits{J.}}:
Deep {Residual} {Learning} for {Image} {Recognition}.
arXiv.
arXiv:1512.03385 [cs]
(2015).
\doiurl{10.48550/arXiv.1512.03385}.
\url{http://arxiv.org/abs/1512.03385}
Accessed 2023-01-16
\end{botherref}
\endbibitem

\bibitem{dendral}
\begin{botherref}
\oauthor{\bsnm{Buchanan}, \binits{B.G.}},
\oauthor{\bsnm{Feigenbaum}, \binits{E.A.}}:
{DENDRAL} and {Meta}-{DENDRAL}: {Their} {Applications} {Dimension}.
Technical report,
STANFORD UNIV CALIF DEPT OF COMPUTER SCIENCE
(February 1978).
\url{https://apps.dtic.mil/sti/citations/ADA054289}
Accessed 2021-11-04
\end{botherref}
\endbibitem

\bibitem{frag_tree1}
\begin{bchapter}
\bauthor{\bsnm{Scheubert}, \binits{K.}},
\bauthor{\bsnm{Hufsky}, \binits{F.}},
\bauthor{\bsnm{Rasche}, \binits{F.}},
\bauthor{\bsnm{Böcker}, \binits{S.}}:
\bctitle{Computing {Fragmentation} {Trees} from {Metabolite} {Multiple} {Mass}
  {Spectrometry} {Data}}.
In: \beditor{\bsnm{Bafna}, \binits{V.}},
\beditor{\bsnm{Sahinalp}, \binits{S.C.}} (eds.)
\bbtitle{Research in {Computational} {Molecular} {Biology}}.
\bsertitle{Lecture {Notes} in {Computer} {Science}},
pp. \bfpage{377}--\blpage{391}.
\bpublisher{Springer},
\blocation{Berlin, Heidelberg}
(\byear{2011}).
\doiurl{10.1007/978-3-642-20036-6\_36}
\end{bchapter}
\endbibitem

\bibitem{frag_tree2}
\begin{barticle}
\bauthor{\bsnm{Böcker}, \binits{S.}},
\bauthor{\bsnm{Dührkop}, \binits{K.}}:
\batitle{Fragmentation trees reloaded}.
\bjtitle{Journal of Cheminformatics}
\bvolume{8}(\bissue{1}),
\bfpage{5}
(\byear{2016}).
\doiurl{10.1186/s13321-016-0116-8}.
Accessed 2021-11-04
\end{barticle}
\endbibitem

\bibitem{csi}
\begin{barticle}
\bauthor{\bsnm{D{\"u}hrkop}, \binits{K.}},
\bauthor{\bsnm{Shen}, \binits{H.}},
\bauthor{\bsnm{Meusel}, \binits{M.}},
\bauthor{\bsnm{Rousu}, \binits{J.}},
\bauthor{\bsnm{B{\"o}cker}, \binits{S.}}:
\batitle{Searching molecular structure databases with tandem mass spectra using
  csi:fingerid}.
\bjtitle{Proceedings of the National Academy of Sciences}
\bvolume{112}(\bissue{41}),
\bfpage{12580}--\blpage{12585}
(\byear{2015})
{\href{https://arxiv.org/abs/https://www.pnas.org/content/112/41/12580.full.pdf}{{https://www.pnas.org/content/112/41/12580.full.pdf}}}.
\doiurl{10.1073/pnas.1509788112}
\end{barticle}
\endbibitem

\bibitem{ss_nn1}
\begin{botherref}
\oauthor{\bsnm{Curry}, \binits{B.U.}},
\oauthor{\bsnm{Rumelhart}, \binits{D.}}:
A {Neural} {Network} {That} {Classifies} {Mass} {Spectra}
(2001).
\url{https://www.semanticscholar.org/paper/A-Neural-Network-That-Classifies-Mass-Spectra-Curry-Rumelhart/09cc37106e61f2fd1f1dc881de24951a6498b354}
Accessed 2021-10-25
\end{botherref}
\endbibitem

\bibitem{ss_nn2}
\begin{botherref}
\oauthor{\bsnm{Lim}, \binits{J.}},
\oauthor{\bsnm{Wong}, \binits{J.}},
\oauthor{\bsnm{Wong}, \binits{M.X.}},
\oauthor{\bsnm{Tan}, \binits{L.H.E.}},
\oauthor{\bsnm{Chieu}, \binits{H.L.}},
\oauthor{\bsnm{Choo}, \binits{D.}},
\oauthor{\bsnm{Neo}, \binits{N.K.N.}}:
Chemical {Structure} {Elucidation} from {Mass} {Spectrometry} by {Matching}
  {Substructures}.
arXiv:1811.07886 [physics, stat]
(2018).
Accessed 2021-11-04
\end{botherref}
\endbibitem

\end{thebibliography}

\clearpage

\beginextendeddata

\section{Extended Data Figures}
\label{s:extended_data_figures}

    \begin{table}[!htb]
        \begin{center}
        \renewcommand{\arraystretch}{1.3}
        \begin{tabular}{c|c|c|c|c|c}
        Configuration & \makecell{Total \\ \# Layers} & Pre-trained & \makecell{ Reinit \\ Layers} & \makecell{Reinit \\ Layernorm} & \makecell{Cosine \\ Similarity} \\
        \hline
        Small & 6 & No & - & - & $.662 \pm .002$\\
        Large & 12 & No & - & - & $.609 \pm .037$\\
        Large+PT & 12 & Yes & 11-12 & No & $.679 \pm .001$\\
        Large+PT+TL & 12 & Yes & 1-12 & No & $.678 \pm .001$\\
        Large+PT+LN & 12 & Yes & 11-12 & Yes & $.686 \pm .001$\\
        \makecell{Large+PT\\+TL+LN} & 12 & Yes & 1-12 & Yes & $\mathbf{.689 \pm .001}$\\
        \end{tabular}
        \end{center}
        \vspace{1em}
        \caption{ A table summarizing the different model configurations in the ablation experiments (all results are on the NIST-Scaffold Test set). The Large configuration used a standard Graphormer module, without pre-training. The Small configuration used a scaled-down Graphormer module ($\approx 27\%$ of the Large model's parameters), also without pre-training. PT refers to initializing the graph transformer and input embedding layers with pre-trained parameters (from the PubChemQC task, see Section \ref{ss:methods:pre-train} for more details) and then fine-tuning. The default fine-tuning strategy (Large+PT) involved randomly re-initializing only the last 2 transformer layers (out of a possible 12) before fine-tuning. LN refers to layer norm parameter re-initialization, and TL refers to re-initializing all of the transformer layers (instead of just the last 2). It appears that pre-training (PT) was essential for training a Large model, and layer norm parameter re-initialization (LN) was also beneficial. Surprisingly, full transformer layer reinitialization (TL) did not seem to result in degraded performance. In fact, the only difference in parameter initialization between Large+PT+TL+LN and Large was the token embedding parameters (which were pre-trained); all other parameters of both models were randomly initialized. This suggests that the token embeddings were the most transferable component of the pre-training procedure, since the other parts of the model could be reset without losing performance. Averages and standard deviations from 10 independently trained models are reported. \label{suptable:ablations}}
    \end{table}

    \begin{figure}[!htb]
        \begin{subfigure}{0.50\linewidth}
            \centering
            \includegraphics[width=0.90\textwidth]{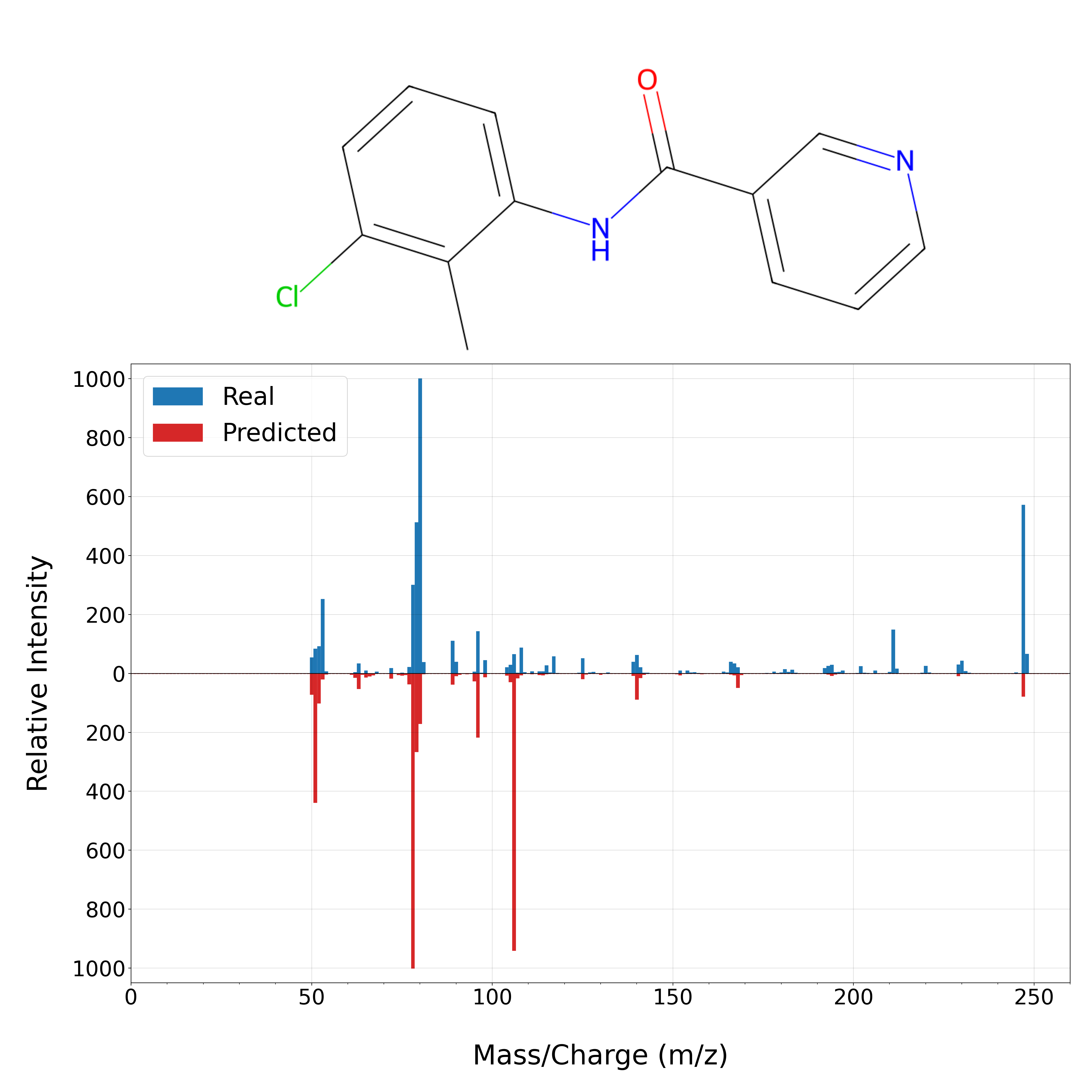}
            \caption{Similarity $=0.40$}
            \label{supfig:spec_examples_1:00}
        \end{subfigure}
        \begin{subfigure}{0.50\linewidth}
            \centering
            \includegraphics[width=0.90\textwidth]{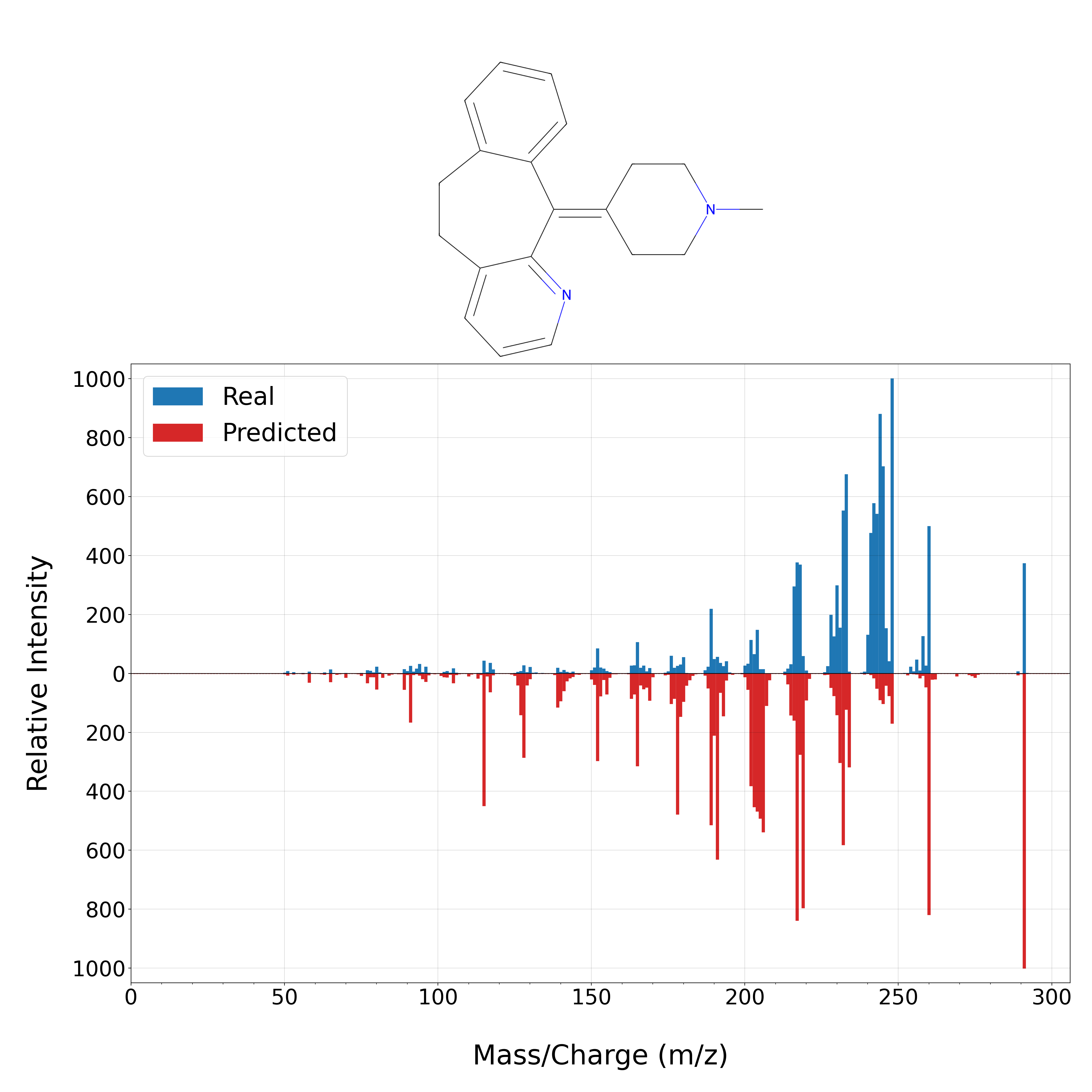}
            \caption{Similarity $=0.46$}
            \label{supfig:spec_examples_1:01}
        \end{subfigure}
        \bigskip
        \begin{subfigure}{0.50\linewidth}
            \centering
            \includegraphics[width=0.90\textwidth]{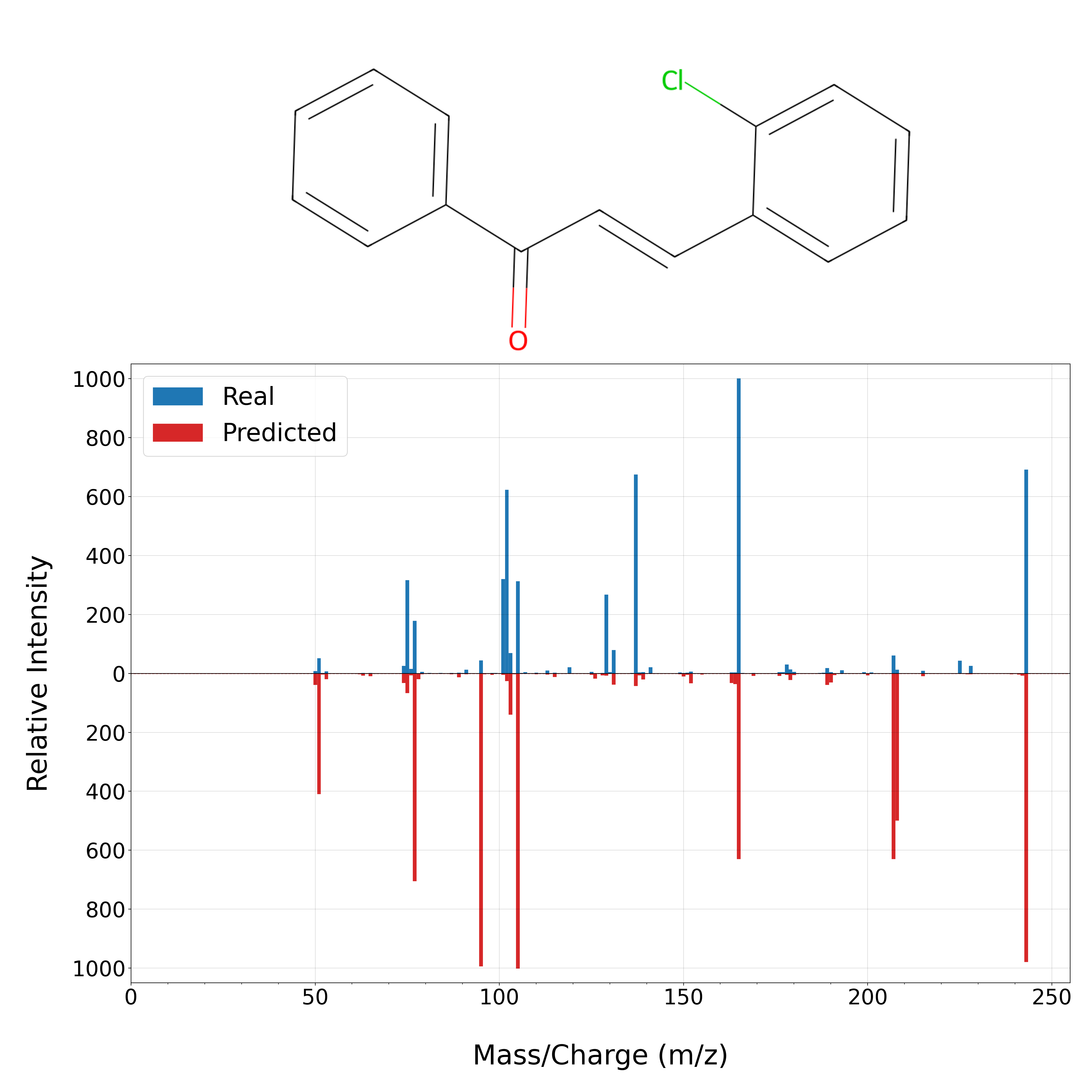}
            \caption{Similarity $=0.54$}
            \label{supfig:spec_examples_1:10}
        \end{subfigure}
        \begin{subfigure}{0.50\linewidth}
            \centering
            \includegraphics[width=0.90\textwidth]{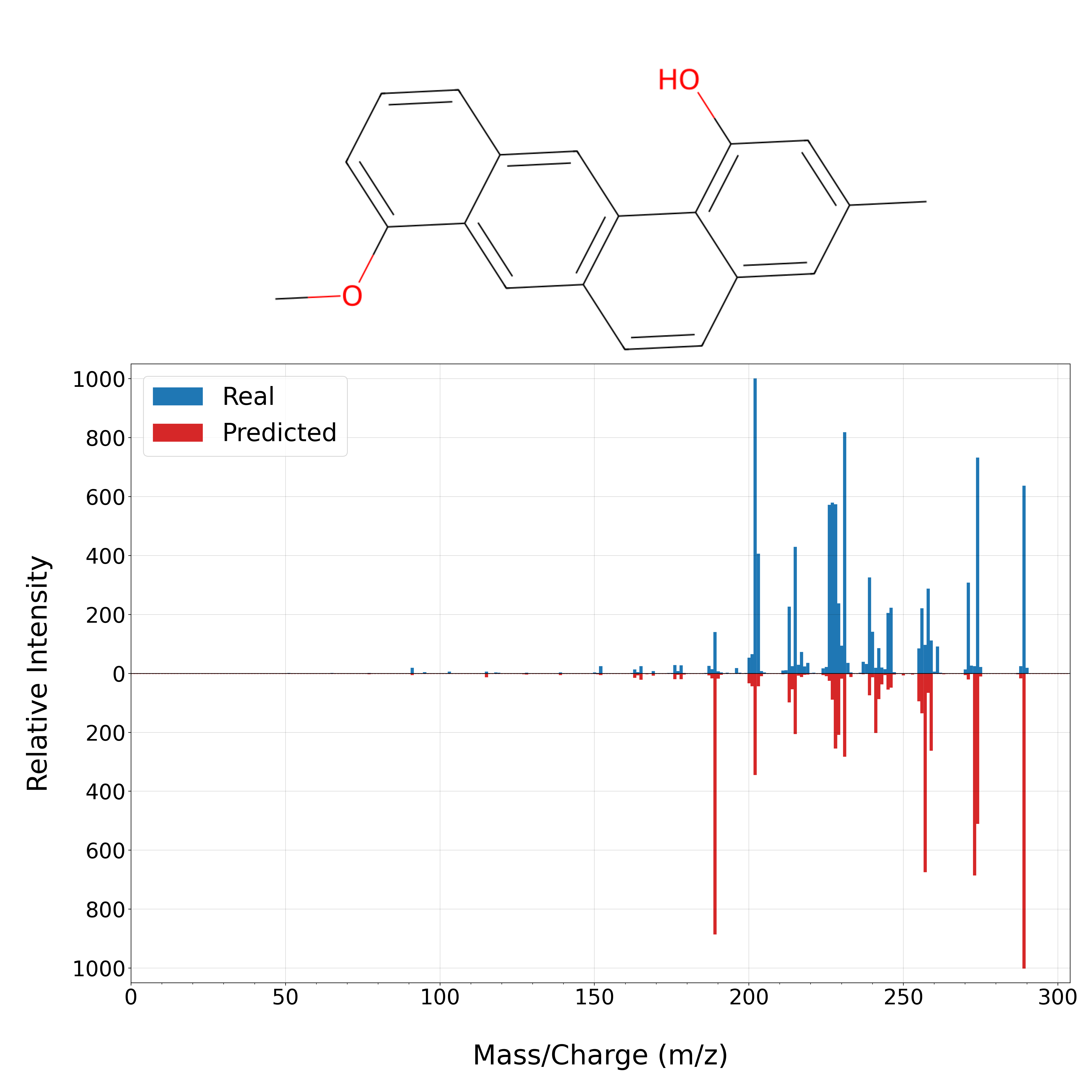}
            \caption{Similarity $=0.58$}
            \label{supfig:spec_examples_1:11}
        \end{subfigure}
        \bigskip
        \begin{subfigure}{0.50\linewidth}
            \centering
            \includegraphics[width=0.90\textwidth]{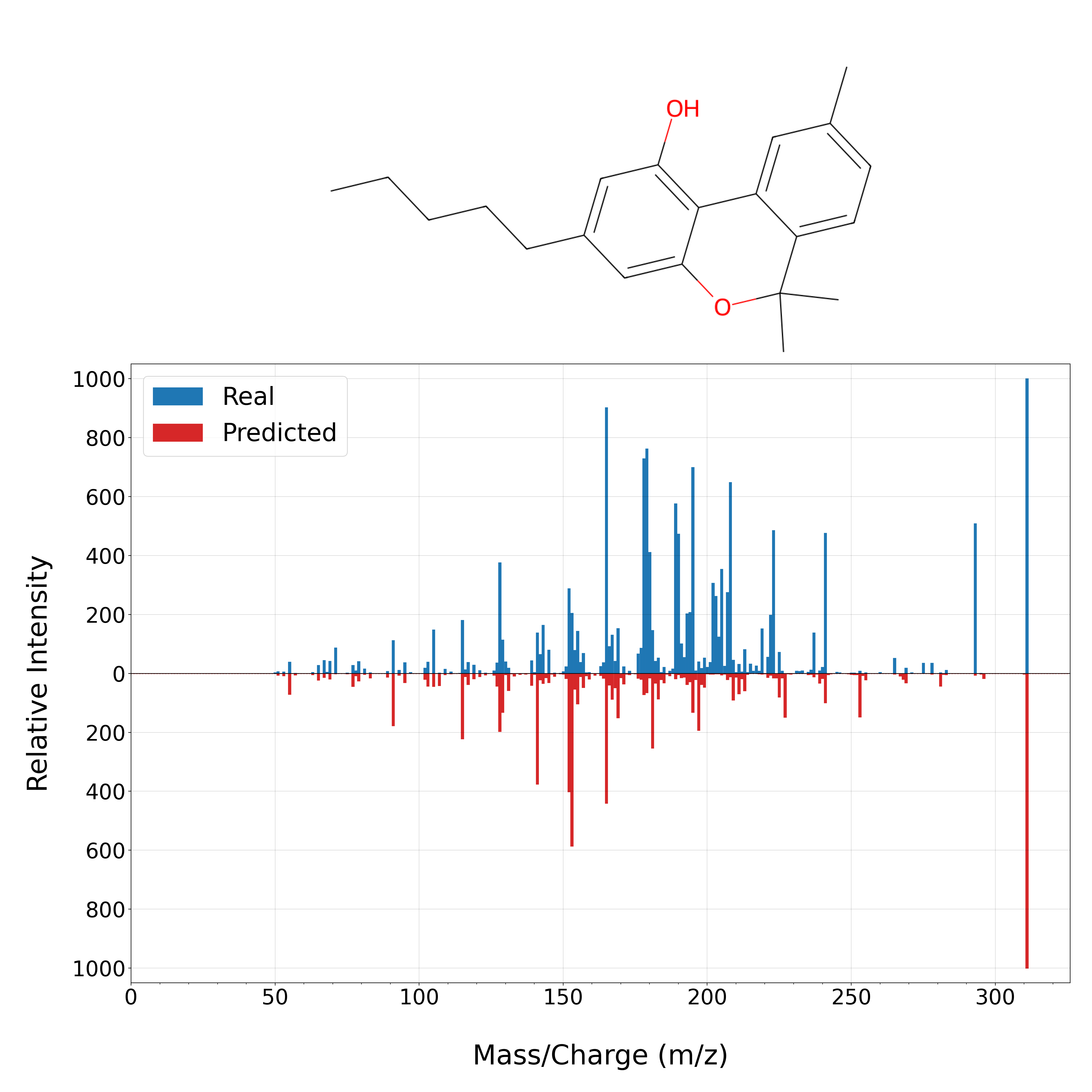}
            \caption{Similarity $=0.60$}
            \label{supfig:spec_examples_1:20}
        \end{subfigure}
        \begin{subfigure}{0.50\linewidth}
            \centering
            \includegraphics[width=0.90\textwidth]{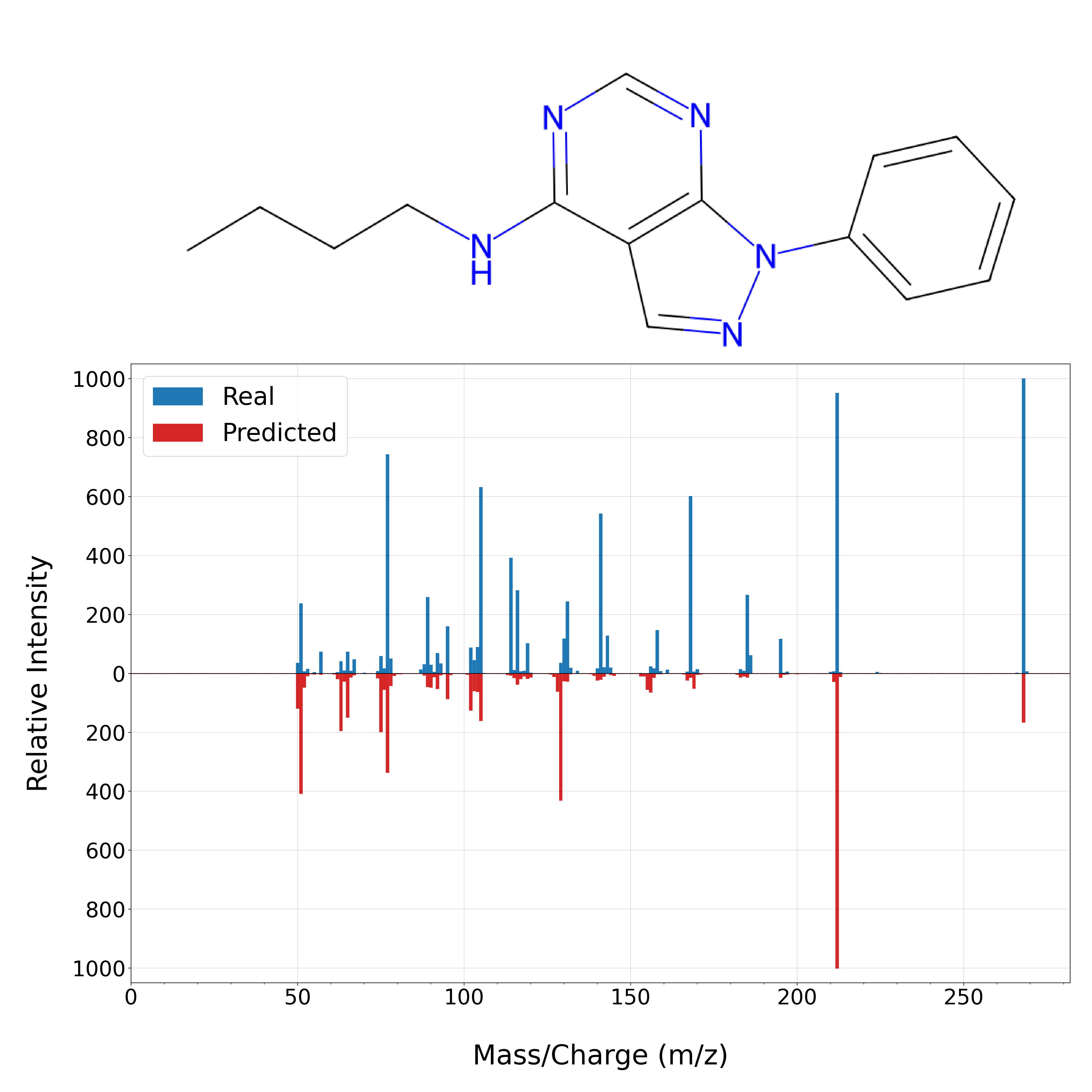}
            \caption{Similarity $=0.65$}
            \label{supfig:spec_examples_1:21}
        \end{subfigure}
        \caption{ \textbf{Additional Spectra (Inaccurate Examples).} True and predicted spectra, roughly covering a range of 0.4 to 0.7 cosine similarity, and their associated attention map visualizations. All spectra correspond to \ce{[M}+\ce{H]+} precursor adducts and have been merged over multiple collision energies. }
        \label{supfig:spec_examples_1}
    \end{figure}
    
    \begin{figure}[!htb]
        \begin{subfigure}{0.50\linewidth}
            \centering
            \includegraphics[width=0.90\textwidth]{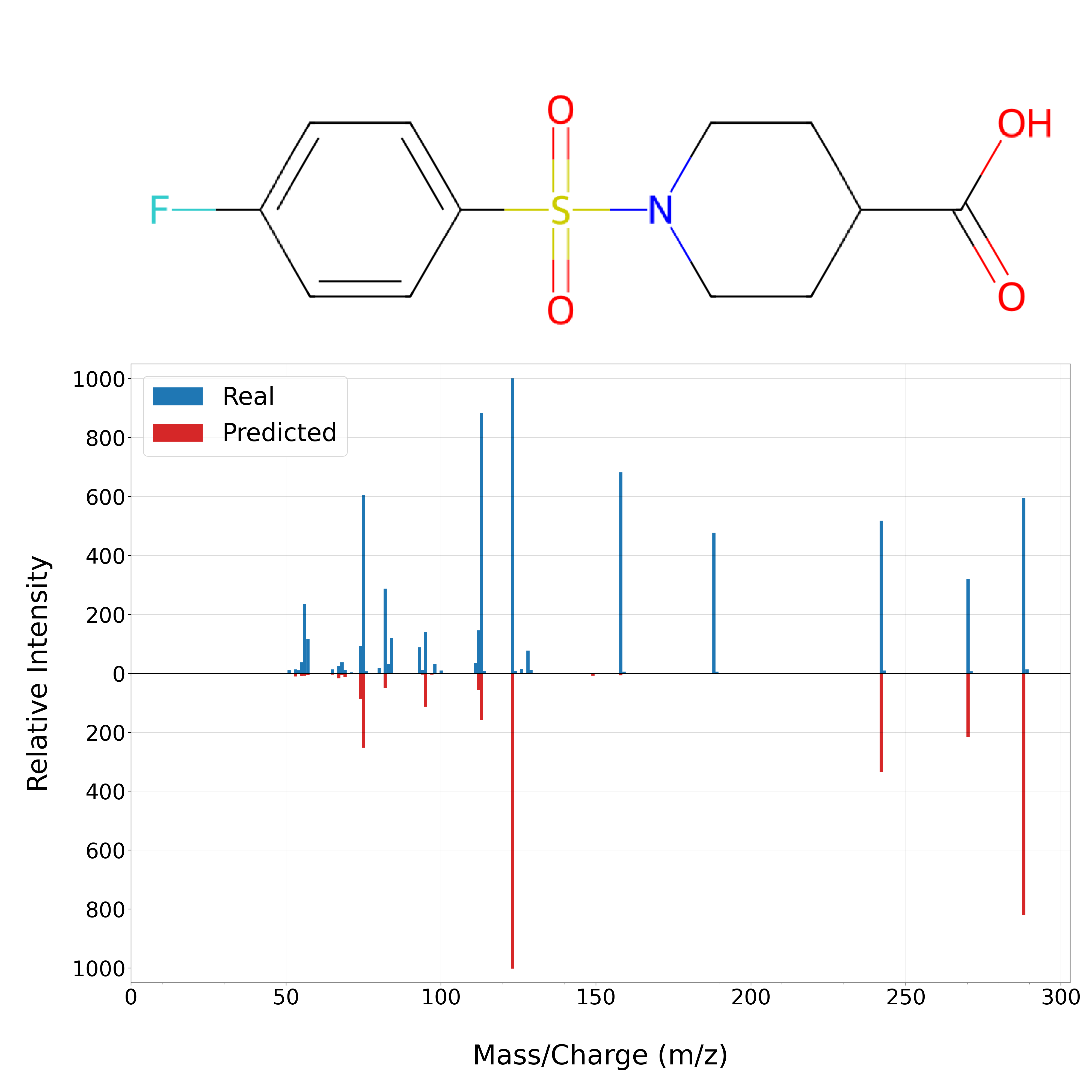}
            \caption{Similarity $=0.76$}
            \label{supfig:spec_examples_2:00}
        \end{subfigure}
        \begin{subfigure}{0.50\linewidth}
            \centering
            \includegraphics[width=0.90\textwidth]{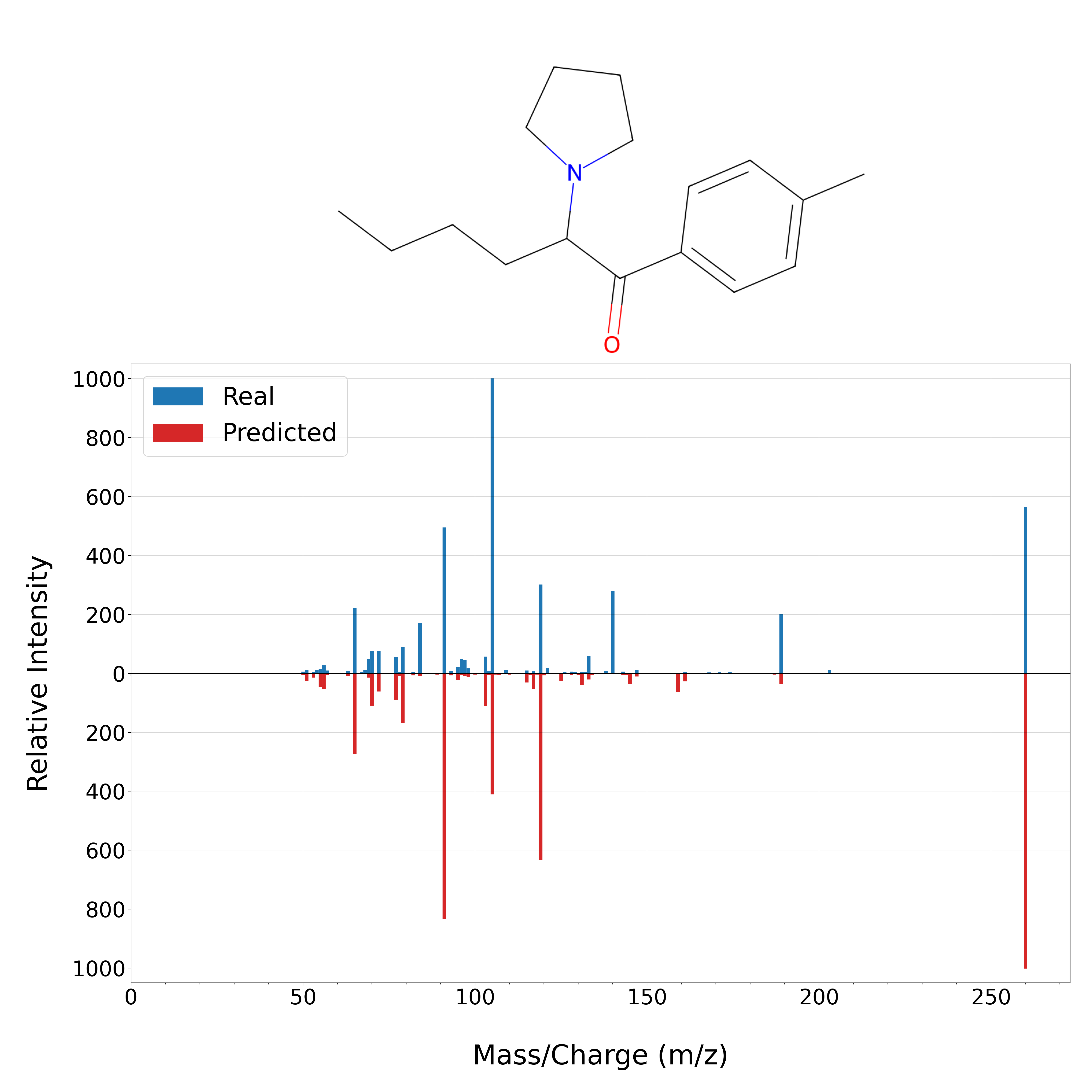}
            \caption{Similarity $=0.79$}
            \label{supfig:spec_examples_2:01}
        \end{subfigure}
        \bigskip
        \begin{subfigure}{0.50\linewidth}
            \centering
            \includegraphics[width=0.90\textwidth]{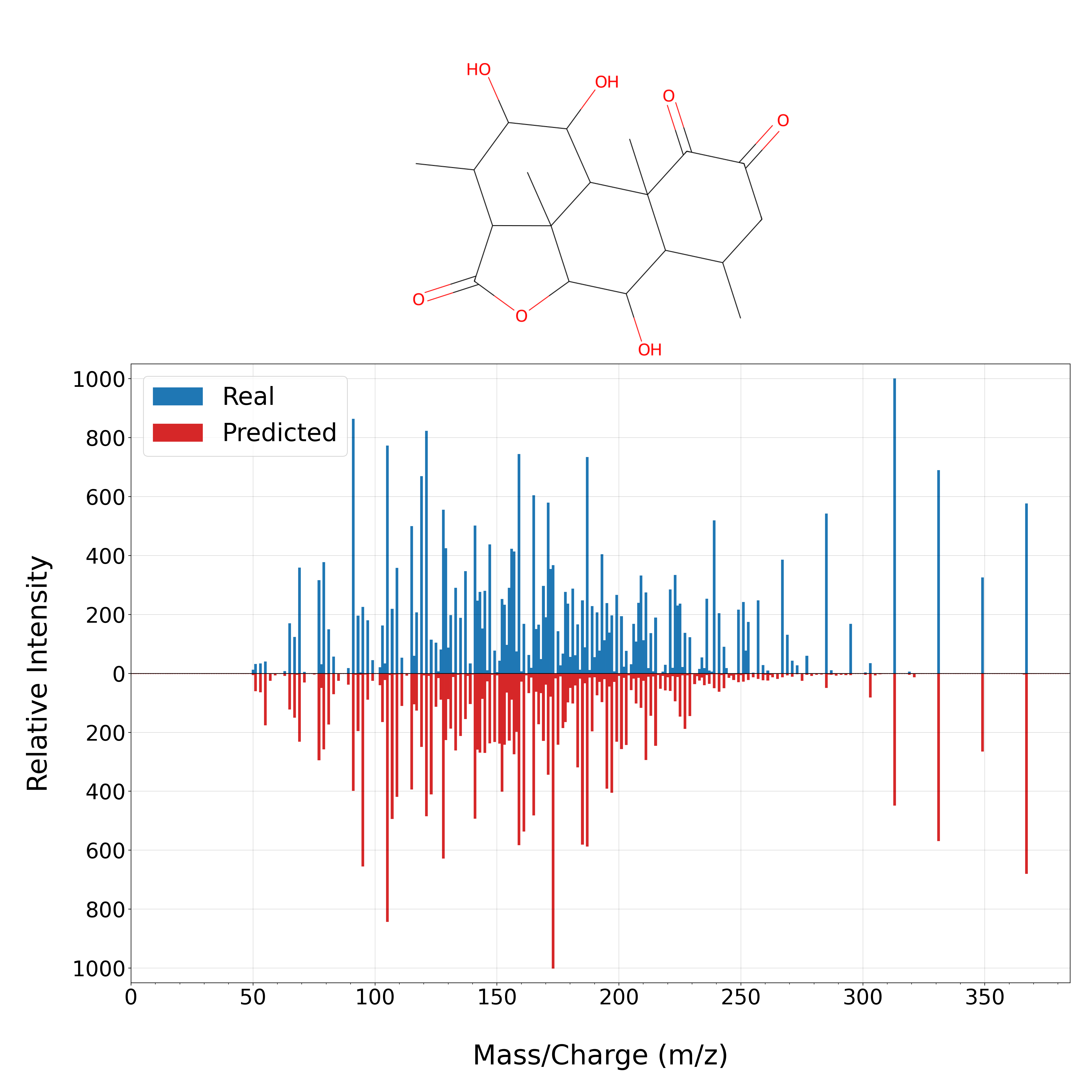}
            \caption{Similarity $=0.82$}
            \label{supfig:spec_examples_2:10}
        \end{subfigure}
        \begin{subfigure}{0.50\linewidth}
            \centering
            \includegraphics[width=0.90\textwidth]{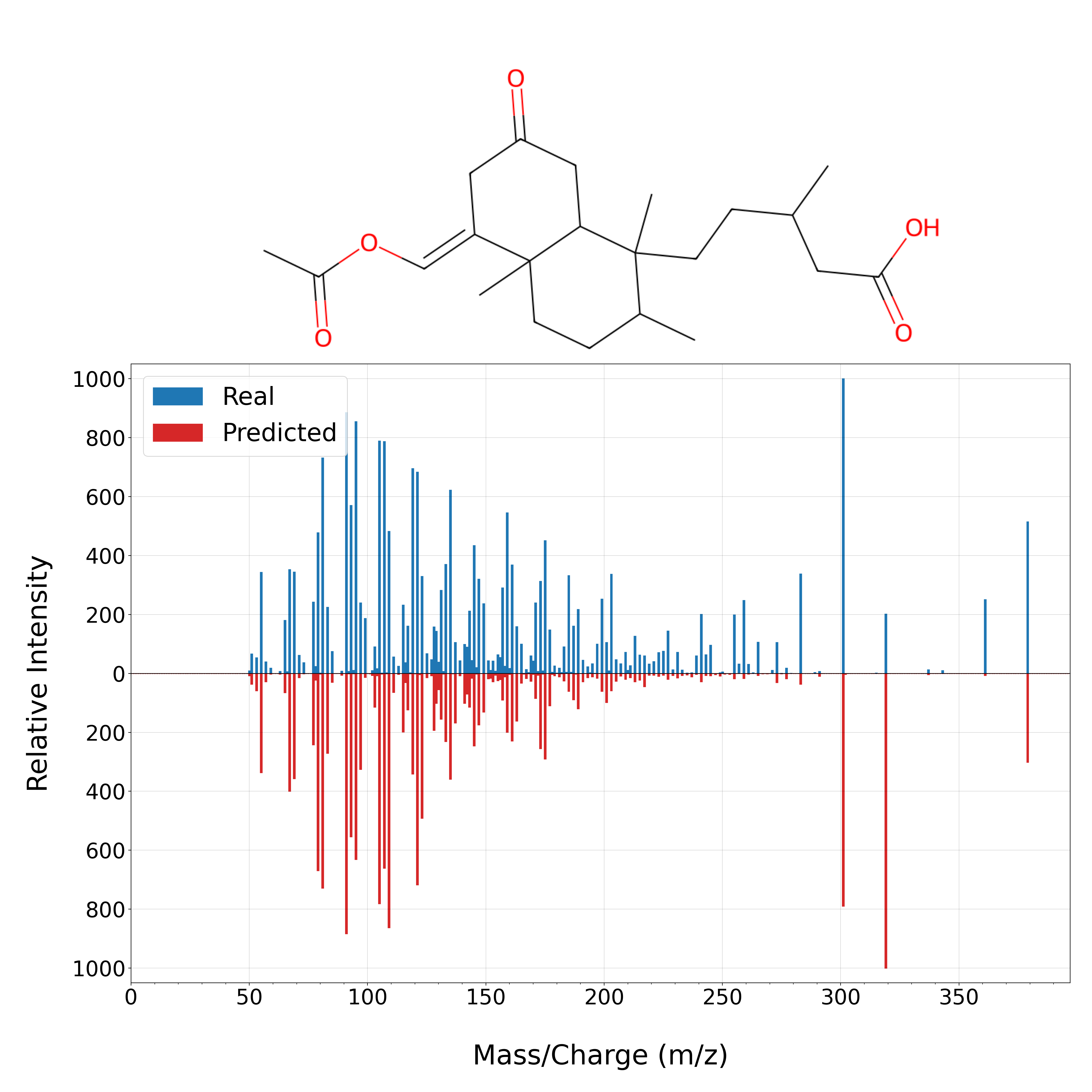}
            \caption{Similarity $=0.88$}
            \label{supfig:spec_examples_2:11}
        \end{subfigure}
        \bigskip
        \begin{subfigure}{0.50\linewidth}
            \centering
            \includegraphics[width=0.90\textwidth]{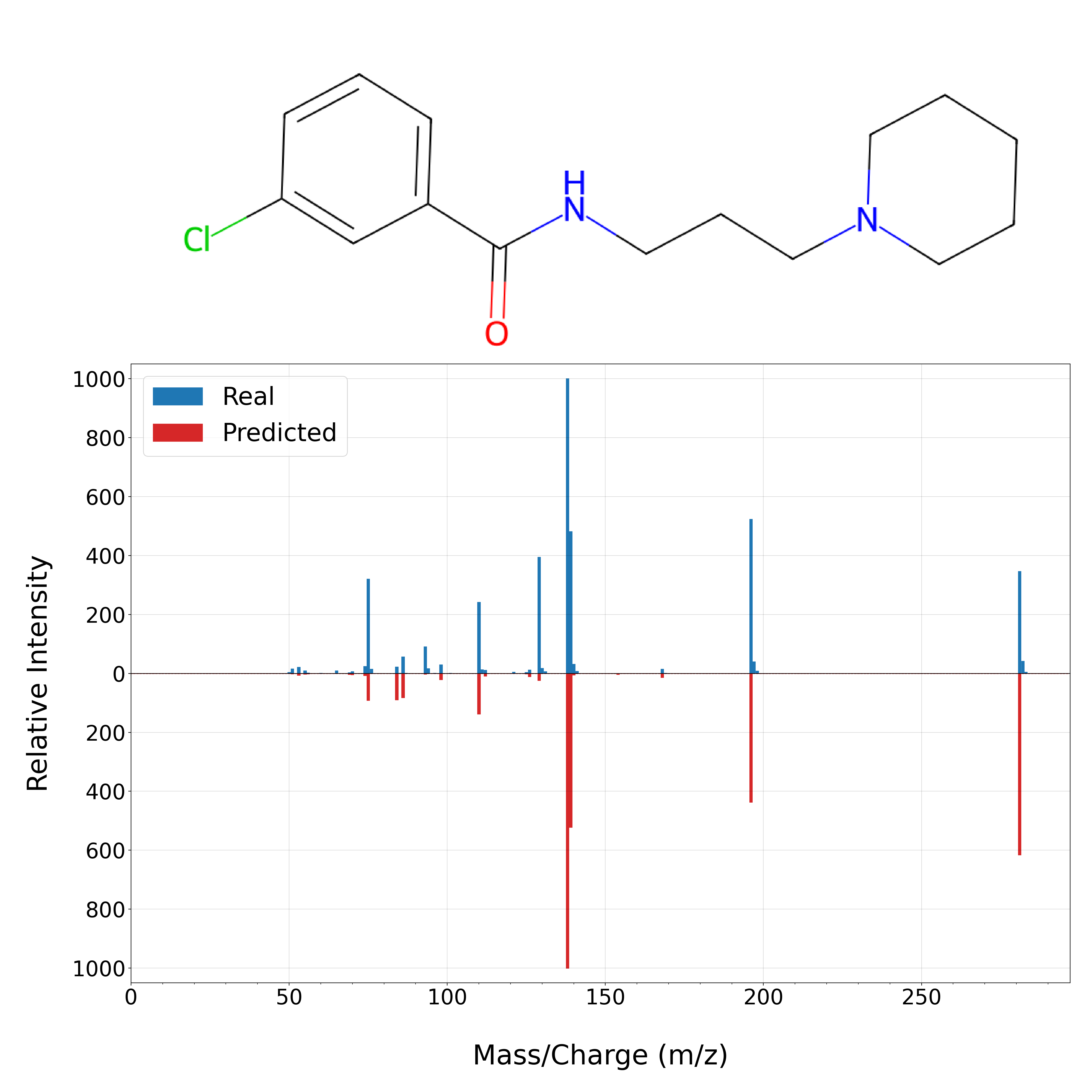}
            \caption{Similarity $=0.92$}
            \label{supfig:spec_examples_2:20}
        \end{subfigure}
        \begin{subfigure}{0.50\linewidth}
            \centering
            \includegraphics[width=0.90\textwidth]{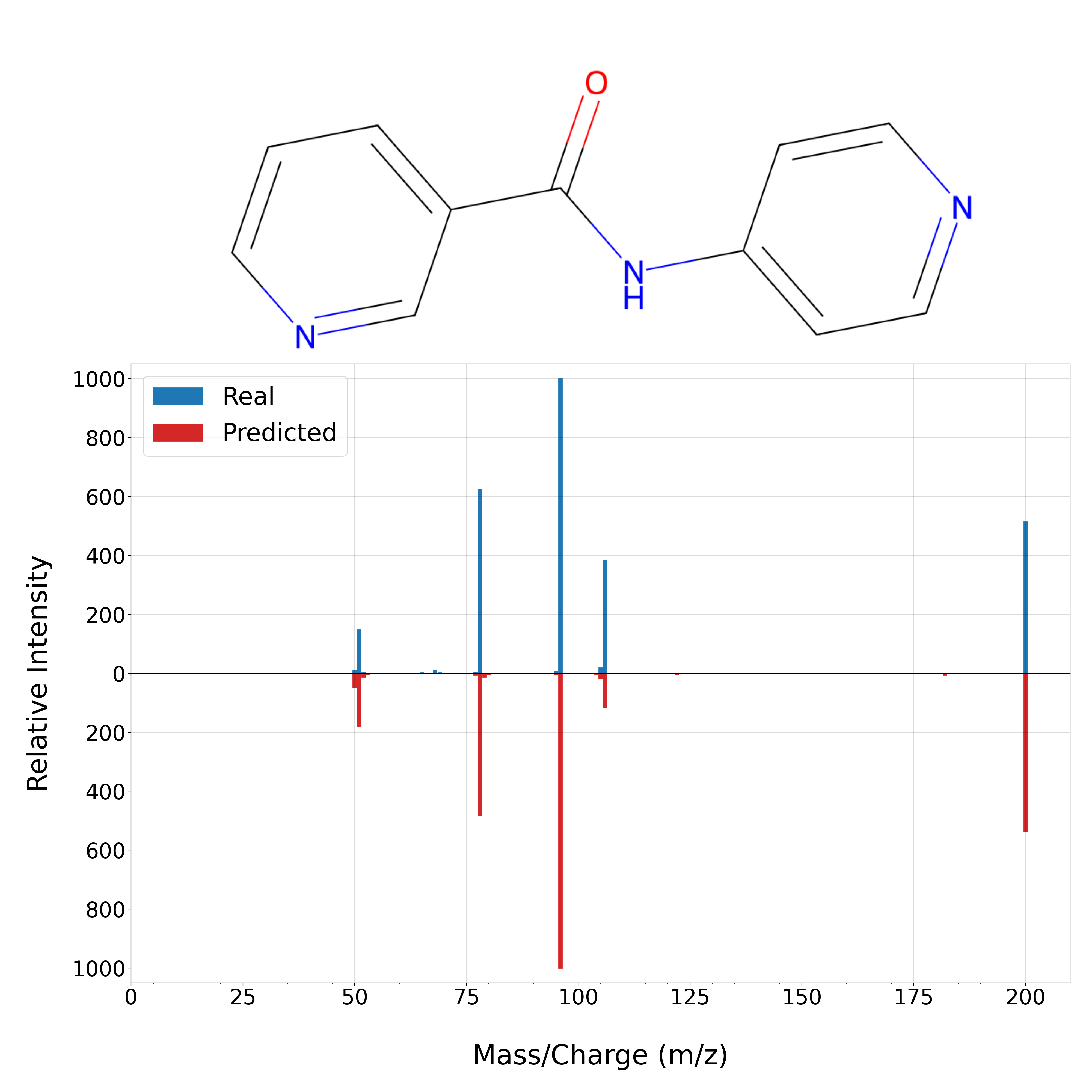}
            \caption{Similarity $=0.97$}
            \label{supfig:spec_examples_2:21}
        \end{subfigure}
        \caption{ \textbf{Additional Spectra (Accurate Examples).} True and predicted spectra, roughly covering a range of 0.7 to 1.0 cosine similarity, and their associated attention map visualizations. All spectra correspond to \ce{[M}+\ce{H]+} precursor adducts and have been merged over multiple collision energies.}
        \label{supfig:spec_examples_2}
    \end{figure}

\clearpage

\section{Supplementary Notes}
\label{ss:sup_notes}

    \subsection{Previous Work in Spectrum Prediction}
    \label{ss:sup_notes:spec_pred_past}

    The statistical theory of mass spectrometry was first formulated in the 1950s \cite{theory_ms_0,theory_ms_1}. While significant progress has been made, an accurate and general model for spectrum prediction remains elusive \cite{theory_ms_2,theory_ms_3}. Over the past 30 years, the proliferation of accessible, powerful computing has resulted in the development of a variety of computational methods for spectrum prediction. Rule-based fragmentation algorithms use experimentally derived fragmentation rules to predict fragment peak locations. They are simple to implement, and can be particularly useful for certain classes of compounds (like lipids \cite{lipid_rules,cfm3,lipidblast} and peptides \cite{abc_xyz,prosit,hq_msms_peptide}) whose regular structure results in predictable fragmentation locations. These methods can also work for more structurally diverse groups of molecules like metabolites \cite{massfrontier,msfragmenter}, but have difficulty generalizing to unseen compounds. Combinatorial methods \cite{metfrag1,metfrag2,msfinder1,msfinder2,magma} rely on iterative bond-breaking to identify possible fragments and their associated peak locations. These methods have high fragment recall, but provide limited information about peak intensity. Competitive Fragmentation Modelling (CFM \cite{cfm,cfm4}) extends the combinatorial approach by defining a probabilistic graphical model over the fragment space. Through modelling bond breakages as a markov process, CFM runs a stochastic simulation to identify which fragments are most likely to appear in the spectrum, and uses this to inform its intensity predictions. While highly interpretable, all combinatorial approaches suffer when the space of fragments explodes, as is the case with larger molecules and particularly molecules with multiple rings.

    In contrast to heuristic and data-driven approaches, first principles methods rely on quantum chemical simulations of the fragmentation process to estimate spectra. A number of techniques have been developed for predicting electron ionization mass spectrometry (EI-MS) \cite{comp_ei_ms_review,comp_ei_ms_1,comp_ei_ms_2}, which is simpler to model than the electrospray ionization spectrometry (ESI-MS) that is commonly used for LC-MS/MS. QCxMS is a Born-Oppenheimer Molecular Dynamics approach originally developed for EI-MS \cite{qceims}, but recently extended to ESI-MS \cite{qcxms}. It uses trajectories sampled from \textit{ab initio} molecular dynamics simulations to define a distribution over fragments. 
    While first-principles approaches promise improved generalization and are highly interpretable, existing implementations require computationally expensive simulations. Such methods are suitable for targeted studies \cite{qc_good_1,qc_good_2}, but have limited utility for high-throughput analysis \cite{qc_bad}. Additionally, current models only support a small range of experimental conditions \cite{qcxms} and lack extensive benchmarking \cite{qc_bad}.

    More recently, deep learning approaches for spectrum prediction have been introduced. The NEIMS model \cite{neims}, developed for EI-MS data, uses a ResNet-inspired architecture \cite{resnet} to predict spectra from a molecular fingerprint representation \cite{ecfp} of the input molecule. Graph neural networks (GNNs) \cite{gnn_msms,esp} have also successfully been applied for ESI-MS prediction. While less interpretable than other methods, deep learning approaches offer a number of compelling advantages. Neural networks can learn complex fragmentation patterns without relying on expensive combinatorial or quantum chemical calculations, which makes their runtimes faster and less dependent on the size of the input molecule. Compared to other data-driven approaches (like CFM), deep learning models can more efficiently leverage larger spectral datasets. They are also more flexible with respect to the properties of the compounds (i.e. inorganic atoms, ring structures) and spectra (i.e. ionization type, collision energy) that are modelled, and can be easily adapted to work with new kinds of data.

    \subsection{Concurrent Work in Spectrum Prediction}
    \label{ss:sup_notes:spec_pred_concurrent}

    A number of deep learning spectrum predictors, developed concurrently to this work, explore different strategies for spectrum prediction. 3DMolMS (3D Molecular Network for Mass Spectra Prediction, \cite{3dmolms}) is perhaps most similar to Massformer, producing relatively low-resolution binned spectrum predictions. However, it incorporates 3D molecule information by running conformer simulations (generated by RDKit \cite{rdkit}) as a preprocessing step on the input molecular graph, and uses a custom GNN-like architecture to levereage the estimated atom locations in 3D space. Despite the limited accuracy of such simple conformer simulations, the additional positional information seems to improve performance, particularly in the low-data regime. Crucially, 3DMolMS is trained on Q-TOF data, unlike MassFormer which is only trained on Orbitrap spectra.  GrAFF (Graph neural network for Approximation via Fixed Formulas of Mass Spectra, \cite{graff}) models spectra as a distribution over a fixed vocabulary of formulae. This method depends on the empirical observation that most peaks in small molecule spectra likely can be explained with either a common small formula or a less common larger formula that can be derived from the precursor formula and a common neutral loss. Formulating spectrum prediction as formula classification allows for arbitrarily high resolution predictions with automatic formula annotations, which is compelling. However, this approach is fundamentally limited by the heuristically-derived set of formulae, and may not generalize to new data. SCARF (Subformulae Classification for Autoregressively Reconstructing Fragmentations, \cite{scarf}) is a closely related method that also performs formula classification. Instead of using a fixed vocabulary of formulae, SCARF parameterizes a distribution over all possible sub-formulae of the precursor formula using a prefix tree. This approach allows auto-regressive formula prediction, which can be trained in a supervised manner using formula annotations derived from formula annotation tools like SIRIUS \cite{sirius} or MAGMa \cite{magma}. A major limitation of SCARF is that it cannot model collision energy, and can only predict spectra that are merged across multiple energies. Additionally, while both GrAFF and SCARF provide formula annotations, they do not provide full substructure annotations like CFM. RASSP (Rapid Approximate Subset-Based Spectra Prediction for Electron Ionization-Mass Spectrometry, \cite{rassp}) is another recent model that attempts to provide more complete annotations, albeit for EI spectra instead of ESI-MS/MS. It works by applying a heuristic bond-breaking algorithm (like CFM) to generate subfragments of the input molecule, then predicting a distribution over these subfragments, which induces a distribution on formulae and can then be mapped to a spectrum. 

    It is not possible to directly compare MassFormer with these other models without a standardized training and evaluation protocol. Three of the four methods (all except GrAFF) provide public code implementations, but differ significantly in terms of the types of spectra that they model: namely, the ionization mode, precursor adducts, and collision energies which are supported. Three of the methods (GrAFF, SCARF, RASSP) present clear advantages in terms of prediction resolution and peak annotations when compared to MassFormer. However, these methods also rely on hand-designed heuristics (either for vocabulary generation or recursive bond breaking), which may affect their generalization on different types of compounds (particularly larger compounds where the heuristics may not apply). Additionally, in terms of prediction speed, MassFormer is likely to be faster than both SCARF (due to its autoregressive formula prediction) and RASSP (due to its combinatorial fragmentation), but may be comparable with 3DMolMS and GrAFF. Rigorous benchmarking is necessary to properly characterize each model's strengths and weaknesses.

    \subsection{Compound Identification}
    \label{ss:sup_notes:comp_id}
    
    The first algorithms for MS-based compound identification arose in the 1960s. DENDRAL and Meta-DENDRAL \cite{dendral} were early AI projects aimed at automatically learning fragmentation rules from MS data, to help identify new molecules. Since then, algorithmic advances and increased data availability have resulted in more powerful methods. Many of the spectrum prediction models from Sections \ref{ss:sup_notes:spec_pred_past} and \ref{ss:sup_notes:spec_pred_concurrent} have also been applied to spectrum identification \cite{massfrontier,msfragmenter,cfm4,metfrag2,msfinder1,msfinder2,magma,qcxms,gnn_msms,esp,neims,scarf,rassp}, typically by generating \textit{in silico} libraries to compare against query spectra. However, there are other methods that do not require directly predicting spectra. Many modern approaches rely on fragmentation trees \cite{frag_tree1,frag_tree2} to represent spectra. Such tree representations are useful because they provide an interpretable description of the fragmentation process, and can be inferred directly from the peak information in the spectrum. Tree kernels are used to compute similarity between spectra \cite{csi}. This allows for the development of kernel algorithms that infer compound identity by predicting properties from the fragmentation tree directly and searching chemical databases to find candidates that match those properties \cite{csi,sirius,msnovelist}. There are also methods for predicting substructure directly from the spectrum, without relying on fragmentation trees. Some works \cite{ss_nn1,ss_nn2} use neural networks to predict substructure labels and find candidate molecules that match those substructures. More recently, transformer approaches for spectrum-to-structure prediction \cite{massgenie} have shown promise with identification of Q-TOF spectra, relying on large libraries of combinatorially simulated spectra \cite{metfrag2} for pre-training.

    \end{document}